\documentclass[pdflatex,sn-mathphys-num, iicol]{sn-jnl}

\usepackage{array}
\usepackage[table]{xcolor}
\usepackage{colortbl}
\usepackage{graphicx}%
\usepackage{caption}
\usepackage{multirow}%
\usepackage{amsmath,amssymb,amsfonts}%
\usepackage{amsthm}%
\usepackage{mathrsfs}%
\usepackage[title]{appendix}%
\usepackage{xcolor}%
\usepackage{textcomp}%
\usepackage{manyfoot}%
\usepackage{booktabs}%
\usepackage{algorithm}%
\usepackage{algorithmicx}%
\usepackage{algpseudocode}%
\usepackage{listings}%
\usepackage{placeins}%
\usepackage{titlesec}%

\theoremstyle{thmstyleone}%
\theoremstyle{thmstyletwo}%

\theoremstyle{thmstylethree}%
\DeclareRobustCommand\onedot{\futurelet\@let@token\@onedot}
\def\@onedot{\ifx\@let@token.\else.\null\fi\xspace}

\newcommand{\red}[1]{#1}
\newcommand{\ray}[1]{#1}

\newcommand{\revision}[1]{#1}

\begin{document}

\title[Ultimate3D: Teaching Multimodal LLMs 3D Camera-Object Relations by Large-Scale Visual Instruction Data]{Ultimate3D: Teaching Multimodal LLMs 3D Camera-Object Relations by Large-Scale Visual Instruction Data}


\author*[1,2]{\fnm{Liu} \sur{He}}\email{riverliuhe71@gmail.com}

\author[2]{\fnm{Yizhi} \sur{Song}}\email{songcx2211@gmail.com}

\author[2]{\fnm{Ruiqi} \sur{Xu}}\email{xu2201@purdue.edu}
\equalcont{These authors contributed equally to this work.}

\author[2]{\fnm{Baris Sarper} \sur{Tezcan}}\email{tezcanb@purdue.edu}
\equalcont{These authors contributed equally to this work.}

\author[1]{\fnm{Xiao} \sur{Zeng}}\email{zenxiao@amazon.com}

\author[1]{\fnm{Albert Y. C.} \sur{Chen}}\email{aycchen@amazon.com}

\author[1]{\fnm{Lu} \sur{Xia}}\email{luxial@amazon.com}

\author[1]{\fnm{Shashwat} \sur{Verma}}\email{shashwv@amazon.com}

\author[1]{\fnm{Sankalp} \sur{Dayal}}\email{sankalpd@amazon.com}

\author[1]{\fnm{Min} \sur{Sun}}\email{minnsun@amazon.com}

\author[1]{\fnm{Cheng-Hao} \sur{Kuo}}\email{chkuo@amazon.com}

\author[2]{\fnm{Daniel} \sur{Aliaga}}\email{aliaga@cs.purdue.edu}

\affil*[1]{\orgname{Amazon}, \country{USA}}

\affil[2]{\orgdiv{Computer Science}, \orgname{Purdue University}, \orgaddress{\street{610 Purdue Mall}, \city{Lafayette}, \postcode{47907}, \state{Indiana}, \country{USA}}}

\vspace{-2em}

\abstract{Multimodal Large Language Models (MLLMs) struggle with accurately capturing camera-object relations, especially for object orientation, camera viewpoint, and camera shots. This stems from the fact that existing MLLMs are trained on images with limited diverse camera-object relations and corresponding textual descriptions. To address this, we propose a synthetic generation pipeline to create large-scale 3D visual instruction datasets. Our framework takes 3D assets as input and uses rendering and diffusion-based image generation models to create photorealistic images preserving precise camera-object relations. Additionally, large language models (LLMs) are used to generate text prompts for guiding visual instruction tuning and controlling image generation. We create Ultimate3D, \revision{a large-scale 3D-grounded 2D visual instruction dataset}. Ultimate3D consists of 240K VQAs with precise camera-object annotations, and corresponding benchmark. MLLMs fine-tuned on our proposed dataset outperform commercial models by a large margin, achieving an average accuracy improvement of \ray{\textbf{29.7\%}} on camera-object relation recognition tasks. Our code, dataset, and benchmark will contribute to broad MLLM applications.}

\maketitle

\section{Introduction}
\label{sec:intro}

\begin{figure*}[t]
    \includegraphics[width=1.0\linewidth]{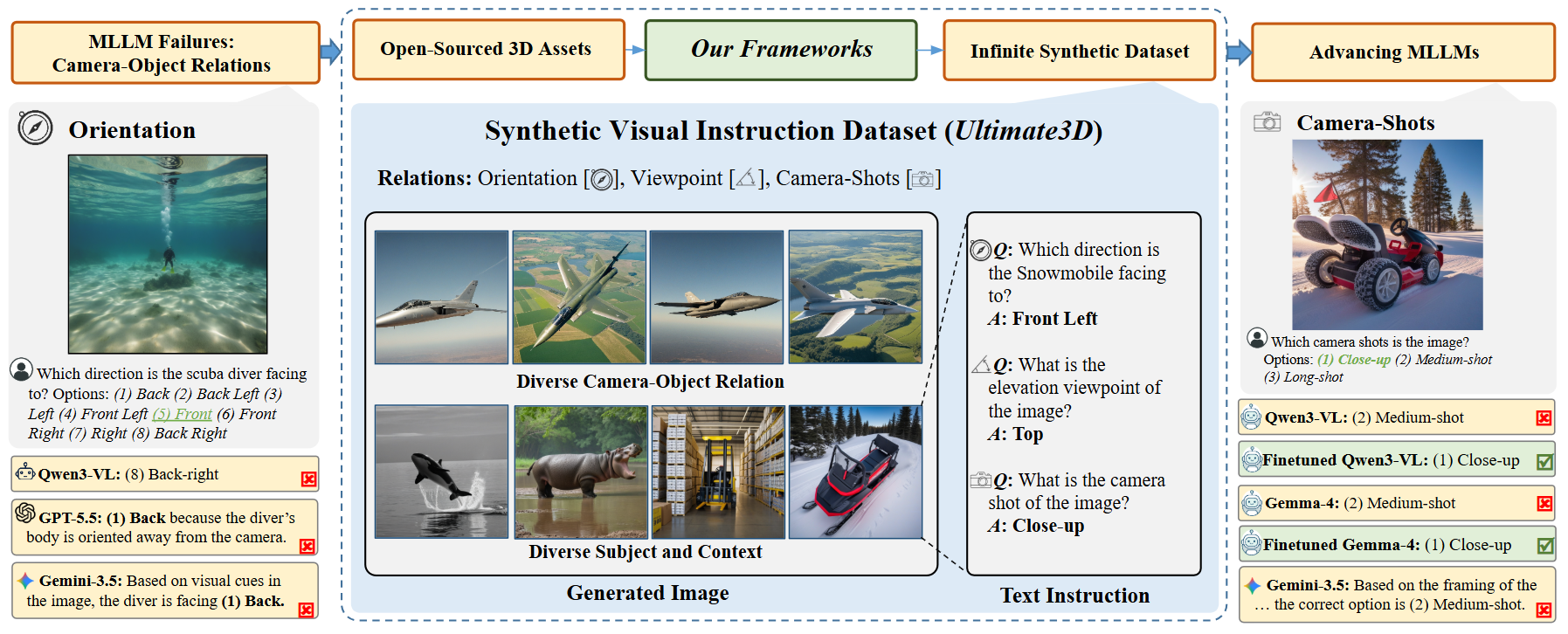}
    \caption{\textbf{Generating Synthetic Visual Instruction Dataset}. Our framework uses open-sourced 3D assets to generate photo-realistic images with precisely controlled camera-object relation. Corresponding text instructions are also generated by Large Language Model (LLM). The generated \textit{Ultimate3D} dataset (240K) and benchmark (8K) advance baseline MLLM models (Gemma-4, Qwen3-VL, etc.) to outperform commercial MLLMs (GPT-5.5, Gemini-3.5, etc.) on camera-object relation recognition tasks. \revision{The left panel illustrates representative MLLM failures, while the right panel provides a direct same-image comparison between base and Ultimate3D-fine-tuned models.}}
    \label{fig:teaser}
\end{figure*}


The advent of Multimodal Large Language Models (MLLMs)~\cite{dai2023instructblip, li2023blip, liu2024visual, liu2024improved, li2024llava, chen2024internvl, lin2024vila, tong2024cambrian, llamavision2024, gemma4, bai2025qwen3, yang2025qwen3} have demonstrated remarkable success on tasks including image captioning, embodied robotic manipulation, and visual question answering (VQA). MLLMs bridge visual understanding from vision encoders~\cite{dosovitskiy2020image, oquab2023dinov2} with powerful verbal capability from large language models (LLMs)~\cite{vicuna2023, touvron2023llama}. Collectively this provides end-to-end inferencing across image and text modalities. In fact, the recently released \ray{Gemini-3.5~\cite{googledeepmind_gemini}, GPT-5.5~\cite{singh2026openaigpt5card} and Claude-4.7-Opus~\cite{anthropic_claude_opus_4_7}} have extended their performances to be comparable to human intelligence in some of the experimented tasks~\cite{yang2023dawn}.

However, research works  have found needs and weaknesses of MLLMs~\cite{tong2024eyes, fu2024blink, yuksekgonul2022and}. Tong \textit{et al.}~\cite{tong2024eyes} point out the deficiency in visual representations learnt by ViT~\cite{dosovitskiy2020image}. In particular, two weakness categories (i.e., orientation/direction, viewpoint/perspective) focus on camera-object relations such as distinguishing simple concepts about object orientation ("right", "front", "back", etc.), camera viewpoint location ("top", "bottom", etc.), and camera-shot styles ("close-up", "long-shot", etc.). As two examples in Fig.~\ref{fig:teaser} show, current SOTA model (i.e., \ray{Gemini-3.5}) struggle with understanding such camera-object relations.
The challenges to improving 3D-aware MLLM abilities on recognizing camera-object relations include: (1) capturing the complexity of the task, and (2) having a dataset containing millions of image-text pairs as is a common requirement for training MLLM on spatial perception tasks~\cite{cheng2024spatialrgpt, chen2024spatialvlm, guo2024regiongpt}. 

Nevertheless, to the best of our knowledge, none of the existing MLLMs or training datasets focus on the aforementioned camera-object relations. Prior methods for MLLM training data generation can be divided into the following two groups:
(1) \textbf{Relabeling of existing real images}. These works focus on generating textual captions by deep learning models given existing images~\cite{chen2024allava, li2024if, zhang2024video, sharifzadeh2024synth, zhang2023llavar, du2023makes}. Specifically, SpatialRGPT~\cite{cheng2024spatialrgpt} and SpatialVLM~\cite{chen2024spatialvlm} address spatial perception and reasoning weakness by providing a comprehensive data annotation pipeline leveraging multiple image segmentation, depth estimation, and camera calibration models to produce dense predictions for real images. But, in practice the accuracy of the predicted labels is not high and the dataset size and distribution is constrained by existing image datasets which predominantly only contain "front" view objects.
%
(2) \textbf{Generating synthetic images}. Typically text-to-image DMs are utilized for synthetic image generation based on provided text prompts~\cite{liu2024synthvlm, du2023training, li2023stablellava}. But, their score-based data curation is not robust to camera-object relations, or generation failures (See Appendix~\ref{supp_sec: aligning}). Alternatively, simulation-based generation pipelines for visual reasoning~\cite{johnson2017clevr, liu2019clevr, greff2022kubric} may provide ground truth camera-object relations, but the generated image quality, complexity, and diversity are still far from realistic. In summary, existing synthetic dataset generation pipelines fail to provide image-text pairs with ground truth camera-object relation, realistic image quality, and diverse image subject and context.

In this work, we propose a synthetic generation pipeline to create unlimited image-text pairs focusing on camera-object relations. With 3D assets (e.g., 3D models with texture, materials, and animation) as input, our framework utilizes a renderer (e.g., Blender) with arbitrary camera-object relation parameters to render a series of 3D ground truth priors (i.e., RGB, depth, segmentation). Those 3D priors are used as comprehensive conditions for stacked diffusion-based generation models (e.g., ControlNet) to generate photorealistic images preserving precise camera-object relations. Simultaneously, LLMs are used to generate versatile text prompts for the controlling of image context and object appearance, and to generate diverse VQA prompts regarding camera-object relations. With our framework, we have created the \textit{Ultimate3D} dataset as a future benchmark for camera-object relations. \revision{The proposed framework serves as a data-generation pipeline that produces photorealistic image-text pairs with precisely controlled camera-object relations from 3D assets. We then use standard fine-tuning protocols to evaluate the effectiveness of Ultimate3D across multiple open-source MLLMs, including LLaVA-, Llama-, Gemma-, and Qwen-based models. Models fine-tuned on Ultimate3D outperform commercial MLLMs by a large margin (i.e., an average 29.7\% accuracy improvement) on recognizing camera-object relations.} 


Our contributions include:

\begin{itemize}
\item A synthetic generation pipeline for unlimited image-text pair creation of diverse camera-object relations.
\item A dataset and benchmark for further improving and evaluating camera-object relations in MLLMs.
\revision{\item Extensive evaluation across multiple open-source MLLMs, showing that Ultimate3D fine-tuning consistently improves camera-object relation recognition and outperforms commercial SOTA models.}

\end{itemize}

\begin{figure*}[t]
    \vspace{-30pt}
  \includegraphics[width=\linewidth]{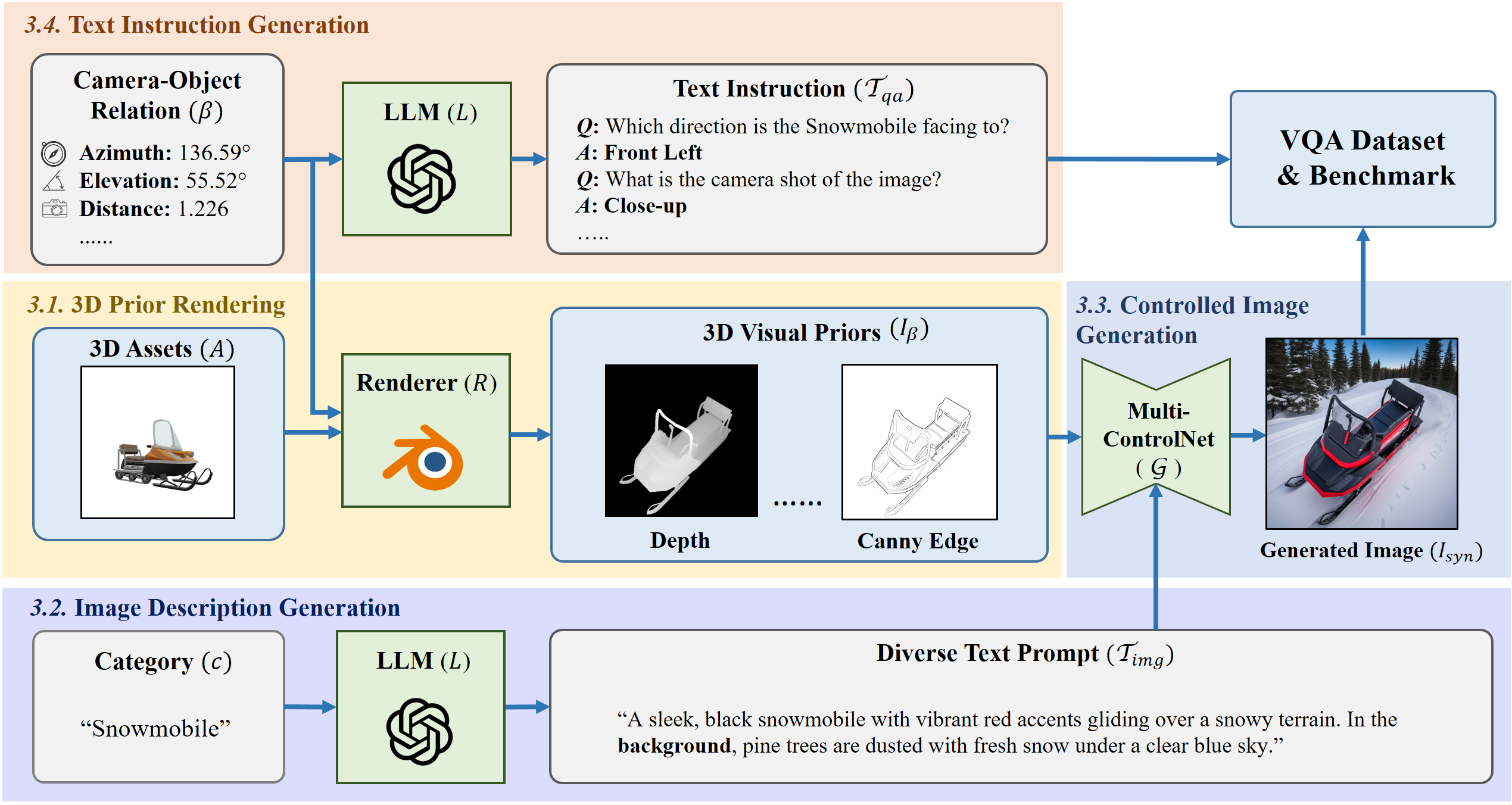}
    \vspace{-10pt}
  \caption{\textbf{Our Framework.} Each shade box corresponds to a section in Sec.~\ref{sec:method}. Given open-sourced 3D assets, our approach leverages a 3D Renderer to generate 3D visual priors ($I_{\beta}$) preserving ground truth camera-object relation ($\beta$). Meanwhile, LLMs take 3D asset category to generate diverse image descriptions ($\mathcal{T}_{img}$) as conditional guidance. Both the 3D visual priors and diverse text prompts are used to generate synthetic images ($I_{syn}$) by multiple ControlNet-based networks. Corresponding text QA instructions ($\mathcal{T}_{qa}$) are generated by LLMs given the ground truth camera-object relation. Our \textit{Ultimate3D} dataset and benchmark (i.e. pairs of $\mathcal{T}_{qa}$ and $I_{syn}$) contribute to fine-tuning and evaluation of MLLMs.}
  \vspace{-13pt}
  \label{fig:framework}
\end{figure*}

\section{Related Works}
\label{sec:relatedworks}

\subsection{Datasets and Benchmarks for MLLMs}
MLLMs use various dataset recipes for visual instruction tuning~\cite{dai2023instructblip, li2023blip, liu2024visual, liu2024improved, li2024llava, chen2024internvl, lin2024vila, tong2024cambrian}. Typically, dozens of datasets, focusing on different areas, are combined to enable the comprehensive abilities of MLLMs. Datasets include image captioning~\cite{lin2014microsoft, young2014image, yu2016modeling}, visual reasoning~\cite{hudson2019gqa, liu2023visual}, general VQA data~\cite{gurari2018vizwiz, wang2023see, liu2024visual} on knowledge grounding~\cite{marino2019ok, lu2022learn}, image understanding~\cite{goyal2017making}, and OCR~\cite{mishra2019ocr,singh2019towards}. Further, most visual instruction datasets are relabeled or created from other existing image-based datasets (e.g., COCO~\cite{lin2014microsoft}, LAION~\cite{schuhmann2022laion}, LVIS~\cite{gupta2019lvis}, Visual Genome~\cite{krishna2017visual}, ADE20K~\cite{zhou2019semantic}). As far as we know, none of the datasets focus on camera-object relations. Thus, the missing ground truth camera-object relation data exacerbates the task for deep learning models, such as SpatialRGPT~\cite{cheng2024spatialrgpt} and SpatialVLM~\cite{chen2024spatialvlm}. In fact, multiple papers have indicated MLLMs' poor distinguishing ability on camera-object relations~\cite{tong2024eyes, fu2024blink}. Our paper fills this gap by providing a generation pipeline, dataset, and benchmark.

In particular for the visual instruction data generation, LLMs, like GPT-4o~\cite{openai2023gpt4}, have been successfully utilized in LLaVA~\cite{liu2024visual, liu2023visual} for dialogue text instruction generation, and used in MMVP~\cite{tong2024eyes} for answer grading. However, generating high-quality images corresponding to proposed text prompts has not been well developed. In this project, we leverage the diverse text generation capability of GPT-4o for the text prompt generation and we provide an image generation pipeline regarding 3D-awareness.

\subsection{Simulation-based Image Generation}
CLEVR~\cite{johnson2017clevr} and Kubric~\cite{greff2022kubric} use 3D engines to generate synthetic images by rendering 3D primitives at multiple arbitrary camera-object relations. Generally, simulation-based image rendering easily preserve ground truth camera-object relations in arbitrary viewpoint settings. However, the generated images are neither photorealistic nor diverse. Realism is limited by the assets' level of detail and textures, and diversity is restricted by the small number of assets. These limitations constrain the usage of synthetic images in MLLM training to only object counting or simple spatial reasoning tasks~\cite{tong2024cambrian}. In this project, we leverage the 3D geometry prior rendered by a 3D renderer to precisely guide the image generation pipeline for preserving ground truth camera-object relation.

\subsection{Generative Images}
Generative models provide advanced image generation by iteratively restoring the original data by denoising from Gaussian noise~\cite{ho2020denoising,song2020denoising,songrefine, song2024imprint, he2023diffusion, he2024kubrick, song2023objectstitch,liao2025longperceptualthoughts, hua2025mmig, sun2025docagent, tang2025caption, he2024coho, tang2025generative, he2023globalmapper,he2023generative,yuan2025newtongen,yuan2025generative}. Open-sourced Stable Diffusion~\cite{rombach2022high}, and SDXL~\cite{podell2023sdxl} enable high-resolution and photo-realistic image generation and editing. 
While generative models provide versatile context and subject, representative works (e.g., AURORA~\cite{krojer2024learning}, EditBench~\cite{wang2023imagen}, Prompt-to-Prompt~\cite{hertz2022prompt}, and MagicBrush~\cite{zhang2024magicbrush}) fail to control camera-object relations in the generated content, and quality guarantees rely on human checking. 
ControlNet~\cite{zhang2023adding}, built on versatile backbone DMs (e.g., SDXL), provides spatial control in the generated images by using various image priors (e.g., Canny edges, semantic segmentation, relative depth).
%

In particular, Ma et al.~\cite{ma2023generating} explore using synthetic image generation for 3D pose estimation by conditioning diffusion models on pose angles. However, their approach does not address advancing MLLMs 3D spatial understanding, nor does it consider VQA instruction generation. Moreover, their reliance on 2D Canny priors (i.e., silhouette) brings severe artifacts for images with backward viewpoint and complex depth variations. In contrast, our method incorporates depth priors to enhance geometric accuracy and leverages SDXL’s superior fidelity to improve image realism. Further discussions and user studies are in Sec.~\ref{sec:ablation} and Sec.~\ref{sec:dataset_quality}.

\section{Method}
\label{sec:method}
Our framework is described in Fig.~\ref{fig:framework} and Appendix~\ref{suppsec:algorithm}. Given input 3D asset $\mathcal{A}$, its category $c$, and arbitrary camera-object relation $\beta$, our framework generates photorealistic image $I_{syn}$ preserving $\beta$, and corresponding text QA instruction $\mathcal{T}_{qa}$. Our pipeline consists of 3D visual prior rendering (Sec.~\ref{sec:prior_gen}), diverse image description generation (Sec.~\ref{sec:text_gen}), controlled image generation (Sec.~\ref{sec:img_gen}), and 3D-aware text instruction generation (Sec.~\ref{sec:vqagen}).

\begin{figure}[h]
    \vspace{-10pt}
  \includegraphics[width=\linewidth]{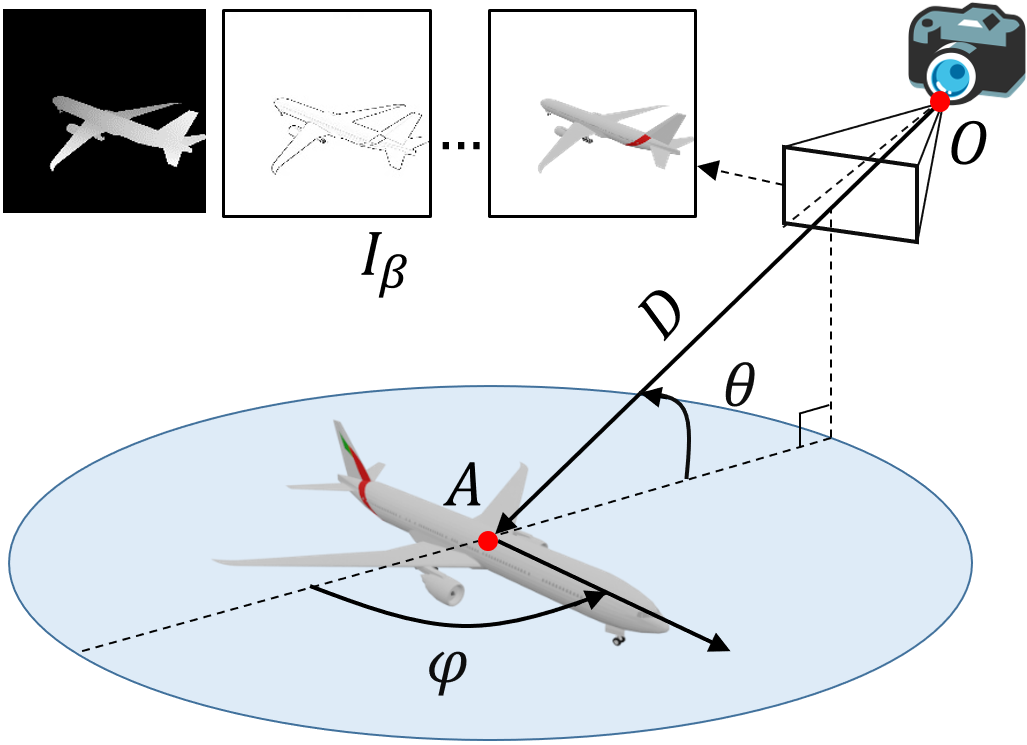}
  \caption{\textbf{3D Visual Prior Rendering.} With a general camera model, our method utilizes a 3D Renderer (e.g., Blender) to render multiple 3D visual priors given arbitrary camera-object relations. $O$: camera center; $A$: object center; $D$: camera-object distance.}
    \vspace{-8pt}
  \label{fig:CAD}
\end{figure}

\subsection{3D Visual Prior Rendering}

Our system renders several 3D visual priors including 2D depth images, Canny-edge images, and segmentation masks from arbitrary viewpoints. These 3D visual priors contain the scene's geometry and will be used together with LLM-generated image descriptions to collectively control the DM-based image generation in Sec.~\ref{sec:img_gen}.

\label{sec:prior_gen}
\revision{
\textbf{Camera-Object Relation Definition.} As Fig.~\ref{fig:CAD} shows, we define camera-object relations using an explicit 3D asset representation and camera-object relative pose.

Each input 3D asset is represented as a textured polygonal mesh
$\mathcal{A}=(V,F,\mathcal{P}),$
where $V=\{\mathbf{v}_i\in\mathbb{R}^3, i=1 ...N\}$ denotes the mesh vertices, $F$ denotes the mesh faces, and $\mathcal{P}$ contains appearance information such as materials and textures. We associate each asset with an object coordinate frame $\mathcal{F}_O$, whose origin is located at the object center $A\in\mathbb{R}^3$ (the red point in Fig.~\ref{fig:CAD}), whose vertical axis specifies the canonical upright direction, and whose unit vector $\mathbf{f}_O$ specifies the canonical object-facing direction.

We further define a world coordinate frame $\mathcal{F}_W$ and a camera coordinate frame $\mathcal{F}_C$. Let $W_{T_{O}}\in SE(3)$ and $W_{T_{C}}\in SE(3)$ denote the object-to-world and camera-to-world rigid transformations, respectively. The relative transformation from the object frame to the camera frame is therefore
$C_{T_{O}} = (W_{T_{C}})^{-1}W_{T_{O}}$, which corresponds to the standard 6-DoF camera-object relative pose.

In our setting, we constrain the asset to remain upright, fix camera roll, and orient the camera optical axis toward the object center $A$. Under these constraints, the camera-object relative pose can be compactly parameterized by
$\beta=(\varphi,\theta,D)$,
consisting of three components:

\begin{itemize}
\item The \textbf{object orientation angle $\varphi\in[0,2\pi)$}, which represents the relative azimuth between the asset's canonical facing direction and the camera-facing direction. Let $\mathbf{f}_W$ denote the canonical facing direction transformed into the world coordinate frame, and let $\mathbf{u}_W$ denote the world-up direction. For any vector $\mathbf{x}$, define its projection onto the horizontal plane as
$
\Pi_h(\mathbf{x})=\mathbf{x}-(\mathbf{x}^{\top}\mathbf{u}_W)\mathbf{u}_W
$,
Let
$
\mathbf{f}_h=
\frac{\Pi_h(\mathbf{f}_W)}
{\|\Pi_h(\mathbf{f}_W)\|_2},
\qquad
\mathbf{q}_h=
\frac{\Pi_h(A-O)}
{\|\Pi_h(A-O)\|_2},
$
where $O$ is the camera aperture center. We define
\[\varphi=
\operatorname{atan2}
\left(
\mathbf{u}_W^\top(\mathbf{q}_h\times\mathbf{f}_h),
\mathbf{q}_h^\top\mathbf{f}_h
\right)
\bmod 2\pi.\]
Thus, $\varphi$ is the counterclockwise azimuth of the asset-facing direction with respect to the camera-facing direction $\overrightarrow{OA}$.

\item The \textbf{viewpoint angle $\theta\in[-\frac{\pi}{2},\frac{\pi}{2}]$}, which represents the elevation of the camera with respect to the horizontal plane of the asset. Let $\mathbf{r}=O-A$ denote the vector from the asset center to the camera. We define
\[
\theta=
\operatorname{atan2}
\left(
\mathbf{r}^{\top}\mathbf{u}_W,
\|\Pi_h(\mathbf{r})\|_2
\right).
\]

\item The \textbf{camera-object distance $D>0$}, defined as the Euclidean distance
between the camera center and the object center, measured in units of the object's size,
\[
D=\frac{\|O-A\|_2}{s(\mathcal{A})},
\]
where $s(\mathcal{A})$ is the largest extent of the asset's axis-aligned bounding box.
In practice each asset is rescaled to unit extent before rendering, so $D$ equals the
camera distance in Blender units. Under our fixed 35mm focal length, $D\approx1$
corresponds to the object spanning the full image frame, and larger $D$ yields a
smaller apparent object size.

\end{itemize}

Thus, $\beta$ should be interpreted as a three-parameter representation of the constrained camera-object relative pose, rather than as a general representation of arbitrary 3D geometry. Compared with~\cite{ma2023generating}, this explicit 3D camera-object parameterization allows us to systematically control object orientation, camera elevation, and camera-object distance during dataset generation.

\textbf{Visual Prior Rendering.}
Given an arbitrary camera-object relation $\beta$, the system renders a corresponding set of geometrically aligned 2D visual priors $I_{\beta}$ (Fig.~\ref{fig:CAD}) using a 3D renderer $R$. We use a perspective camera with a default focal length of 35mm, while its extrinsic parameters are determined by $\beta$. The asset remains upright and the camera is oriented toward its center. The rendered RGB images are further processed to provide Canny edges as additional visual priors.

Because all rendered modalities are produced from the same underlying 3D mesh and camera parameters, they are pixel-aligned and geometrically consistent with the specified camera-object relation $\beta$. Images rendered under different values of $\beta$ therefore provide controlled variations in object orientation, camera viewpoint, and camera-shot distance while preserving exact 3D geometric supervision.
}
\subsection{Diverse Image Description Generation}
\label{sec:text_gen}
Our approach uses GPT-4o as a LLM to generate diverse text prompts for image descriptions ($\mathcal{T}_{img}$) for DM-based image generation; an example is in Fig.~\ref{fig:framework}. Our framework uses system prompting and few-shot prompting tricks which are broadly utilized for image captioning and QA composing~\cite{liu2023visual, liu2024visual}.

For the generation of $\mathcal{T}_{img}$, the LLM is guided to generate diverse descriptions of background context and object details given the object category $c$. The diversity and rationality of the description directly influences the image generation quality (see Appendix~\ref{suppsec:genfailure}). More details are in Appendix~\ref{suppsec:textprompt}.

\subsection{Controlled Image Generation}
\label{sec:img_gen}

We utilize DMs to generate photorealistic images $I_{syn}$ given comprehensive visual prior $I_{\beta}$. A successful generation of $I_{syn}$ shall preserve ground truth camera-object relation information, and be of high fidelity. We first describe a traditional text-to-image DM pipeline, extend it to stacked multiple ControlNets using various images of $I_{\beta}$, and obtain precisely-controlled image generation.

\textbf{Text-to-Image Generation.}
We use text-to-image DM (i.e., SDXL-1.0~\cite{podell2023sdxl}) as our backbone. The denoising U-Net, $\mathcal{G}_{\theta}$ parameterized by $\theta$, conducts iterative denoising diffusion steps of Gaussian noise $\epsilon$ for a total of $T$ time steps. A text-encoder transforms the text prompt $\mathcal{T}_{img}$ into text embedding for guidance injection. The iterative denoising is defined as:

\begin{equation}
    z_{t-1} = \mathcal{G}_{\theta}(\mathcal{T}_{img}, z_{t}, t), t = T, ..., 1
    \label{eq:sdxl}
\end{equation}

where $z_{T} = \epsilon \sim \mathcal{N}(0, I)$, and $z_{0}$ is the latent variable of the generated image. It can be decoded to an RGB image by pretrained image decoder $\mathcal{D}$ as $I_{0} = \mathcal{D}(z_{0})$. 

\textbf{Precise Controlling via Multiple ControlNets.}
ControlNet~\cite{zhang2023adding} guides image generation by injecting additional controlling features derived from various 2D image visual priors. There are dozens of off-the-shelf ControlNets, each pretrained by a single type of visual prior. In our work, we leverage multiple ControlNets to capture comprehensive visual priors from $I_{\beta}$ (depth, Canny edge, etc.) where each one is weighted by coefficients $w$. The denoising process using multiple ControlNets ($\mathcal{C}$) is defined as:

\begin{equation}
    z_{t-1} = \mathcal{G}_{\theta}(\mathcal{T}_{img}, z_{t}, t, \sum_{k=1}^{|\beta|}w_{k} \cdot \mathcal{C}^{k}(I^{k}_{\beta})), t = T, ..., 1.~
    \label{eq:controlnet}
\end{equation}
Similarly, the output generated image $I_{syn} = \mathcal{D}(z_{0})$ and $I_{syn}$ is expected to have ground truth camera-object relations.

\subsection{Text Instruction Generation}
\label{sec:vqagen}
We utilize GPT-4o as a LLM to generate QA pairs on camera-object relation ($\mathcal{T}_{qa}$) for visual instruction tuning (more details in Appendix~\ref{suppsec:vqagen}).

Specifically, the LLM is prompted to generate QA pairs given the camera-object relation $\beta$. We subdivide each component of $\beta$ into categorical types for effective referring through textual response by MLLMs:
\begin{itemize}
    \item Object orientation angle $\varphi$ is classified into eight object orientation types, each covers a bin of $\frac{\pi}{4}$ azimuth (i.e. "right", "front right", "front", "front left", "left", "back left", "back", "back right").
    \item The viewpoint angle $\theta$ is classified into three viewpoint types, each covering a range of $\frac{\pi}{3}$ elevation angle (i.e. "horizontal", "top", "bottom").
    \item The camera-object distance $D$ (Sec.~\ref{sec:prior_gen}) is measured in
    multiples of the object's size, with $D\approx1$ corresponding to the object filling
    the image frame. The three camera-shot types (i.e. "close-up", "medium-shot",
    "long-shot") are decided by two cutoff values, 1.25 and 3.0.
\end{itemize}

Examples of generated multiple-choice style QA pairs as MMVP benchmark~\cite{tong2024eyes} are showed in Fig.~\ref{fig:framework}. For question generation, we prompt LLMs to generate template questions for each prediction task of object orientation, viewpoint type, and camera-shot type. We list all possible options, including the ground truth $\beta$, in random order. MLLMs are expected to choose the option according to $\beta$.

\begin{figure*}[!htbp]
  \vspace{-30pt}
  \centering
  \includegraphics[width=\linewidth]{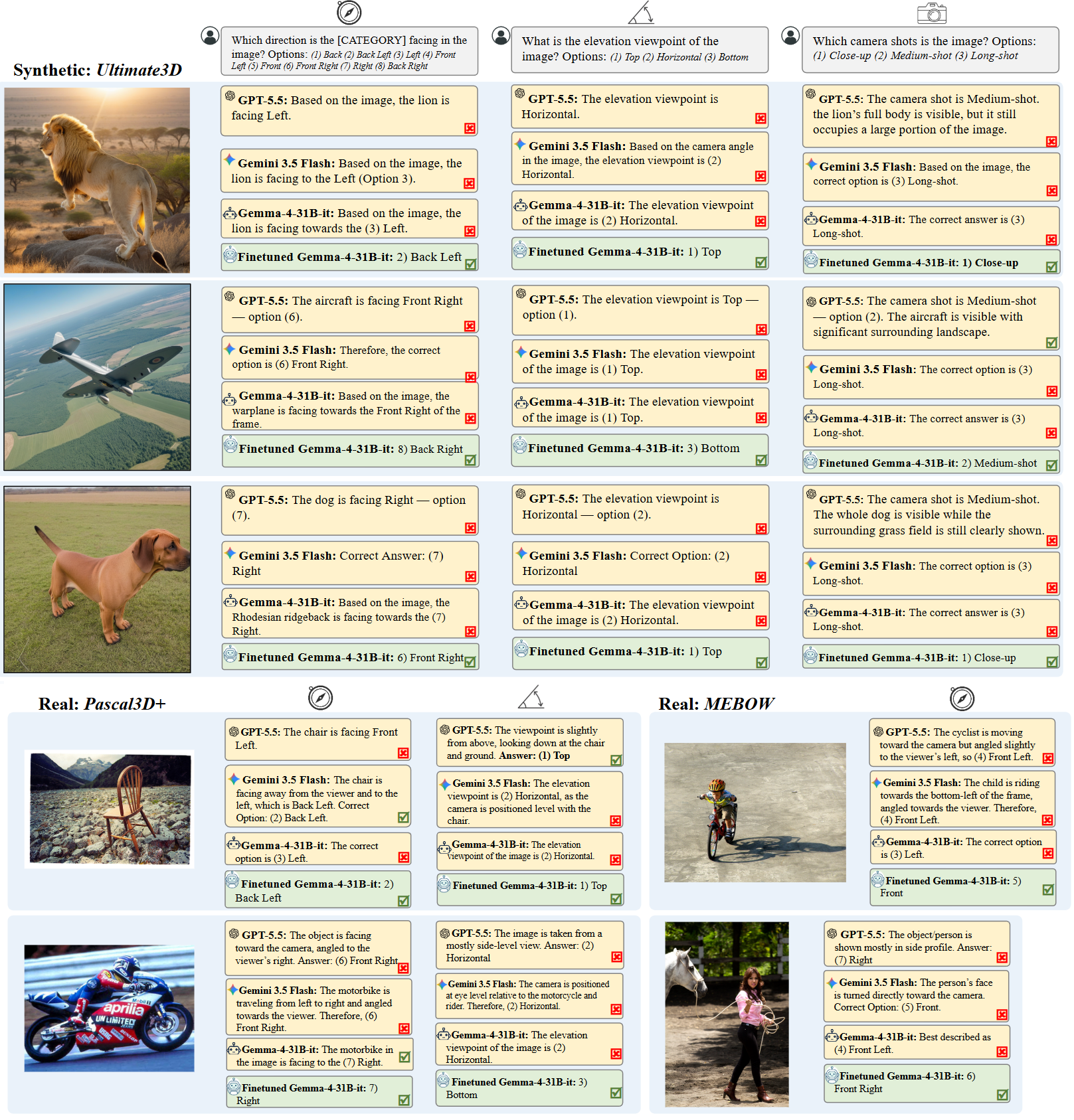}
  \vspace{-17pt}
  \caption{\textbf{Qualitative Results.} We show responses by GPT-5.5, Gemini-3.5-Flash with high thinking, Gemma-4-31B-it, and finetuned Gemma-4-31B-it on the \textit{Ultimate3D} benchmark. Each model is asked questions (in gray boxes) regarding object orientation, camera viewpoint, and camera-shot type, together with the input images on the left. The model responses illustrate that the Gemma-4-31B-it model finetuned on \textit{Ultimate3D} can precisely recognize camera-object relations on synthetic images. GPT-5.5 and Gemini-3.5-Flash with high thinking sometimes provide correct answers, but also exhibit inconsistent predictions across relation types. The original Gemma-4-31B-it frequently fails to produce the correct camera-object relation labels, while the finetuned model provides substantially more accurate and task-relevant answers.}
  \label{fig:qualitatiev}
  \vspace{-12pt}
\end{figure*}

\section{Experiments}
\label{sec:experiments}

\subsection{Visual Instruction Generation}
\label{subsec:dataset}
\textbf{Data Source.}
Our framework uses open-sourced 3D assets datasets from ObjaverseXL~\cite{deitke2023objaverse, deitke2024objaverse} and ShapeNet~\cite{chang2015shapenet}. Partial 3D assets are labeled following the synset list from ImageNet-1K~\cite{russakovsky2015imagenet}. We leverage this subset and define the asset category by its synset label. In particular, we manually select 100 synsets with explicit "facing" orientation from the original list of ImageNet. In total, there are 1196 assets used for synthetic visual instruction generation.

\noindent\textbf{ControlNets.}
We utilize SDXL-1.0~\cite{podell2023sdxl} as the backbone of ControlNets. For selection of multiple visual priors, we found using both Canny edges and relative depth provides the best mix of camera-object relation preservation and image quality. The diffusion step $T=30$, and relative weights for depth and Canny edge ControlNets are 0.5 and 0.8, respectively. Several seconds are needed to generate an image by a single H100 GPU. Ablations and user studies are shown in Sec.~\ref{sec:ablation} and Sec.~\ref{sec:dataset_quality}.

\subsection{Ultimate3D Dataset and Benchmark.}
Our framework creates the \textit{Ultimate3D}, \revision{a large-scale 3D-grounded 2D visual instruction dataset whose samples are generated from explicit 3D assets and controlled camera geometry, providing precise ground-truth camera-object relation annotations.} \textit{Ultimate3D} consists of 85K synthetic images and 18K cropped images of a single human per photograph and corresponding human orientation labels from MEBOW~\cite{wu2020mebow}. For the synthetic images, we use 1180 3D assets covering 100 categories. For each 3D asset, there are 72 possible camera-object relations (8 orientations, 3 viewpoints, and 3 camera shot types) and we render a corresponding synthetic image for each relation. Then, each synthetic image is described by up to 3 textual QAs corresponding to object orientation, camera viewpoint, and camera-shot type, yielding a total of 240K corresponding QAs. Examples of synthetic images are in Appendix~\ref{suppsec:ultimate3D}. The images from MEBOW are useful because otherwise there is a strong lack of 3D assets of human which we find crucial to fine-tuning. The MEBOW images only have one QA corresponding to human orientation.

The \textit{Ultimate3D} benchmark (Fig.~\ref{fig:qualitatiev}) has two sets: (1) Synthetic Set. It contains 1200 synthetic images and 3600 QAs generated by our framework. Each image and QA has been manually reviewed to ensure correctness. This set covers all 100 categories with explicit facing orientation. The 3D assets utilized for benchmark generation do not overlap with those used for dataset generation. (2) Real Set. It contains two portions. The first portion is 2443 real images collected from Pascal3D+~\cite{xiang_wacv14} covering 10 categories (e.g. aeroplane, bicycle, etc.). Each image corresponds to a QA based on provided labels of object orientation or camera viewpoint. The second portion is 800 real images of humans collected from the MEBOW dataset~\cite{wu2020mebow}. Each image corresponds to a QA on human orientation based on provided labels. In total, there are 3243 VQAs in the Real set. We resample the real images to keep a uniform distribution for each orientation and viewpoint type. Since no camera-shot information is provided in the real image set, the QAs in the real set exclude camera-shot type questions.

Both the \textit{Ultimate3D} dataset and the benchmark have no camera-object relation bias, which is very uncommon in the real image datasets (e.g. most images capture objects facing straight to the camera~\cite{xiang_wacv14}). In contrast, in our dataset the number of images in each orientation, viewpoint, and camera-shot type are \textbf{evenly} distributed.

\subsection{Advancing 3D-Aware MLLMs}
We explore how much our synthetic \textit{Ultimate3D} dataset can improve the MLLMs recognition capability on camera-object relation. The evaluation results on both synthetic and real image VQA benchmarks indicate the fine-tuned baseline models are significantly improved from random guess and outperform commercial SOTAs.

\red{
\noindent\textbf{Training Details.}
We fine-tune and evaluate several recent open-source MLLM families, including LLaVA-1.5/1.6~\cite{liu2024improved}, Llama-3.2-Vision-11B~\cite{llamavision2024}, Gemma-4 models~\cite{gemma4}, and Qwen-based models~\cite{bai2025qwen3, yang2025qwen3}. \revision{All models are fine-tuned for one epoch on the \textit{Ultimate3D} training set.} \revision{To evaluate the effect of training on Ultimate3D, we compare each base model with its corresponding fine-tuned version.} We adopt two fine-tuning protocols depending on the model family. For LLaVA and Llama-3.2-Vision, we follow their official full fine-tuning recipes~\cite{liu2024improved, llamavision2024}, updating the MLP vision-language connector and the LLM decoder while freezing the vision encoder. \revision{For LLaVA, we use BF16 precision, a learning rate of $2\times10^{-5}$, cosine scheduling with a warm-up ratio of 0.03, zero weight decay, and gradient checkpointing. For Llama-3.2-Vision-11B, we use BF16 precision, fused AdamW, a learning rate of $2\times10^{-5}$, a per-device batch size of 2 with gradient accumulation of 4, and gradient clipping at 1.0.} For Gemma-4 and Qwen-based models, we use parameter-efficient fine-tuning primarily based on LoRA. \revision{For the Qwen3.5 and Qwen3-VL series, we use Unsloth-based parameter-efficient fine-tuning with a hybrid trainable-parameter policy: the vision encoder is kept frozen, the language-model attention and MLP projections are adapted with LoRA (rank 16, $\alpha$=16, dropout 0, no bias terms, no RSLoRA), and the vision–language connector (merger) is optimized as full weights. For Gemma-4 models, we use plain LoRA with rank 16, $\alpha=32$, and dropout 0.05, applied to compatible linear layers in the language model and vision embedding projection. Gemma-4 models are trained in BF16 using fused AdamW, an effective batch size of 4, and a constant learning rate of $1\times10^{-4}$ without warmup or weight decay. For Gemma-4-31B, we resumed from the best intermediate checkpoint with a learning rate of $3\times10^{-5}$, cosine decay, and weight decay of 0.01 due to late-stage instability.} Training takes approximately 12 hours on 4 H100 GPUs for LLaVA and Llama-3.2-Vision, up to 2.5 days on a single H100 GPU for Gemma-4, and up to 8 days on a single RTX 6000 GPU for Qwen-based models.
}



\noindent\textbf{Evaluation Settings.}
We evaluate the baseline MLLMs and their fine-tuned versions on \textit{Ultimate3D} benchmark and MMVP benchmark~\cite{tong2024eyes}. MMVP benchmark is an intentionally selected real image dataset which is claimed to be difficultly distinguished by MLLMs. We manually label 89 VQAs regarding camera-object relations. During evaluation, MLLMs' textual responses are evaluated by GPT-4o, which is prompted to judge if the responses correspond to the ground truth answers. We calculate the accuracy rate ($\%$) of the MLLMs responses as the evaluation metric. We also include commercial SOTA models (e.g. GPT-4o and Claude-3.5-Sonnet) in the evaluation.

\begin{table*}[!htbp]
\caption{\textbf{Quantitative Comparisons.}
We evaluate open-source and closed-source MLLMs on the \textit{Ultimate3D} and \textit{MMVP} benchmarks. LLaVA and Llama models are fine-tuned using their original full fine-tuning recipes, while Gemma-4 and Qwen models are fine-tuned with LoRA-based parameter-efficient fine-tuning (for Qwen, the vision-language merger is also fully trained). Results report MLLM response accuracy ($\%$) for object orientation, camera viewpoint, and camera-shot recognition. Superscripts indicate the fine-tuning setup: \textsuperscript{F} full fine-tuning and \textsuperscript{L} LoRA fine-tuning.}

\vspace{-1mm}
\centering
\fontsize{7.6pt}{9pt}\selectfont
\setlength{\tabcolsep}{3.0pt}
\renewcommand{\arraystretch}{1.2}

\begin{tabular}{l|cccc|cccc|cc}
\noalign{\hrule height 1.0pt}
\multirow{3}{*}{\textbf{MLLMs}} 
& \multicolumn{4}{c}{\textbf{Orientation} $(\%\uparrow)$} 
& \multicolumn{4}{|c|}{\textbf{Viewpoint} $(\%\uparrow)$} 
& \multicolumn{2}{c}{\textbf{Camera-Shots} $(\%\uparrow)$} \\

\cmidrule(lr){2-11}
& \multicolumn{3}{c|}{\textit{Ultimate3D}} 
& \multirow{2}{*}{\textit{MMVP}} 
& \multicolumn{3}{c|}{\textit{Ultimate3D}} 
& \multirow{2}{*}{\textit{MMVP}} 
& \multicolumn{1}{c|}{\textit{Ultimate3D}} 
& \multirow{2}{*}{\textit{MMVP}} \\

\cmidrule(lr){2-4} \cmidrule(lr){6-8} \cmidrule(lr){10-10}
& Both & Syn & \multicolumn{1}{c|}{Real} 
&  
& Both & Syn & \multicolumn{1}{c|}{Real} 
&  
& \multicolumn{1}{c|}{Syn} 
&  \\

\noalign{\hrule height 1.0pt}

\cellcolor{orange!15}LLaVA-1.5-7B 
& 18.3 & 11.1 & 22.5 & 25.0 
& 31.8 & 38.4 & 25.3 & 29.2 
& 46.0 & 66.7 \\

\cellcolor{orange!15}LLaVA-1.6-7B 
& 18.2 & 14.7 & 20.3 & 25.0 
& 36.8 & 43.9 & 29.8 & 33.3 
& 46.6 & 71.4 \\

\cellcolor{orange!15}LLaVA-1.5-13B 
& 17.9 & 18.7 & 17.4 & 22.7 
& 38.9 & 35.0 & 42.8 & 41.7 
& 44.6 & 61.0 \\

\cellcolor{orange!15}LLaVA-1.6-13B 
& 16.1 & 16.4 & 16.0 & 15.9 
& 30.7 & 30.3 & 31.1 & 33.3 
& 42.3 & 61.9 \\

\cellcolor{orange!15}Llama-3.2-V-11B 
& 5.6 & 5.3 & 5.7 & 6.8 
& 19.8 & 26.2 & 13.6 & 38.1 
& 29.6 & 25.0 \\

\noalign{\hrule height 1.0pt}

\cellcolor{green!12}\textbf{Finetuned LLaVA-1.5-7B}\textsuperscript{F} 
& 68.8 & 85.6 & 58.8 & 54.5 
& 70.6 & 81.3 & 60.0 & 66.7 
& 94.0 & 66.7 \\

\cellcolor{green!12}\textbf{Finetuned LLaVA-1.6-7B}\textsuperscript{F} 
& 71.5 & 86.9 & 62.4 & 61.4 
& 71.3 & 83.3 & 59.5 & 66.7 
& 94.1 & 50.0 \\

\cellcolor{green!12}\textbf{Finetuned LLaVA-1.5-13B}\textsuperscript{F} 
& 70.0 & 85.6 & 60.8 & 50.0 
& 69.7 & 81.9 & 57.7 & 66.7 
& 94.2 & 76.2 \\

\cellcolor{green!12}\textbf{Finetuned LLaVA-1.6-13B}\textsuperscript{F} 
& 72.4 & 88.1 & 63.1 & 65.9 
& 72.3 & 83.0 & 61.8 & 75.0 
& 94.8 & 71.4 \\

\cellcolor{green!12}\textbf{Finetuned Llama-3.2-V-11B}\textsuperscript{F} 
& 74.2 & 87.3 & 66.4 & 50.0 
& 69.1 & 79.6 & 58.7 & 58.3 
& 93.1 & 76.2 \\

\noalign{\hrule height 1.0pt}

\cellcolor{orange!15}Qwen3.5-2B 
& 28.0 & 36.1 & 20.2 & 34.1 
& 40.6 & 37.1 & 44.1 & 50.0 
& 45.2 & 42.9 \\

\cellcolor{orange!15}Qwen3.5-4B 
& 37.0 & 42.8 & 31.3 & 45.5 
& 45.7 & 64.3 & 27.3 & 50.0 
& 45.5 & 47.6 \\

\cellcolor{orange!15}Qwen3.5-9B 
& 39.6 & 48.3 & 31.1 & 40.9 
& 47.1 & 63.7 & 30.8 & 50.0 
& 37.7 & 28.6 \\

\cellcolor{orange!15}Qwen3-VL-4B 
& 21.9 & 28.6 & 15.3 & 45.5 
& 45.7 & 52.1 & 39.4 & 33.3 
& 43.4 & 47.6 \\

\cellcolor{orange!15}Qwen3-VL-8B 
& 34.3 & 41.9 & 26.9 & 52.3 
& 47.1 & 59.7 & 34.6 & 45.8 
& 44.9 & 38.1 \\

\cellcolor{orange!15}Qwen3-VL-32B 
& 41.1 & 46.7 & 35.6 & 56.8 
& 52.5 & 66.4 & 38.8 & 50.0 
& 44.3 & 57.1 \\

\noalign{\hrule height 1.0pt}

\cellcolor{green!12}\textbf{Finetuned Qwen3.5-2B}\textsuperscript{L} 
& 68.0 & 80.4 & 60.2 & 61.4 
& 66.8 & 76.5 & 57.1 & 62.5 
& 91.7 & 70.7 \\

\cellcolor{green!12}\textbf{Finetuned Qwen3.5-4B}\textsuperscript{L} 
& 71.3 & 79.0 & 63.5 & 60.9 
& 71.0 & 79.0 & 63.0 & 66.7 
& 92.1 & 72.8 \\

\cellcolor{green!12}\textbf{Finetuned Qwen3.5-9B}\textsuperscript{L} 
& 73.3 & 82.5 & 64.0 & 63.6 
& 71.8 & 73.5 & 70.2 & 62.8 
& 91.9 & 71.2 \\

\cellcolor{green!12}\textbf{Finetuned Qwen3-VL-4B}\textsuperscript{L} 
& 67.5 & 83.8 & 56.2 & 68.2 
& 72.8 & 81.5 & 59.5 & 75.1 
& 91.5 & 73.3 \\

\cellcolor{green!12}\textbf{Finetuned Qwen3-VL-8B}\textsuperscript{L} 
& 69.5 & 80.8 & 62.5 & 62.7 
& 71.2 & 76.2 & 67.1 & 70.8 
& 89.7 & 77.0 \\

\cellcolor{green!12}\textbf{Finetuned Qwen3-VL-32B}\textsuperscript{L} 
& 73.1 & 83.5 & 65.5 & 69.5 
& 73.4 & 82.3 & 65.1 & 73.3 
& 92.2 & 78.7 \\

\noalign{\hrule height 1.0pt}

\cellcolor{orange!15}Gemma-4-E2B-it 
& 25.6 & 28.2 & 24.0 & 40.9 
& 39.5 & 48.2 & 30.9 & 37.5 
& 37.7 & 47.6 \\

\cellcolor{orange!15}Gemma-4-E4B-it 
& 29.8 & 35.2 & 26.6 & 38.6 
& 44.0 & 58.0 & 30.1 & 58.3 
& 39.2 & 28.6 \\

\cellcolor{orange!15}Gemma-4-26B-A4B-it 
& 47.3 & 54.6 & 43.0 & 56.8 
& 52.6 & 64.2 & 41.1 & 66.7 
& 36.9 & 47.6 \\

\cellcolor{orange!15}Gemma-4-31B-it 
& 44.9 & 55.2 & 38.9 & 52.3 
& 52.8 & 63.7 & 42.1 & 54.2 
& 40.1 & 52.4 \\

\noalign{\hrule height 1.0pt}

\cellcolor{green!12}\textbf{Finetuned Gemma-4-E2B-it}\textsuperscript{L} 
& 67.3 & 80.0 & 59.8 & 61.4 
& 66.7 & 76.8 & 56.7 & 62.5 
& 91.8 & 71.4 \\

\cellcolor{green!12}\textbf{Finetuned Gemma-4-E4B-it}\textsuperscript{L} 
& 65.7 & 78.6 & 58.2 & 59.1 
& 69.4 & 76.2 & 62.8 & 58.3 
& 89.8 & 76.2 \\

\cellcolor{green!12}\textbf{Finetuned Gemma-4-26B-A4B-it}\textsuperscript{L} 
& 71.9 & 83.2 & 65.3 & 61.4 
& 67.7 & 76.8 & 58.6 & 58.3 
& 92.4 & 71.4 \\

\cellcolor{green!12}\textbf{Finetuned Gemma-4-31B-it}\textsuperscript{L} 
& 73.9 & 86.8 & 66.3 & 45.5 
& 69.8 & 80.1 & 59.7 & 62.5 
& 92.2 & 76.2 \\

\noalign{\hrule height 1.0pt}

\cellcolor{cyan!10}GPT-4o 
& 43.1 & 51.0 & 38.5 & 40.9 
& 54.1 & 63.5 & 44.9 & 70.8 
& 41.7 & 42.9 \\

\cellcolor{cyan!10}GPT-4o-mini 
& 43.5 & 52.3 & 38.3 & 38.6 
& 54.5 & 63.9 & 45.1 & 66.7 
& 41.5 & 42.9 \\

\cellcolor{cyan!10}Claude-3-Sonnet 
& 43.1 & 52.6 & 37.5 & 36.4 
& 54.3 & 64.2 & 44.5 & 66.7 
& 41.7 & 47.6 \\

\cellcolor{cyan!10}Claude-3.5-Sonnet 
& 40.3 & 47.7 & 35.8 & 54.5 
& 55.6 & 65.9 & 45.3 & 66.7 
& 41.8 & 42.9 \\

\cellcolor{cyan!10}Gemini-3.5-Flash \textit{(min. thinking)}
& 56.8 & 66.9 & 50.8 & 61.4 
& 61.4 & 68.4 & 54.4 & 83.3 
& 37.1 & 52.4 \\

\cellcolor{cyan!10}Gemini-3.5-Flash \textit{(high thinking)}
& 65.3 & 73.8 & 60.3 & 56.8 
& 61.6 & 67.0 & 56.3 & 79.2 
& 33.8 & 42.9 \\

\cellcolor{cyan!10}Claude-Sonnet-4.6 
& 36.4 & 44.1 & 28.9 & 50.0 
& 56.4 & 67.1 & 45.8 & 66.7 
& 40.9 & 47.6 \\

\cellcolor{cyan!10}GPT-5.5 
& 54.1 & 62.5 & 45.9 & 70.5 
& 59.2 & 69.7 & 48.9 & 58.3 
& 37.3 & 52.4 \\

\cellcolor{cyan!10}Claude-Opus-4.7 
& 46.0 & 52.0 & 40.4 & 45.5 
& 57.2 & 67.1 & 47.5 & 62.5 
& 44.3 & 52.4 \\

\noalign{\hrule height 1.0pt}
\end{tabular}
\vspace{-8pt}
\label{tbl:quan_ulti3d}
\end{table*}

\red{
\noindent\textbf{Quantitative Results.}
The quantitative results of response accuracy ($\%$) are shown in Tab.~\ref{tbl:quan_ulti3d}. \revision{We assess the effect of training on \textit{Ultimate3D} through base-to-fine-tuned comparisons within each model family.} Overall, we observe a consistent trend: off-the-shelf MLLMs struggle with camera-object relation recognition, whereas task-specific fine-tuning on \textit{Ultimate3D} substantially improves performance across multiple model families. This difficulty appears even for strong general-purpose vision-language models, including LLaVA, Llama-3.2-Vision, Gemma-4, Qwen-based models, and closed commercial MLLMs. Several baseline models tend to collapse to a small set of frequent answers, such as predicting ``front'' orientation or ``top'' viewpoint regardless of the actual image content. As a result, their accuracies are often close to random-guess levels, e.g., $33.3\%$ for three-way viewpoint recognition and $12.5\%$ for eight-way orientation recognition. This weakness may be related to biases in commonly used visual instruction datasets, where most images are captured from canonical front-facing directions and from common side or above viewpoints. The poor off-the-shelf performance across both older and more recent MLLM families suggests that camera-object relation understanding remains underrepresented in existing MLLM training data.

Fine-tuning on our synthetic \textit{Ultimate3D} dataset substantially improves camera-object relation recognition across all evaluated open-source model families. On the \textit{Ultimate3D} benchmark, fine-tuned LLaVA models significantly outperform their corresponding base models across object orientation, camera viewpoint, and camera-shot recognition. For example, fine-tuned LLaVA-1.6-13B improves orientation accuracy from $16.1\%$ to $72.4\%$, viewpoint accuracy from $30.7\%$ to $72.3\%$, and camera-shot accuracy from $42.3\%$ to $94.8\%$. The smaller LLaVA-1.6-7B model achieves similarly competitive performance, reaching $71.5\%$, $71.3\%$, and $94.1\%$ on these three tasks, respectively. This suggests that the ability to distinguish camera-object relations does not necessarily require a very large backbone, but instead benefits strongly from task-specific supervision.

The same conclusion holds for Llama-3.2-Vision, Gemma-4, and Qwen-based models. Llama-3.2-Vision-11B obtains the largest average improvement among the models in Tab.~\ref{tbl:quan_ulti3d}, with an average gain of $60.46$ percentage points across the three camera-object relation tasks. The original released model often refuses to answer some recognition questions, e.g., responding that it cannot provide information about the individual in the image, while the fine-tuned model produces valid relation predictions and achieves the best orientation accuracy in Tab.~\ref{tbl:quan_ulti3d}. These results indicate that \textit{Ultimate3D} supervision can correct both answer-format failures and visual relation recognition failures.

Recent Gemma-4 and Qwen-based models further confirm that the benefit of \textit{Ultimate3D} supervision is not specific to the LLaVA architecture. Before task-specific fine-tuning, Gemma-4 models achieve only $36.0$--$51.1\%$ overall accuracy when averaging all reported relation-recognition columns. After LoRA fine-tuning, they reach $69.4$--$71.3\%$ overall accuracy, corresponding to an average improvement of $26.3$ percentage points. The improvements are consistent across all three relation types, with average gains of $20.0$--$37.5$ percentage points on object-direction recognition, $9.2$--$26.7$ percentage points on viewpoint recognition, and $38.0$--$49.1$ percentage points on camera-shot recognition.

Qwen-based models show a similar pattern. Their base versions achieve $37.3$--$48.9\%$ overall accuracy, while the fine-tuned versions reach $69.5$--$75.7\%$ overall accuracy, improving by $30.5$ percentage points on average. In particular, fine-tuned Qwen3-VL-32B achieves the strongest overall performance among the evaluated open models in Tab.~\ref{tbl:quan_ulti3d}, reaching $75.7\%$ average accuracy across all reported columns. It also obtains strong performance on both synthetic and real subsets, suggesting that the learned camera-object relation knowledge is not restricted to rendered synthetic images.

Among closed commercial MLLMs, newer models generally perform better than earlier GPT-4o- and Claude-3-era models on some orientation and viewpoint columns, but the improvement is not uniform across tasks. For example, Gemini-3.5-Flash with the high thinking setting achieves stronger orientation accuracy than its minimal-thinking counterpart, improving from $56.8\%$ to $65.3\%$ on the \textit{Ultimate3D} orientation column. However, increasing the thinking budget does not consistently improve all relation types: its camera-shot accuracy decreases from $37.1\%$ to $33.8\%$. This suggests that additional reasoning budget can help in some spatial-relation cases, but it does not resolve the lack of grounded camera-object relation understanding.

Compared with closed commercial MLLMs, our fine-tuned open models achieve especially strong performance on the \textit{Ultimate3D} relation-recognition tasks. Although recent commercial models improve over earlier commercial baselines on some orientation and viewpoint columns, they remain substantially weaker on several \textit{Ultimate3D} tasks, especially camera-shot recognition, where fine-tuned open models consistently reach around $90\%$ accuracy or higher. Overall, these results demonstrate that \textit{Ultimate3D} provides effective supervision for camera-object relation understanding, improves both synthetic and real-image evaluation performance, and generalizes across multiple MLLM families, including LLaVA, Llama, Gemma, and Qwen.
}

\noindent\textbf{Qualitative Results.} 

We show qualitative comparisons in Fig.~\ref{fig:qualitatiev}. Commercial SOTA models (e.g. \red{Gemini-3.5-Flash}) and open-source models (e.g. \red{Gemma-4-31B-it}) struggle with precise recognition of all three types of camera-object relations. \red{For the two commercial models, their object-orientation predictions are often unreliable despite producing plausible natural-language explanations. We also observe systematic biases in viewpoint and camera-shot recognition: both commercial models tend to underestimate viewpoint angle, e.g., mistaking top views as horizontal, and overestimate camera-shot distance, e.g., mistaking medium-shot images as long-shot.} The original \red{Gemma-4-31B-it} model also shows strong response biases across all types of questions. \red{Its orientation predictions are biased mostly toward ``Left'' and ``Right'', its viewpoint predictions are dominated by ``Horizontal'', and its camera-shot predictions are biased toward ``Long-shot''.} In comparison, the \red{Gemma-4-31B-it} model fine-tuned on our \textit{Ultimate3D} dataset provides more plausible predictions across all three tasks on both the synthetic and real subsets of our benchmark. Additional comparisons are provided in Supplemental Sec. ~\ref{suppsec:addqual}.

We hypothesize that \red{these behavioral biases arise from dataset bias and the lack of camera-object visual instruction data in current MLLM training pipelines.} Without \red{modifying the model architecture or training scheme}, our fine-tuned \red{Gemma-4-31B-it} model \red{substantially improves camera-object relation recognition and outperforms strong commercial models on several \textit{Ultimate3D} tasks, despite relying only on task-specific supervision. This indicates that the lack of targeted camera-object relation data is a major factor behind current MLLMs' weakness on camera-object relation recognition.}

\noindent\textbf{Limitations.} 
The numerical prediction of camera-object relation $\beta$ is more challenging for MLLMs compared to the categorical prediction we report. Preliminary tests show MLLMs can handle numerical prediction of camera-object relation for a single category, but struggle with a hundred categories as in our dataset. Since our main contribution is the discovery of the dataset bottleneck and the generation pipeline, we leave addressing this limitation as future work.

\subsection{Generalization of \textit{Ultimate3D} to General VQA}
To explore the generalization of 3D camera-object relation capturing with general VQA tasks, we fine-tune LLaVA-1.6-13B models by a half-half mixture of \textit{Ultimate3D} (240K VQAs) and randomly sampled 240K VQAs from LLaVA-Instruct-665K~\cite{liu2024improved}. Tab.~\ref{tab:comparison_vqav2} presents the dataset mixture finetuning performance on VQAv2~\cite{goyal2017making}, our \textit{Ultimate3D}, and \textit{MMVP} benchmarks. The results demonstrate that incorporating \textit{Ultimate3D} significantly enhances camera-object relation reasoning while maintaining general VQA performance.

\begin{table}[t]
\caption{\textbf{Generalization.} We finetune LLaVA-1.6-13B model using a mixture of both general VQA and camera-object relation VQA datasets. Without degradation on general VQA tasks, finetuned model improves its ability on camera-object relation recognition and beats SOTA (i.e. GPT-4o) by 28.42\% accuracy on \textit{Ultimate3D} benchmark. $*$ denotes the average accuracy ($\%$) across all 3 types of camera-object relation VQAs in Tab.~\ref{tbl:quan_ulti3d}.}
\vspace{-4mm}
\centering
\footnotesize
\renewcommand{\arraystretch}{1.2}

\begin{tabular*}{\columnwidth}{@{}p{0.36\columnwidth}@{\extracolsep{\fill}}ccc@{}}
\toprule
MLLMs & VQAv2~\cite{goyal2017making} & \textit{Ultimate3D}\textsuperscript{*} & \textit{MMVP}\textsuperscript{*} \\
\midrule
\cellcolor{orange!15}LLaVA-1.6-13B~\cite{liu2024improved} & \underline{80.00} & 29.70 & 37.03 \\
\midrule
\cellcolor{green!12}\textit{Ulti3D} (100\%) & 78.77 & \textbf{79.83} & \textbf{70.77} \\
\cellcolor{green!12}\textit{Ulti3D} (50\%) + ~\cite{liu2024improved} (50\%) & \textbf{80.01} & \underline{74.72} & \underline{65.80} \\
\midrule
\cellcolor{cyan!10}GPT-4o & / & 46.30 & 51.57 \\
\cellcolor{cyan!10}Claude-3.5-Sonnet & / & 45.90 & 54.70 \\
\botrule
\end{tabular*}

\vspace{-4mm}
\label{tab:comparison_vqav2}
\end{table}

\begin{figure}[!htbp]
  \includegraphics[width=\linewidth]{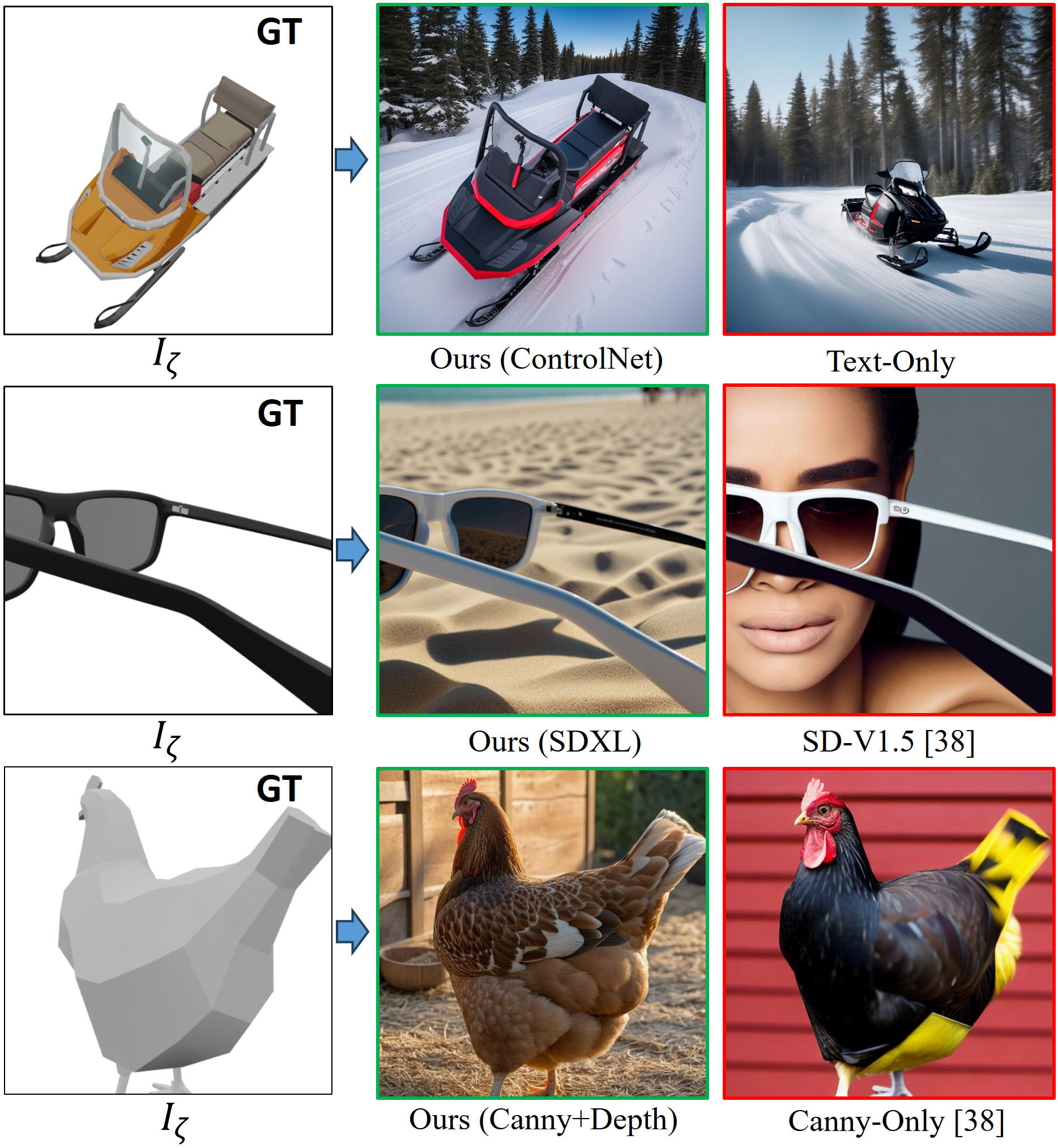}
  \vspace{-6mm}
  \caption{\textbf{Ablations on Image Generation Pipeline.}
  Our framework shows superior performance by using multi-ControlNet with SDXL backbones, and introducing both Canny edges and depth maps as visual priors. It consistently delivers better robustness and quality compared to text-only SDXL and~\cite{ma2023generating}. When these visual priors are removed (top-right: removing all visual priors; bottom-right: removing depth prior), the generator often fails to produce good results and brings lower generation success rate (see Sec~\ref{sec:dataset_quality},~\cite{ma2023generating} only achieves $55.11\%$ success rate v.s. ours $93.07\%$)
  }
  \vspace{-4mm}
  \label{fig:ablation}
\end{figure}

\subsection{Ablations on Image Generation Pipeline}
\label{sec:ablation}
In Fig.~\ref{fig:ablation}, we ablate our image generation pipeline and illustrate better quality than~\cite{ma2023generating} by: (1) removing the visual priors; and (2) using alternative image generation backbones; (3) changing the group of visual priors given to the image generator. Compared to text-only SDXL and~\cite{ma2023generating}, we find that multiple visual priors may significantly improve the robustness of image generation under complex camera-object relation (especially for viewpoints of the object's back-side, and for 3D assets with low level-of-detail). In particular, the depth and Canny edge prior are crucial. However, we find that adding more priors (e.g., RGB, semantic masks, normal maps) does not explicitly improve quality. Moreover, our SDXL backbone outperforms other alternatives (i.e. SD-V1.5~\cite{rombach2022high}). Discussions in Appendices~\ref{suppsec:dm_backbone_ablation} and~\ref{suppsec:genfailure}.

\subsection{User Study on Dataset Quality}
\label{sec:dataset_quality}
We investigate using text-image aligning metric and pose-estimation model prediction to evaluate quality of our Ultimate3D dataset but find those metrics to be deficient (Appendix~\ref{supp_sec: aligning}). Instead, we perform a comprehensive user study as a quality check of \textit{Ultimate3D} dataset. As Appendix~\ref{suppsec:userstudy} shows, the user will see two images side-by-side (Fig.~\ref{suppfig:userstudy}): an RGB image rendered by Blender (representing $I_{\beta}$), and a synthetic image generated by our method using $I_{\beta}$. Preserving the RGB image's 3D geometry and structure in the synthetic image is regarded as success. Our study includes 225 randomly sampled image pairs where each image is independently reviewed by 3 users. Overall success rate is \textbf{93.07$\%$} (agreement rate among 3 users is 94.4\%). In comparison, we perform another similar-objective user study on ~\cite{ma2023generating}. The overall success of this alternative method was only 55.11\% (we explain the cause of this reduced performance in Sec.~\ref{sec:ablation}).

\section{Conclusion}
\label{sec:conclusion}
In this work, we discover the dataset bottleneck of MLLMs for distinguishing camera-object relations. Given only 3D assets, our framework is able to generate unlimited image-text pairs of diverse camera-object relations with no distribution bias. We also provided \textit{Ultimate3D} dataset and benchmark MLLM community. Moreover, fine-tuned open-sourced MLLMs using our dataset outperform SOTA commercial models by a significant margin. We plan to extend our framework for multi-object image generation with additional viewpoints, and improve numerical prediction capability on camera-object relation.

\section{Acknowledgments}
This work was performed during Liu He's internship at Amazon, and later extended by Ruiqi Xu and Baris Tezcan at Purdue University. Ruiqi Xu fine-tuned Qwen3 series and evaluated GPT5.5 \& Claude models, and Baris Tezcan fine-tuned Gemma-4 series and evaluated Gemini models. We thank Jingjing Zheng, Tony Qi, Nan Qiao, Ke Zhang, and Yuyin Sun for their insightful discussions and valuable contributions to the formulation of the ideas in this work.


\begin{appendices}
\titleformat{\section}
  {\normalfont\normalsize\bfseries}
  {\thesection}{1em}{}

\titlespacing*{\section}
  {0pt}{1.0ex plus .2ex}{0.6ex}
\clearpage

\vspace{3em}
\noindent \textbf{\Large Appendix Overview}

Below is a summary of the contents in each section of this supplemental material:
\begin{itemize}
\item Sec.~\ref{suppsec:textprompt}: Details of text prompt generation as image description utilized for controlled image generation.
\item Sec.~\ref{suppsec:vqagen}: Details of VQA prompt generation given the ground truth camera-object relations. 
\item Sec.~\ref{suppsec:LLMgrading}: Details of LLM-based grading for MLLM response evaluation.
\item Sec.~\ref{supp_sec: aligning}: Preliminary studies on data curating for camera-object relation VQA dataset by text-image aligning metrics, and by pose estimation models.
\item Sec.~\ref{suppsec:dm_backbone_ablation}: Ablations on selections of DM backbones. 
\item Sec.~\ref{suppsec:genfailure}: Factors influence generation quality of our image generation pipelines.
\item Sec.~\ref{suppsec:userstudy}: Details of user study on image quality evaluation.
\item Sec.~\ref{suppsec:addqual}: More qualitative comparisons between finetuned LLaVA and Gemma models to commerical SOTA models.
\item Sec.~\ref{suppsec:ultimate3D}: 48 image examples illustrate the diversity of \textit{Ultimate3D} dataset and benchmarks on object categories, camera-object relations, and background context.
\item Sec.~\ref{app:imagenet3d}: Generalization to real images: evaluation of base and \textit{Ultimate3D}-fine-tuned models on the independently collected ImageNet3D benchmark.
\item Sec.~\ref{app:failure}: Failure analysis of fine-tuned models: residual weak relations, error structure, boundary effects, and annotated image conditions.
\item Sec.~\ref{suppsec:finetune_dicussion}: Additional discussions and insights on using synthetic dataset generated by our framework for MLLM finetuning.
\item Sec.~\ref{suppsec:algorithm}: Algorithm of our 3D visual instruction dataset generation pipeline.
\end{itemize}

\section{Image Description Generation}
\label{suppsec:textprompt}
We provide system prompts given to GPT-4oto generate context description for image generation. Corresponding section in main paper is Sec.~\ref{sec:text_gen}. In the prompt, we provide one-shot example for better robustness of the text generation.

Specifically, generated text prompt will be merged with default positive prompt like \textit{"detailed, 4K, 35mm photograph, professional"}, and customized camera-object prompt \textit{"the image shows a [$\beta$] view of a [$c$]"}, in order to enhance image generation quality.

\begin{lstlisting}
    SYSTEM_PROMPT_FOR_IMAGE_GEN = "You are an expert in generating concise and diverse descriptions of an object to guide image generation model like DAll-E3. You should use your common sense to generate 1 sentence description of the given object. Additionally, you should also generate 1 sentence of a common real-world scene given the object. Both sentences of descriptions should not include more than one of the given object to avoid ambiguity.
    
    Example Input: Please generate the visual prompt of chicken.
    Example Output: A white and brown chicken standing in a cage made of metal bars. The chicken has a long, curved beak and its feathers are fluffy and white.
    "
\end{lstlisting}

\section{Text Instruction Generation}
\label{suppsec:vqagen}
We provide the system prompt to generate QA pairs given camera-object relation as below. Corresponding section in main paper is Sec.~\ref{sec:vqagen}. In the prompt, we provide few-shot example for better instructions for diversity of the generated text.

\begin{lstlisting}
    SYSTEM_PROMPT_FOR_VQA_GEN = "You are an AI visual assistant, and you are seeing a single image. What you see are provided with several sentences, describing the same image you are looking at. Answer all questions as you are seeing the image.

    Design a conversation between you and a person asking about this photo. The answers should be in a tone that a visual AI assistant is seeing the image and answering the question.
    Ask diverse questions and give corresponding answers.
    
    Include questions asking about the visual content of the image, including the object orientation and its corresponding azimuth degree, the camera viewpoint of the image and corresponding camera elevation angle, the camera shot type of the image and the distance from camera to the target object. Only include questions that have definite answers:
    (1) one can see the content and its orientation degree in the image that the question asks about and can answer confidently;
    (2) one can recognize camera viewpoint and camera shot of the image that the question asks about and can answer confidently;
    The question can be multiple choice questions given the choice, or open-ended questions. For multiple choice, the option expression can be diverse but with the same meaning. For example, "front" also means "directly towards the camera"; 'back' also means 'away from the camera', etc. Use the common sense to make options diverse if possible.
    Provide at most 5 pairs of question and answer pairs. Prior the confident questions. Do not ask any question that cannot be answered confidently.
    
    You should ask questions from multiple QA templates as the below examples show.
    
    Example Input Description:
    The image shows a front view of police van. A blue and white police van with the word "POLICE" emblazoned on the side is parked on a city street. Nearby, a couple of officers are discussing a recent incident while pedestrians walk by, some glancing curiously at the vehicle.
    The police van is with an azimuth of 181.518 degree, facing Front direction.
    The elevation angle of the camera to the police van is 81.41 degree, the camera viewpoint is Horizontal view.
    The relative distance from the camera to the police van is 1.132 meters, the camera shot type is Close-up.
    
    Example Output Question and Answer Pairs:
    Question:
    From the camera's perspective, is the police van in the picture facing straight or oriented at an angle? Options: (a) Directly towards the camera (b) At an angle
    ===
    Answer:
    (a) Directly towards the camera
    ===
    Question:
    Is the police van in the picture facing the camera or away from the camera? Options: (a) Away from the camera (b) Facing the camera
    ===
    Answer:
    (b) Facing the camera
    ===
    Question:
    Which direction is the police van facing in the image? Options: (1) Back (2) Front
    ===
    Answer:
    (2) Front
    ===
    Question:
    Is the police van facing back or front from the camera's perspective? Options: (a) Back (b) Front
    ===
    Answer:
    (b) Front
    ===
    Question:
    Is the photo taken directly above the police van or from the side? Options: (a) Taken directly (b) From the side
    ===
    Answer:
    (b) From the side
    ===
    Question:
    Is the photo taken far away the police van or taken closely?
    ===
    Answer:
    The relative distance from the camera to the police van is 1.132 meters, thus it is with a close-up shot. This indicates the photo is taken closely.
    ===
    Question:
    What is the elevation viewpoint of the image? Options: (1) Top (2) Horizontal (3) Bottom
    ===
    Answer:
    (2) Horizontal
    ===
    Question:
    Which camera shots is the image? Options: (1) Close-up (2) Medium-shot (3) Long-shot
    ===
    Answer:
    (1) Close-up"
\end{lstlisting}

\section{LLM-based Response Grading}
\label{suppsec:LLMgrading}
We provide the system prompt for evaluation of the MLLM response to our camera-object relation multiple choice questions.

\begin{lstlisting}
SYSTEM_PROMPT_FOR_GRADING_MLLM_RESPONSE ='You are a helpful and precise assistant for checking the quality of the answer. You should review all listed choices and compared to the Answer content, and judge whether the Answer is correct (yes), or not (no). You should only focus on the major content of the answer, not the detailed number or symbol. Please answer in only yes or no'
\end{lstlisting}



\section{Metric Limitations for Evaluating Camera-Object Relation}
\label{supp_sec: aligning}

\begin{figure}[t]
  \centering
  \includegraphics[width=\linewidth]{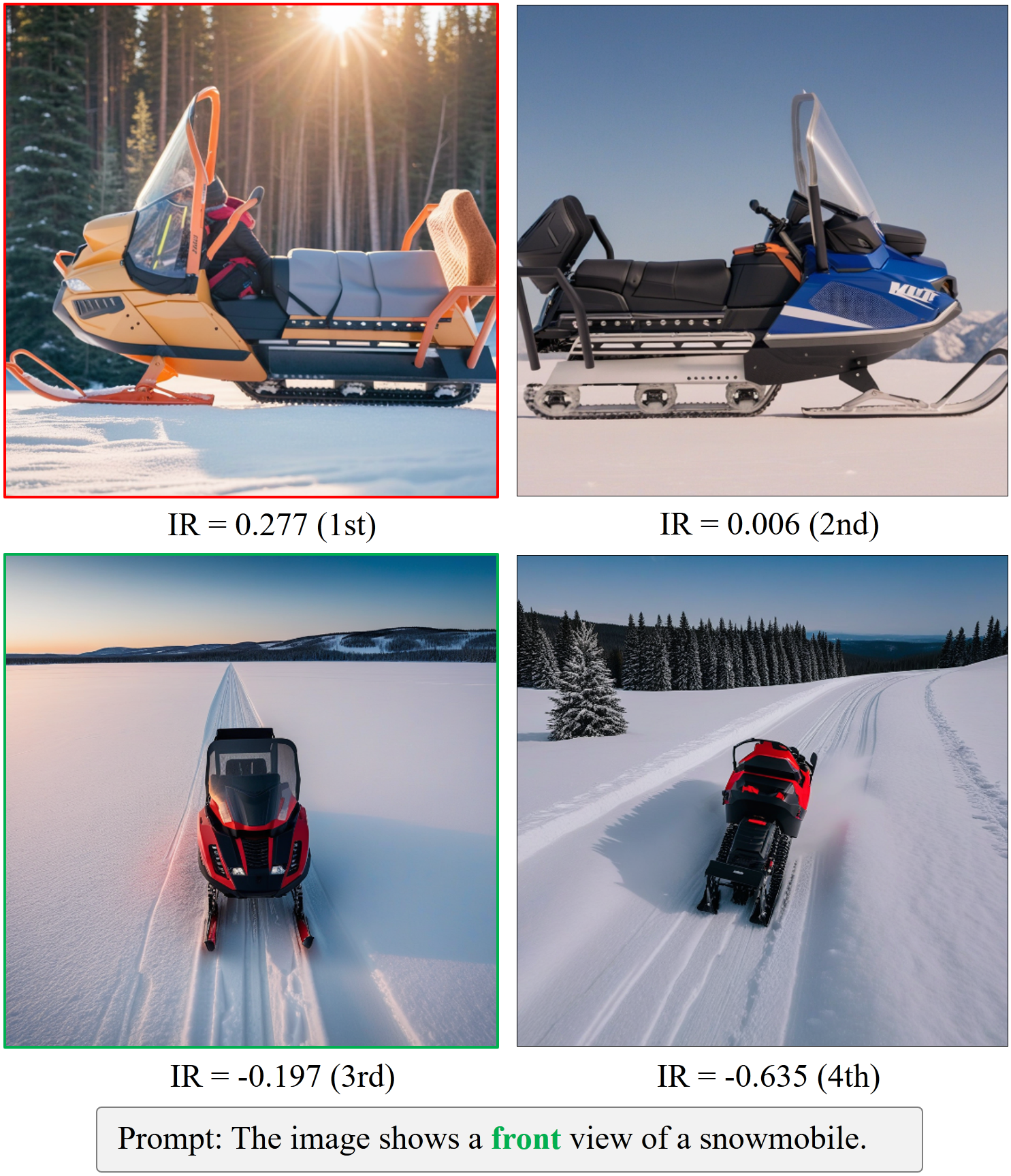}
  \caption{\textbf{Limitations of Text-Image Aligning Metrics.} We evaluate the ImageReward~\cite{xu2024imagereward} (IR) between the text promopt regarding object orientation with above 4 synthetic images. The image matches the correct camera-object relation text prompt ("front") may reach the highest score. However, image with the highest matching score (it faces to "left" direction) is not the correct one with expected orientation. This failure is prevailing across all categories.}
  \label{suppfig:aligning}
\end{figure}

In Fig.~\ref{suppfig:aligning}, we show an example of using ImageReward~\citep{xu2024imagereward} for the dataset curation by evaluating the alignment between generated image and text prompts. Preliminary test indicates that text-image aligning metric is not sensitive to camera-object relation text prompts. Thus we may not rely on those metrics for data curation purpose.

\begin{figure}[t]
  \centering
  \includegraphics[width=\linewidth]{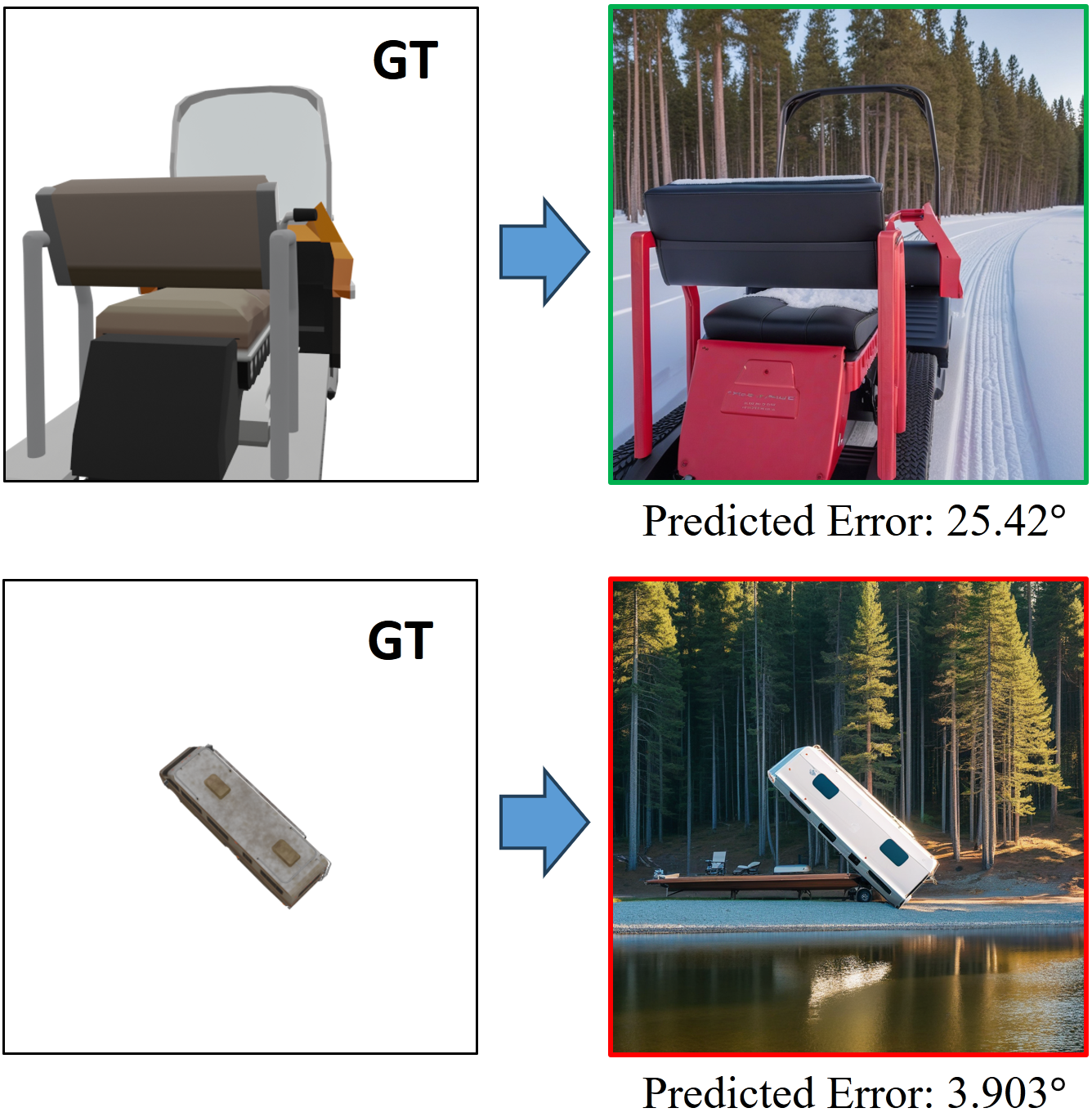}
  \caption{\textbf{Failures of Pose Estimation Model.} Enlighten by~\cite{ma2023generating}, we train a transformer-based pose estimation model on a 10-fold manner for generated images on each single category. The images with less prediction errors may indicate successful generation. However, the pose estimation model is not robust to complex objects and diverse camera-object relations in our dataset.}
  \label{suppfig:poseest}
\end{figure}

Alternatively, in Fig.~\ref{suppfig:poseest}, enlighten by~\cite{ma2023generating}, we train a transformer-based pose estimation model on a 10-fold manner for generated images on each single category. The images with less prediction errors may indicate successful generation. However, the pose estimation model is not robust to complex objects and diverse camera-object relations in our dataset. Using pose estimation model prediction error is not robust for data curation purpose.

\section{Ablation of DM Backbones}
\label{suppsec:dm_backbone_ablation}
In Tab.~\ref{tab:abltion_sd}, we perform quantitative comparisons on general image visual quality between different DM backbones: SD V1.5 and SDXL. We randomly sample 1,000 generated images ($I_{syn}$), each image corresponds to: (1) the RGB visual prior rendered using Blender (one element of $I_{\beta}$); (2) the text prompts ($\mathcal{T}_{img}$) provided to the diffusion model for image generation. We evaluate the generated images on the following aspects:

\begin{itemize}
    \item Prompt following. CLIP-T (CLIP-Text Score \cite{hessel2021clipscore}) and IR (Image Reward \cite{xu2024imagereward}) metrics are employed to assess the alignment between the generated images and the input prompts. The results indicate that images produced by the SDXL backbone exhibit superior prompt alignment.
    \item Fidelity. CLIP-I (CLIP-Image Score \cite{hessel2021clipscore}) and DINO Score \cite{oquab2023dinov2} are used to measure the similarity between the visual prior images and the synthesized outputs. Findings demonstrate that SDXL more effectively preserves fidelity to the visual prior.
    \item Realism. FID \cite{heusel2017gans} is used to compare the distributions of generated images with a subset of LAION-aesthetic \cite{schuhmann2022laion} images, revealing that SDXL significantly surpasses SD V1.5 in terms of image quality and realism.
\end{itemize}

\begin{table}[t]
  \setlength{\tabcolsep}{2pt}
  \footnotesize
  \centering
  \begin{tabular}{l|ccccc}
    \hline
    Ablations & CLIP-T $\uparrow$ & CLIP-I $\uparrow$ & DINO $\uparrow$ & IR $\uparrow$ & FID $\downarrow$ \\ \hline
    Ours (SD-V1.5) & 29.80 & 83.56 & 69.36 & -0.29 & 289.99 \\ 
    Ours (SDXL) & \textbf{34.40} & \textbf{87.44} & \textbf{70.79} & \textbf{0.23} & \textbf{256.94} \\
    \hline
  \end{tabular}
  \caption{\textbf{Ablations on DM Backbones}. SDXL provides better fidelity (CLIP-I, DINO), realism (FID), and text-image alignment (CLIP-T, IR) than other alternatives.}
  \label{tab:abltion_sd}
\end{table}

\section{Factors influencing Image Generation}
\label{suppsec:genfailure}
\begin{figure*}[!t]
  \centering
  \includegraphics[width=\linewidth]{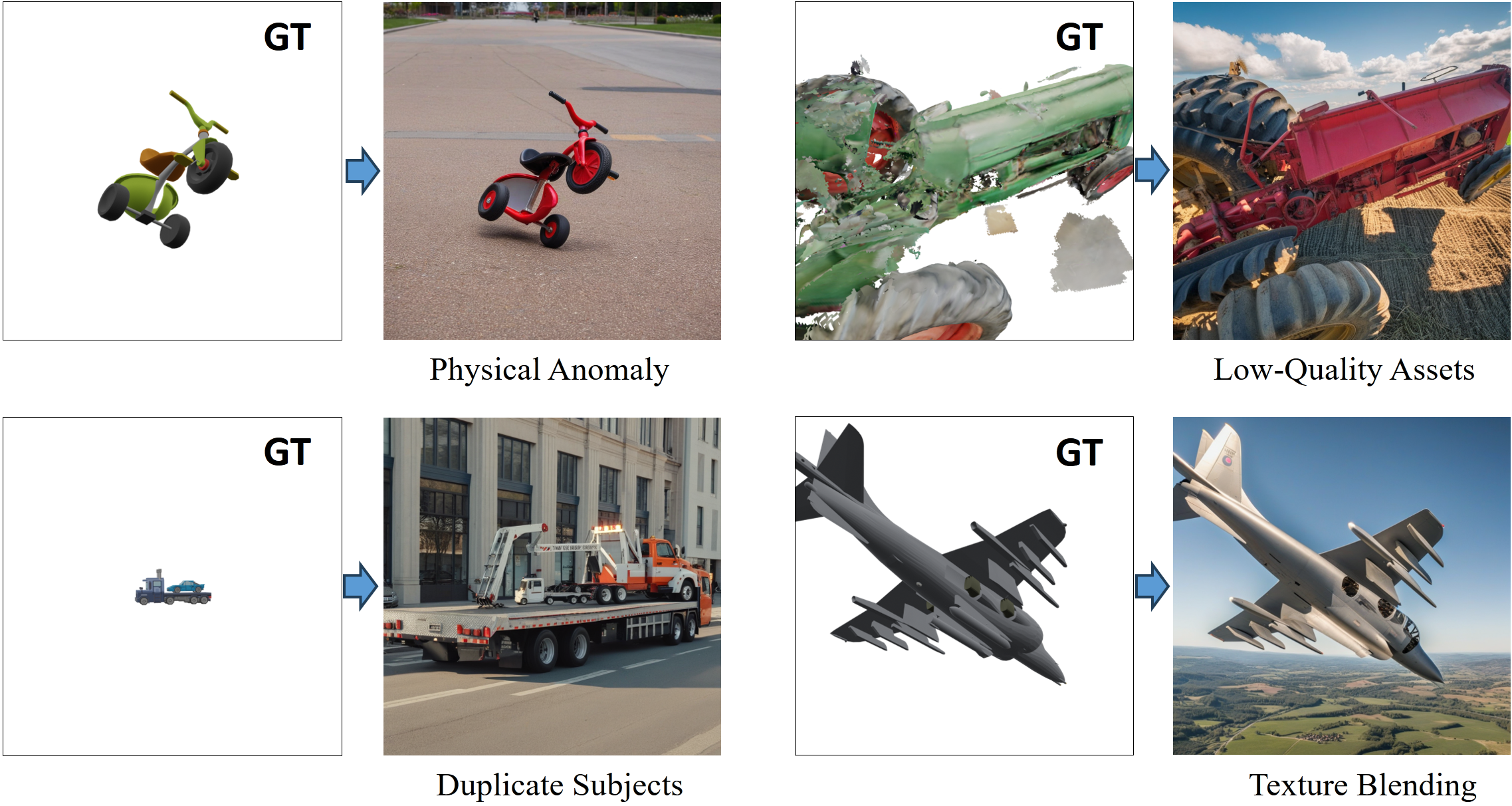}
  \caption{\textbf{Factors influencing Image Generation.} The upper-left example shows that bottom-view camera-object relations can introduce physical anomalies, such as the tricycle appearing with its front lifted unnaturally. The upper-right example highlights that low-quality 3D assets can produce fragmented depth maps or Canny edge visual priors, resulting in unrealistic generated images. The lower-left example shows that small subjects viewed from a long-shot perspective may lead to duplicate subjects in the generated image. The lower-right example illustrates how complex 3D asset structures can cause texture blending issues, particularly where intricate geometry is present.}
  \label{fig:supp_genfailure}
\end{figure*}

Several factors may lead to unreasonable artifacts in our image generation pipeline. In Fig.~\ref{fig:supp_genfailure}, the upper-left example shows that bottom-view camera-object relations can introduce physical anomalies, such as the tricycle appearing with its front lifted unnaturally. The upper-right example highlights that low-quality 3D assets can produce fragmented depth maps or Canny edge visual priors, resulting in unrealistic generated images. The lower-left example shows that small subjects viewed from a long-shot perspective may lead to duplicate subjects in the generated image. The lower-right example illustrates how complex 3D asset structures can cause texture blending issues, particularly where intricate geometry is present.

There are several ways to reduce the influences from those factors. First, choosing 3D assets with reasonable number of meshes, and with plausible textures may filter the low-quality and texture blending failures. Second, using simple and compact image description prompts to generate long-shot camera-shot images will significantly reduce the duplicate subject failures. Reduce the image generation with high elevation angle from bottom view will significantly reduce physical anomaly.

\section{User Study Details}
\label{suppsec:userstudy}

\begin{figure*}[!tp]
  \centering
  \includegraphics[width=\linewidth]{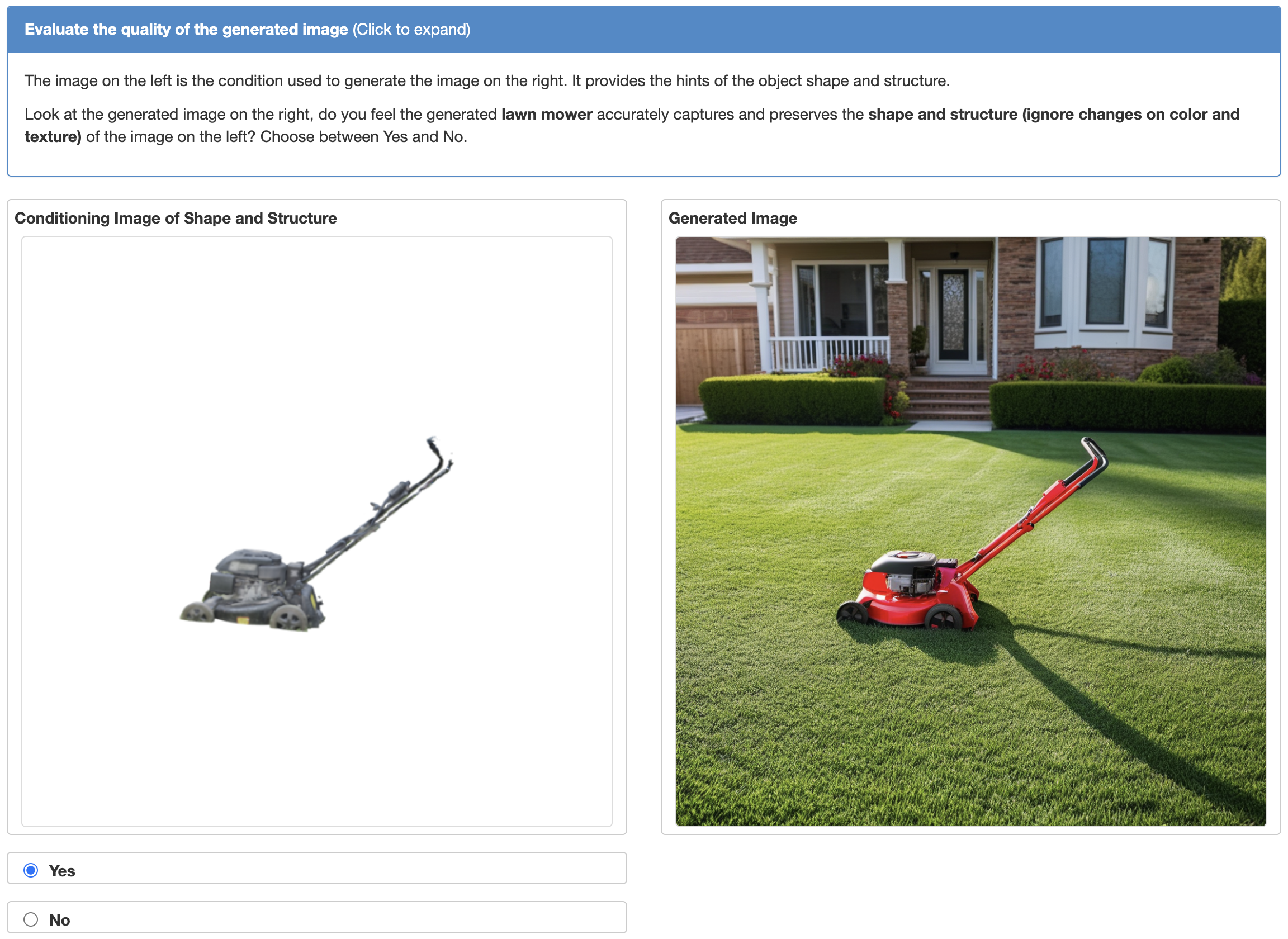}
  \caption{\textbf{User Study UI Page.} Users will see a pair of side-by-side images. The left hand side is an RGB image generated by Blender, and the right hand side is the synthetic image generated by DMs using that RGB image as guidance prior. A vivid preserving of 3D geometry and structure in the synthetic image is a success. Users are required to provide a binary answer regarding the success of the image generation.}
  \label{suppfig:userstudy}
\end{figure*}

The Fig.~\ref{suppfig:userstudy} shows the UI page of our user study. We randomly selected 200 images from the \textit{Ultimate3D} dataset. Each image corresponds to an RGB prior generated by Blender. During the user study, users will see a pair of side-by-side images. One is an RGB image generated by Blender, and the other is the synthetic image generated by DMs using that RGB image as guidance prior. A vivid preserving of 3D geometry and structure in the synthetic image is a success. Users are required to provide a binary answer regarding the success of the image generation. Each image pair will be manually reviewed by at least 3 users.

\section{Additional Qualitative Results}
\label{suppsec:addqual}
\begin{figure*}[!tp]
  \centering
  \includegraphics[width=0.95\linewidth]{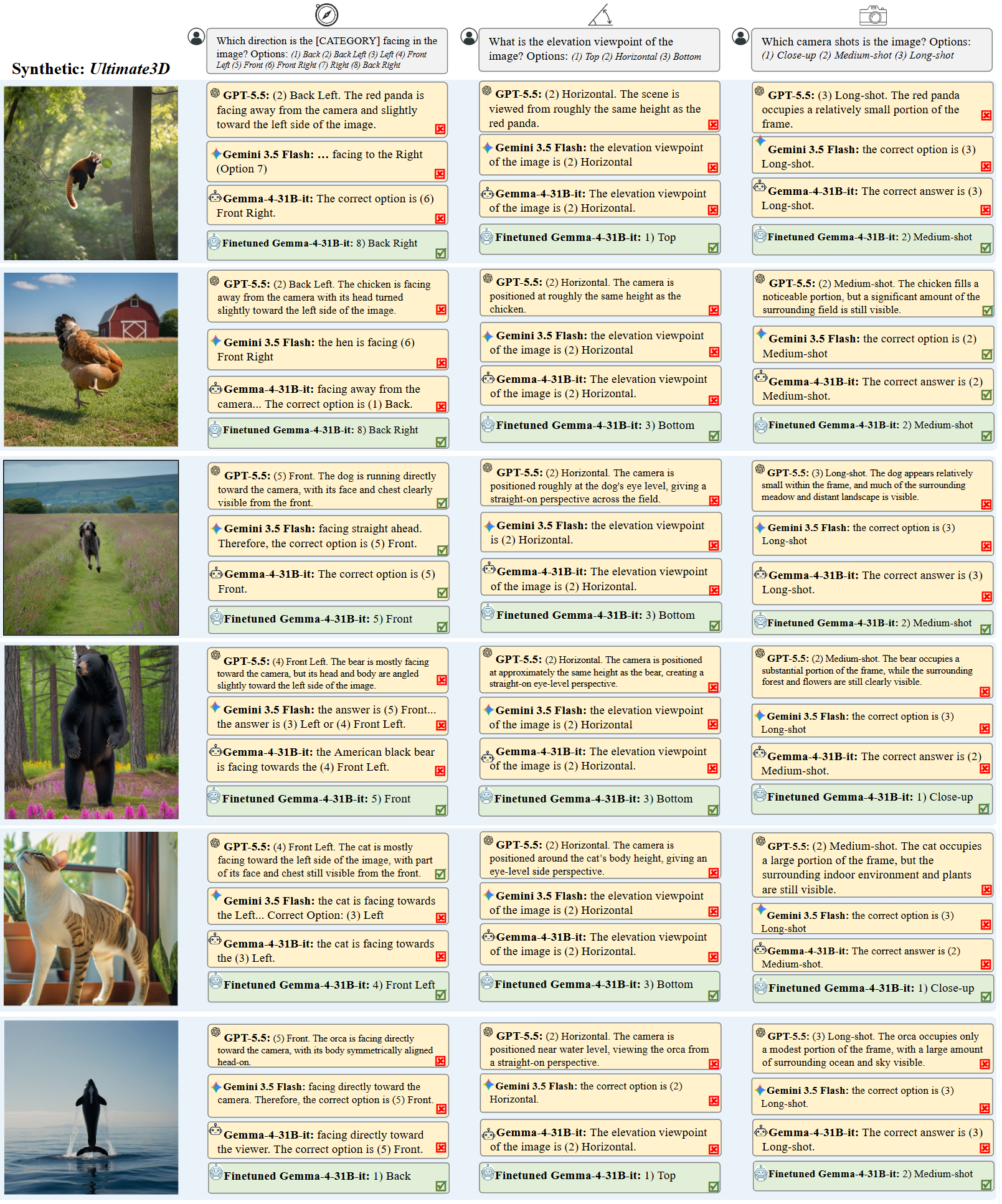}
    \caption{\textbf{Additional Qualitative Results with Gemma.} We evaluate camera-object relation recognition for GPT-5.5, Gemini-3.5-Flash under the high thinking setting, Gemma-4-31B-it, and fine-tuned Gemma-4-31B-it on the \textit{Ultimate3D} benchmark. Each model is asked questions, shown in gray boxes, about object orientation, camera viewpoint, and camera-shot type, together with the input images on the left. The responses show that fine-tuning Gemma-4-31B-it on \textit{Ultimate3D} substantially improves camera-object relation recognition compared with the base Gemma-4-31B-it and closed-source models.}
  \label{fig:supp_qualitatiev}
\end{figure*}

Based on \textit{Ultimate3D} benchmark, we provide additional Fig.~\ref{fig:supp_qualitatiev} to show more qualitative comparisons between finetuned LLaVA model to commercial SOTAs. The finetuned LLaVA model outperforms other models.


\begin{figure*}[!tp]
  \centering
  \includegraphics[width=0.95\linewidth]{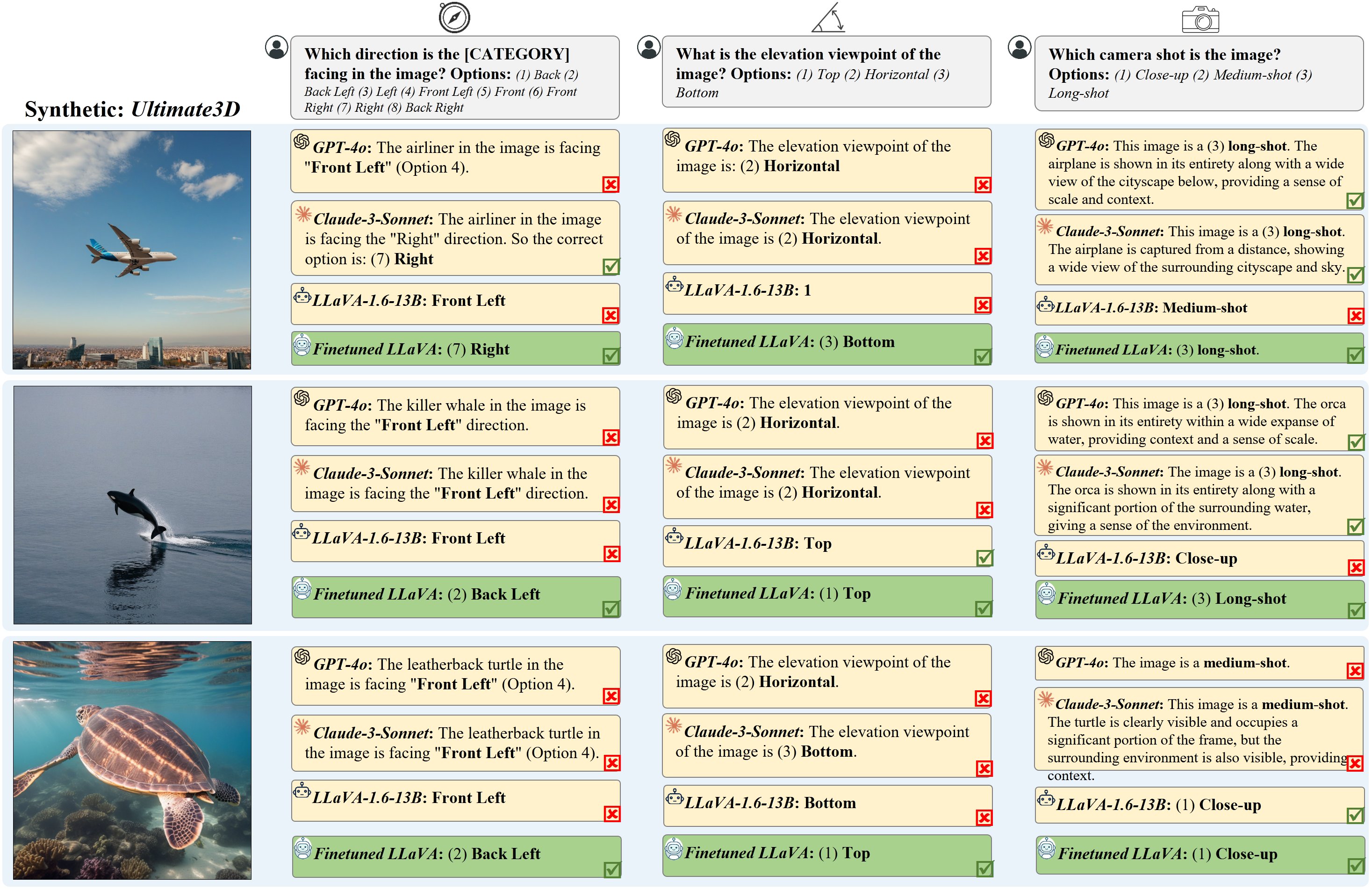}
  \includegraphics[width=0.95\linewidth]{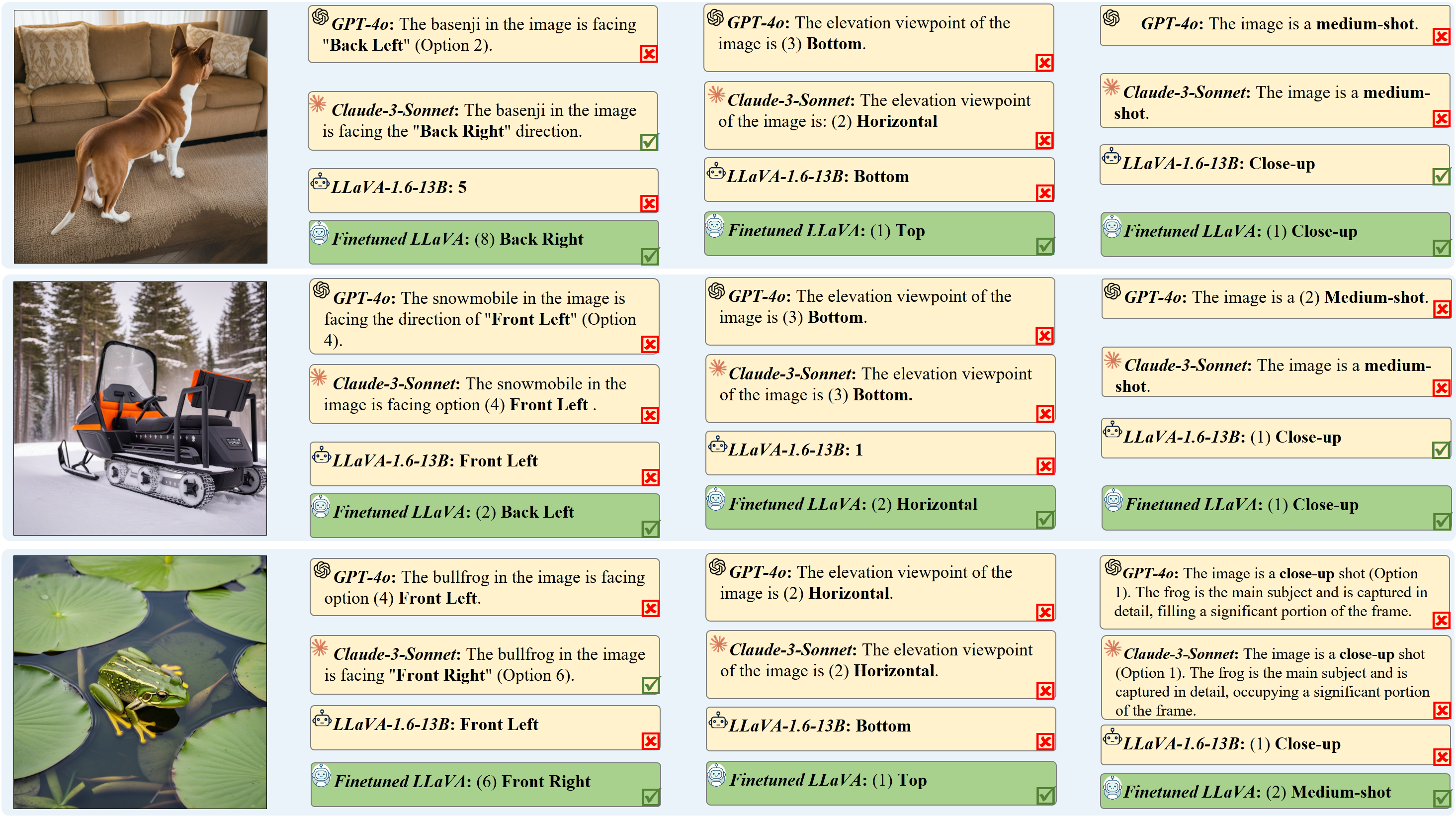}
  \caption{\textbf{Additional Qualitative Results with LLaVA.} We evaluate camera-object relation recognition for GPT-4o, Claude-3-Sonnet, LLaVA-1.6-13B, and finetuned LLaVA-1.6-13B on the \textit{Ultimate3D} benchmark. Each model is asked questions, shown in gray boxes, about object orientation, camera viewpoint, and camera-shot type, together with the input images on the left. The responses show that finetuning LLaVA-1.6-13B on \textit{Ultimate3D} substantially improves camera-object relation recognition compared with the base LLaVA model and closed-source models.}
  \label{fig:supp_qual_llava}
\end{figure*}

\section{Additional Visuals of \textit{Ultimate3D} Dataset}
\label{suppsec:ultimate3D}
TIn Fig.~\ref{suppfig:ultimate3D}, we shows more examples of diversity on object categories, camera-object relation, and background contexts.

\revision{
\section{Generalization to Real Images: ImageNet3D}
\label{app:imagenet3d}

To test whether the camera--object relation knowledge learned from our synthetic data transfers to real photographs, we evaluate on ImageNet3D~\cite{ma2024imagenet3d}, a benchmark of ${\sim}$69K ImageNet images with human-annotated 3D object poses across 189 categories, collected independently of our pipeline. We convert its validation split into our VQA format, keeping images that contain a single quality-checked annotated object (23,224 images, 186 categories). Annotated azimuth and elevation are discretized into our eight orientation and three viewpoint classes; we validated this mapping against PASCAL3D+ labels on the 1,109 images shared by both datasets (97.5\% and 99.0\% agreement, respectively). As ImageNet3D does not annotate camera shot, we derive shot labels from object framing (the maximum bounding-box extent relative to the image), calibrated to our three shot classes, with ambiguous boundary ranges excluded. Since real photographs are strongly biased toward frontal, eye-level views, we sample an answer-balanced evaluation set of 1,200 questions per task. ImageNet3D is used for evaluation only.

Table~\ref{tab:imagenet3d} compares each baseline with its Ultimate3D-fine-tuned counterpart. Fine-tuning improves every model on every task (+15.2, +7.8, and +7.8 points on average for orientation, viewpoint, and camera shot). The baselines largely collapse onto frequent answers (e.g., ``Front'' for most orientation questions), whereas the fine-tuned models spread predictions across all classes and gain most on rear and oblique views, indicating improved camera--object understanding rather than better answer priors. The one residual gap is the Long-shot class, whose accuracy decreases after fine-tuning; we attribute this to our synthetic long shots being clean, object-centered renders, unlike cluttered real-world wide shots, and leave closing this gap to future work.

\begin{table*}[t]
\centering
\caption{Accuracy (\%) on the answer-balanced ImageNet3D evaluation set, reported as base $\rightarrow$ Ultimate3D-fine-tuned. Chance is 12.5\% for orientation and 33.3\% for viewpoint and camera shot.}
\label{tab:imagenet3d}
\small
\setlength{\tabcolsep}{5pt}
\begin{tabular}{@{}lccc@{}}
\toprule
Model & Orientation & Viewpoint & Camera shot \\
\midrule
Qwen3.5-2B  & 17.2 $\rightarrow$ \textbf{32.4} & 33.3 $\rightarrow$ \textbf{45.2} & 40.9 $\rightarrow$ \textbf{49.7} \\
Qwen3.5-9B  & 20.7 $\rightarrow$ \textbf{34.3} & 41.3 $\rightarrow$ \textbf{46.9} & 42.4 $\rightarrow$ \textbf{48.8} \\
Qwen3-VL-8B & 20.5 $\rightarrow$ \textbf{34.4} & 37.8 $\rightarrow$ \textbf{50.1} & 43.6 $\rightarrow$ \textbf{51.8} \\
Gemma-4-E2B-it & 12.25 $\rightarrow$ \textbf{27.17} & 31.67 $\rightarrow$ \textbf{43.25} & 39.08 $\rightarrow$ \textbf{46.58} \\
Gemma-4-E4B-it & 16.83 $\rightarrow$ \textbf{28.42} & 40.00 $\rightarrow$ \textbf{46.33} & 41.08 $\rightarrow$ \textbf{49.58} \\
Gemma-4-26B-A4B-it & 16.83 $\rightarrow$ \textbf{34.08} & 43.42 $\rightarrow$ \textbf{47.33} & 42.33 $\rightarrow$ \textbf{51.42} \\
Gemma-4-31B-it & 14.08 $\rightarrow$ \textbf{34.00} & 40.67 $\rightarrow$ \textbf{43.92} & 43.33 $\rightarrow$ \textbf{49.42} \\
\midrule
Avg.\ gain  & +15.2 & +7.8 & +7.8 \\
\bottomrule
\end{tabular}
\end{table*}

\section{Failure Analysis}
\label{app:failure}

We analyse the errors that remain after Ultimate3D fine-tuning on the synthetic
Ulti3D benchmark, the real-image PASCAL3D+ and MEBOW splits, and the ImageNet3D
set of Appendix~\ref{app:imagenet3d}. All ranges below are over the three
fine-tuned models.

\noindent \textbf{Which relations remain challenging.} Table~\ref{tab:fail:classes} gives per-class accuracy. On synthetic images
accuracy is nearly uniform across classes, so the answer classes themselves are
no longer a source of systematic error. On real photographs three relations stay
hard: \emph{rear-oblique orientations} (Back-Left/Back-Right, $42$--$61\%$ on
PASCAL3D+ against $74$--$85\%$ for the pure side views; Back-Left$\to$Left and
Back-Right$\to$Right alone account for $117$ of Qwen3.5-9B's $449$ PASCAL3D+
orientation errors), the \emph{Bottom viewpoint} ($41$--$59\%$ on PASCAL3D+,
$25$--$47\%$ on ImageNet3D, mostly answered Horizontal), and \emph{Long-shot
framing} on real photographs ($14$--$22\%$), where our clean object-centred
renders transfer poorly to cluttered wide shots. People (MEBOW) are the weakest
object class overall, and only $2.3\%$ of our synthetic training questions
depict humans, none of them ordinary pedestrians.

\begin{table*}[!t]
\centering\small\setlength{\tabcolsep}{5pt}
\caption{Per-class accuracy (\%) after Ultimate3D fine-tuning; each cell is the
range over Qwen3.5-2B, Qwen3.5-9B and Qwen3-VL-8B. PASCAL3D+ carries no
camera-shot annotation and MEBOW only orientation. Rear-oblique orientations,
Bottom and Long-shot are the residual weak classes.}
\label{tab:fail:classes}
\begin{tabular}{llcccc}
\toprule
Relation & Class & Synthetic & PASCAL3D+ & MEBOW & ImageNet3D \\
\midrule
\multirow{9}{*}{Orientation}
 & Back        & 79--91 & 48--62 & 48--68 & 28--43 \\
 & Back Left   & 76--89 & 43--61 & 45--70 & 26--43 \\
 & Left        & 87--89 & 74--85 & 33--47 & 35--53 \\
 & Front Left  & 77--90 & 54--73 & 29--39 & 24--35 \\
 & Front       & 83--89 & 66--71 & 28--34 & 29--39 \\
 & Front Right & 72--85 & 62--69 & 30--39 & 20--48 \\
 & Right       & 87--88 & 75--85 & 46--58 & 40--41 \\
 & Back Right  & 71--93 & 42--57 & 28--47 & 13--37 \\
 & \emph{Overall} & 82--86 & 59--64 & 37--38 & 32--34 \\
\midrule
\multirow{4}{*}{Viewpoint}
 & Top         & 70--85 & 52--64 & -- & 42--61 \\
 & Horizontal  & 63--91 & 66--76 & -- & 47--59 \\
 & Bottom      & 57--72 & 41--59 & -- & 25--47 \\
 & \emph{Overall} & 67--83 & 58--62 & -- & 45--50 \\
\midrule
\multirow{4}{*}{Camera shot}
 & Close-up    & 89--96 & -- & -- & 82--90 \\
 & Medium-shot & 83--88 & -- & -- & 40--52 \\
 & Long-shot   & 88--91 & -- & -- & 14--22 \\
 & \emph{Overall} & 87--92 & -- & -- & 49--52 \\
\bottomrule
\end{tabular}
\end{table*}

\noindent\textbf{Residual errors are one bin off.} Fine-tuning changes the shape of the errors, not just their number: on PASCAL3D+
the share of orientation errors landing in a bin \emph{adjacent} to the truth
rises from $38$--$55\%$ (base) to $63$--$64\%$, while gross ($90^\circ$) errors
fall from $15$--$30\%$ to $8$--$15\%$. Tolerating a one-bin error raises
PASCAL3D+ accuracy from $59$--$64\%$ to $85$--$87\%$ and synthetic accuracy from
$82$--$86\%$ to $91$--$93\%$ (base models reach only $55$--$67\%$ and
$69$--$75\%$ under the same tolerance). A second mode sits at $180^\circ$
($20$--$22\%$ of synthetic and $8$--$16\%$ of PASCAL3D+ orientation errors) and
concentrates on objects whose silhouette hides front from back --- fans,
monitors, sunglasses, and boats, buses or trains seen end-on
(Figure~\ref{fig:fail:examples}, row 2).

\noindent\textbf{Errors concentrate at the label boundaries.} Our benchmark discretises continuous camera parameters, and the residual errors
are largely poses near a class boundary: using the exact render parameters,
synthetic accuracy falls to $69$--$71\%$ for orientations within $8^\circ$ of a
$45^\circ$ bin edge, to $54$--$59\%$ for elevations within $5^\circ$ of the
$\pm15^\circ$ band edge and to $36$--$44\%$ for Medium-shots within $0.15$ of the
Close-up threshold, against $70$--$93\%$ away from those boundaries; on
ImageNet3D, whose azimuths are human-annotated, accuracy falls from $39$--$41\%$
to $25$--$28\%$ within $8^\circ$ of a boundary, and from $35$--$37\%$ to
$15$--$20\%$ for objects covering less than $30\%$ of the frame.

\noindent \textbf{Image conditions.} We annotated $456$ images: per (source, relation) group, $30$ that all three
fine-tuned models answered incorrectly, $20$ answered incorrectly by one or two
of them, and $20$ answered correctly by all three as a matched control. Two
annotators labelled every image independently and blind to all model
predictions, with $80$--$99\%$ agreement per field.
Table~\ref{tab:fail:conditions} compares the shared failures with the matched
successes. Failures are enriched for subtle geometric cues, near-boundary poses,
clutter, small objects and shape-symmetric objects, but \emph{not} for occlusion
or truncation ($33\%$ vs.\ $32\%$ on real photographs): partial visibility is not
the limiting factor, cue strength is. Person images behave differently, with
clutter the strongest signal ($43\%$ vs.\ $0\%$) alongside head--torso direction
conflicts --- conditions absent from our single-object renders.

\begin{table*}[!t]
\centering\small\setlength{\tabcolsep}{12pt}
\caption{Prevalence (\%) of image conditions among images all three fine-tuned
models answered incorrectly (F) and matched images all three answered correctly
(S); $n=90$ and $n=60$ per column pair. Real = PASCAL3D+ and MEBOW.}
\label{tab:fail:conditions}
\begin{tabular}{lcccc}
\toprule
& \multicolumn{2}{c}{Real photographs} & \multicolumn{2}{c}{Synthetic} \\
\cmidrule(lr){2-3}\cmidrule(lr){4-5}
Condition & F & S & F & S \\
\midrule
Subtle directional / elevation cue & \textbf{52} & 27 & \textbf{41} & 28 \\
Pose near a class boundary         & \textbf{47} & 28 & \textbf{41} & 27 \\
Cluttered background               & \textbf{37} & 18 & 18 & 13 \\
Small object ($<15\%$ of frame)    & \textbf{24} & 15 & \textbf{36} & 20 \\
Shape-symmetric object             &  9 &  2 &  6 &  3 \\
Unusual or articulated pose        & 13 & 12 & 17 & 13 \\
Multiple instances of the category & 12 & 10 &  4 &  7 \\
Occluded or truncated by the frame & 33 & 32 & 23 & 43 \\
\midrule
Ground-truth label judged questionable & 38 & 0 & 34 & 15 \\
\bottomrule
\end{tabular}
\end{table*}

\noindent\textbf{Label ambiguity.} The last row of Table~\ref{tab:fail:conditions} is a limitation of the
benchmarks. On orientation the annotators reproduced the ground-truth label for
$83$--$93\%$ of the shared failures, so those are genuine model errors; on
viewpoint and camera shot only $43$--$70\%$, because the discretisation admits no
visual resolution ($16^\circ$ of elevation is Top, $14^\circ$ is Horizontal). We
therefore also report adjacent-bin accuracy for orientation and note the
$\pm15^\circ$ elevation band and the fixed distance thresholds as design choices
whose boundaries are inherently ambiguous.

\begin{figure*}[!tp]
\centering
\includegraphics[width=\textwidth]{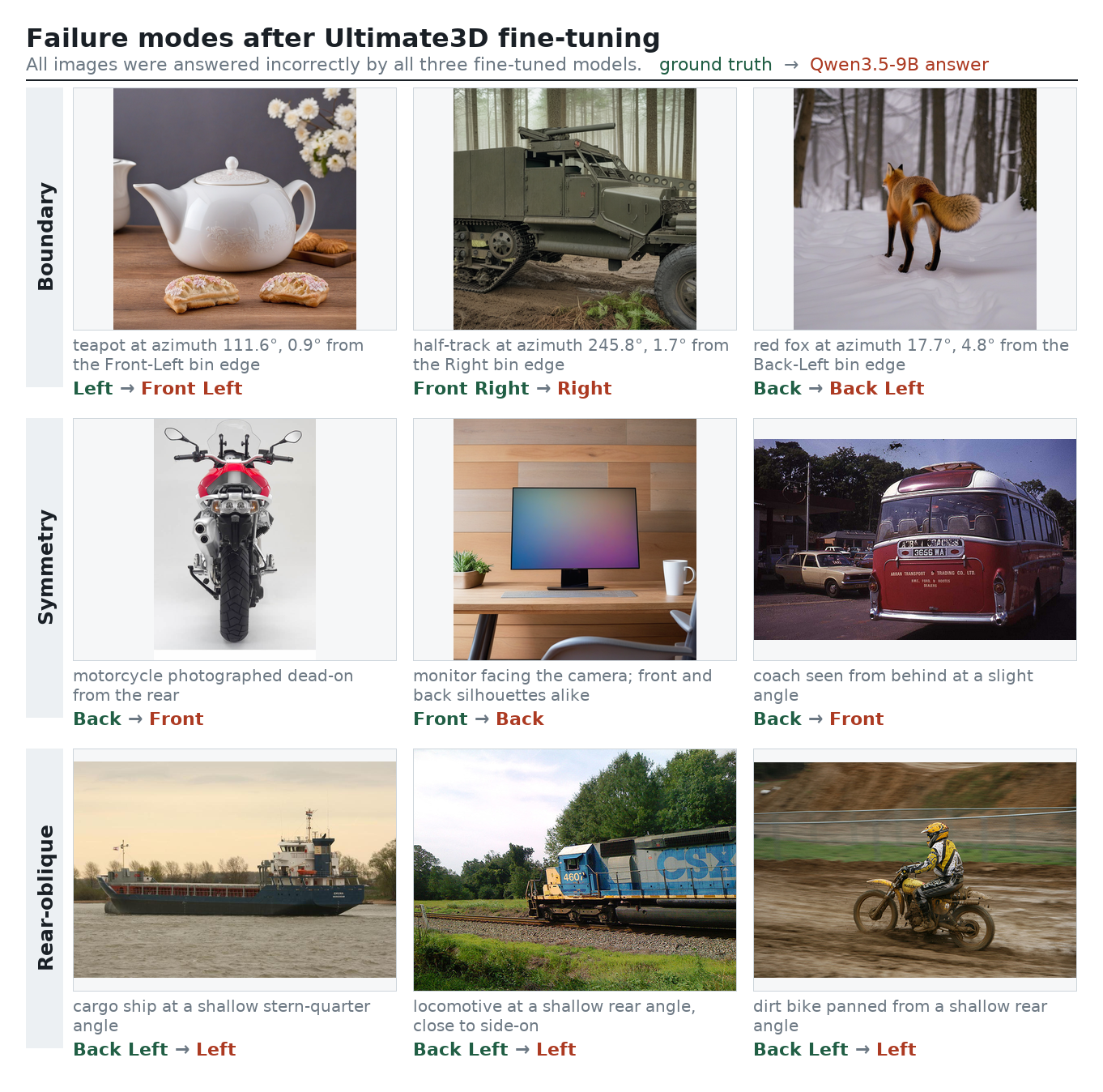}
\caption{Representative failures, part 1 of 2: geometric cues in the image
plane. Every image was answered incorrectly by all three fine-tuned models;
panels give the ground-truth label (green) and the answer of
Qwen3.5-9B\,+\,Ultimate3D (red).
\textbf{Boundary:} the true azimuth is within a few degrees of a bin edge.
\textbf{Symmetry:} front and back are barely distinguishable and the prediction
is inverted. \textbf{Rear-oblique:} a shallow rear three-quarter view is answered
with the neighbouring side bin. Elevation, framing and people failures are shown
in Fig.~\ref{fig:fail:examples:b}.}
\label{fig:fail:examples}
\end{figure*}

\begin{figure*}[!tp]
\centering
\includegraphics[width=\textwidth]{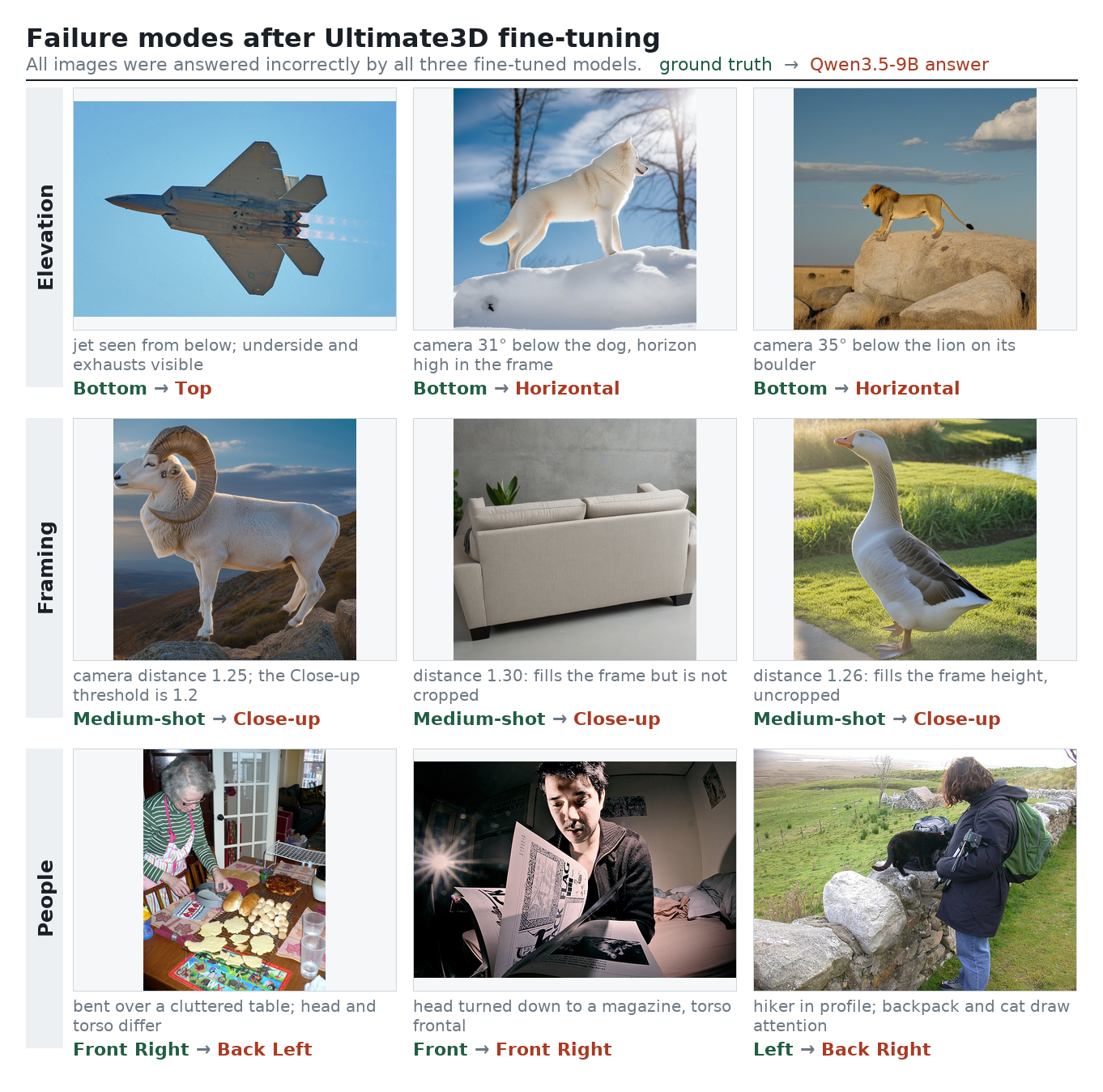}
\caption{Representative failures, part 2 of 2: viewpoint, framing and people;
continued from Fig.~\ref{fig:fail:examples}. Every image was answered incorrectly
by all three fine-tuned models; panels give the ground-truth label (green) and
the answer of Qwen3.5-9B\,+\,Ultimate3D (red).
\textbf{Elevation:} the camera is above or below but the cue is faint.
\textbf{Framing:} the object fills the frame uncropped, at the
Close-up/Medium-shot threshold. \textbf{People:} clutter, small scale, or a
head facing away from the torso.}
\label{fig:fail:examples:b}
\end{figure*}

\section{Additional Discussions}
\label{suppsec:finetune_dicussion}
We provide some insights of using synthetic generated visual instruction dataset for MLLM finetuning. For the task of camera-object relation recognition for 100 categories (as we collected in \textit{Ultimate3D} dataset), a dataset comprising 100K to 1M VQAs will significantly improve baseline MLLMs by one-epoch finetuning. Further scaling up the dataset volume, finetuneing epochs, or model parameters may yield saturation.

Moreover, the quality of synthetic dataset is crucial to the improvement margin. Using dataset with higher success rate (ours is 93.07\% as main paper Sec.~\ref{sec:dataset_quality}) provides better performances than using dataset generated other alternative backbones (i.e. SD-V1.5).

The coverage of arbitrary camera-object relation is also important for generalization. Our preliminary test using real images which are highly biased to front-facing direction from Pascal3D+~\cite{xiang_wacv14}, results in poor generalization on uncovered orientations and viewpoints. This illustrates the importance of using synthetic dataset to eliminate the dataset biases.

For numerical prediction of camera-object relation, our preliminary test shows plausible success using MEBOW~\cite{wu2020mebow} dataset to improve MLLMs on predicting human orientation angles. The average degree prediction accuracy is competitive to the original paper~\cite{wu2020mebow}. However, the extended numerical prediction of 100 categories and 3 types of camera-object relations may require even larger dataset volume than \textit{Ultimate3D}. Our framework is able to scale up the dataset volume. But due to the limitation of computing power, we leave it as interesting future work.
}

\begin{figure*}[!tp]
\centering
\setlength{\tabcolsep}{2pt}
\begin{tabular}{p{1em}cccccc}
   \rotatebox{90}{\textbf{Front}} & 
   \includegraphics[width=0.145\textwidth]{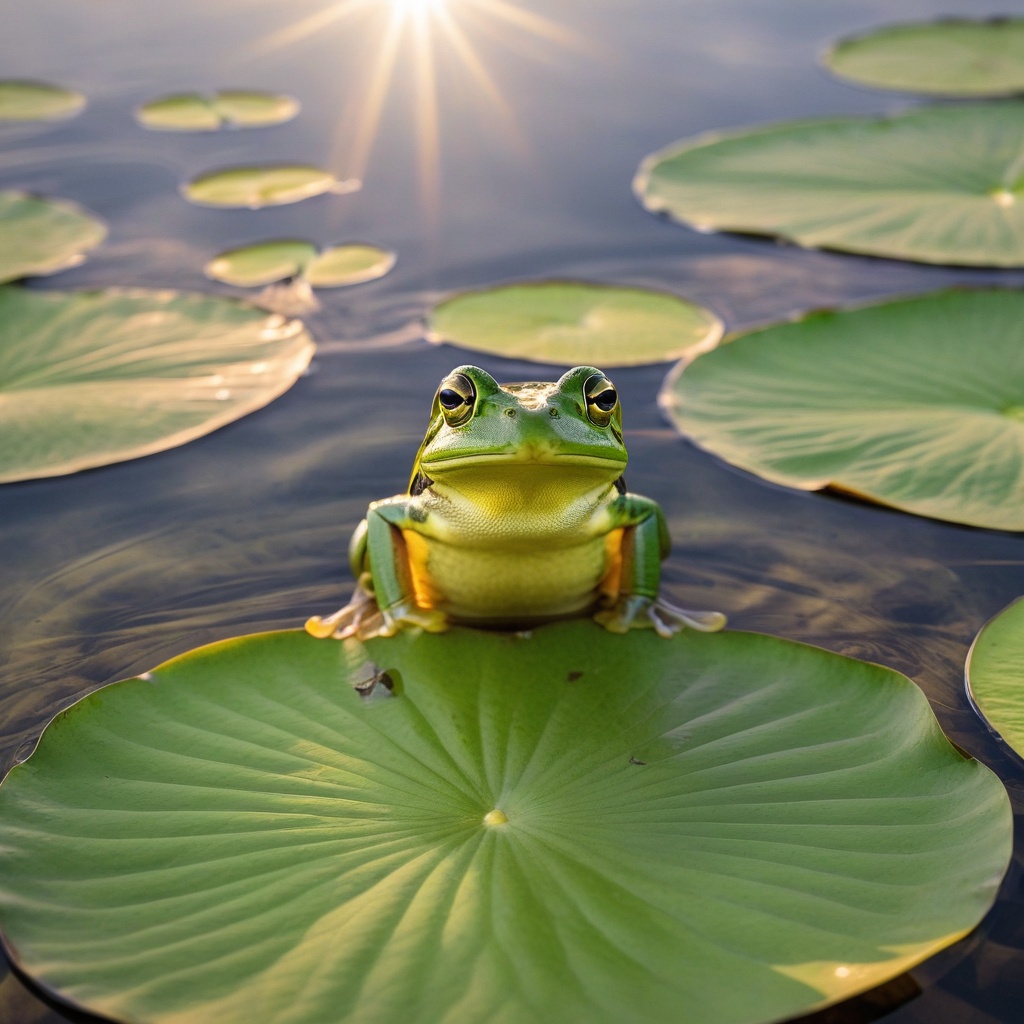} &    
    \includegraphics[width=0.145\textwidth]{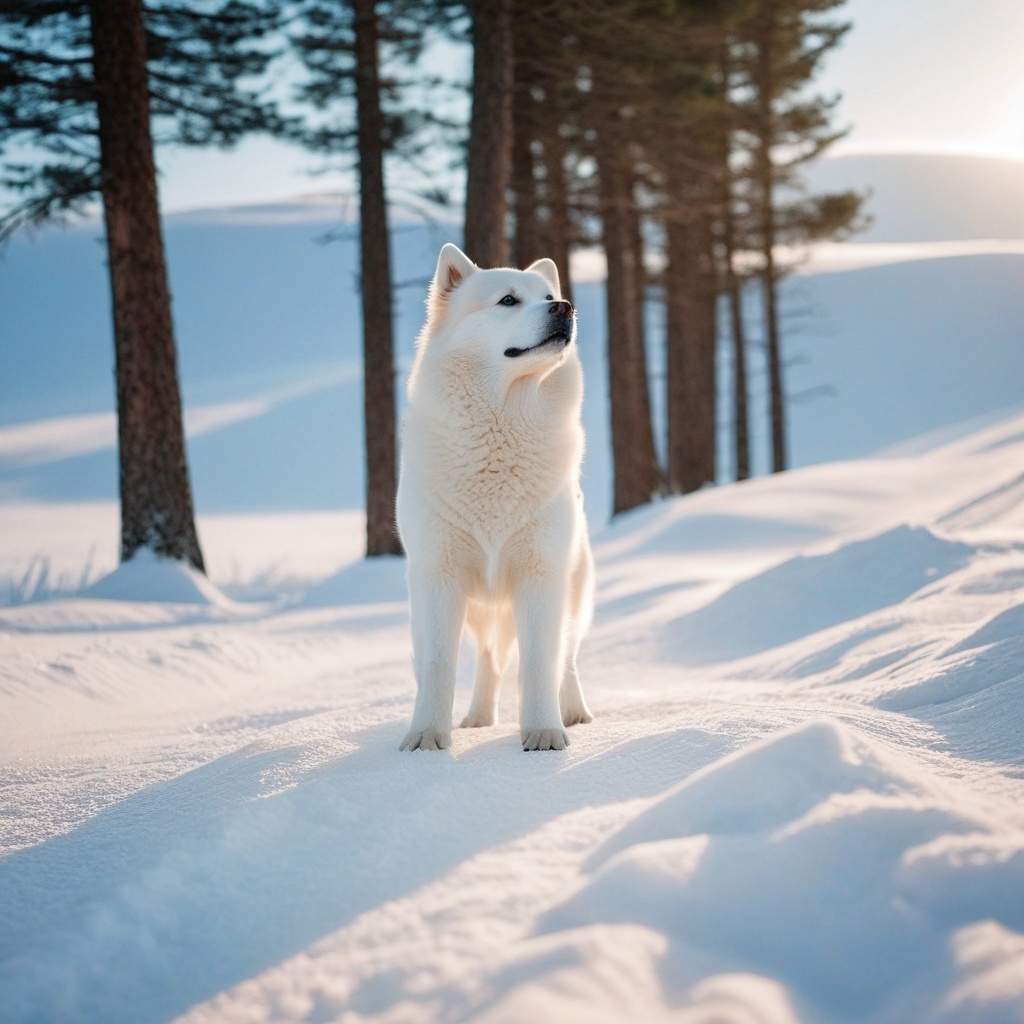} &
   \includegraphics[width=0.145\textwidth]{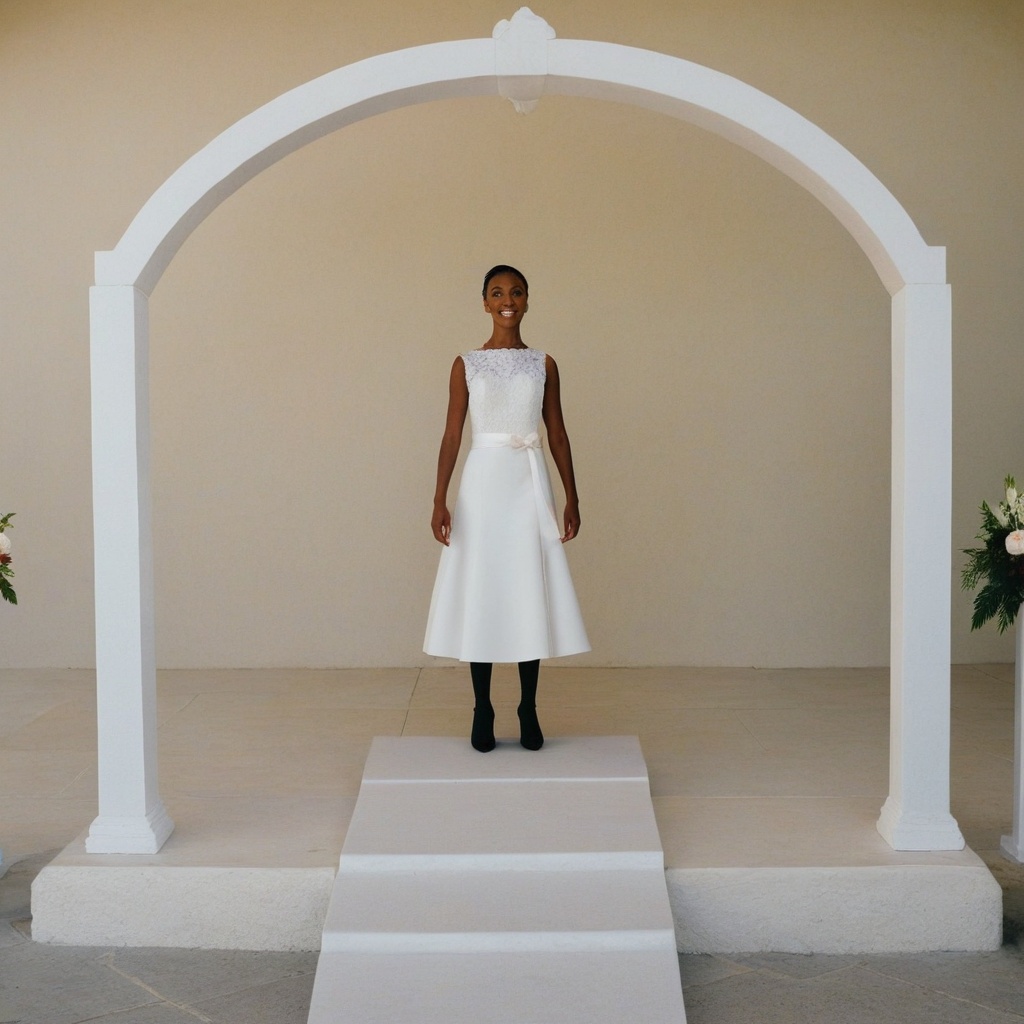} &
    \includegraphics[width=0.145\textwidth]{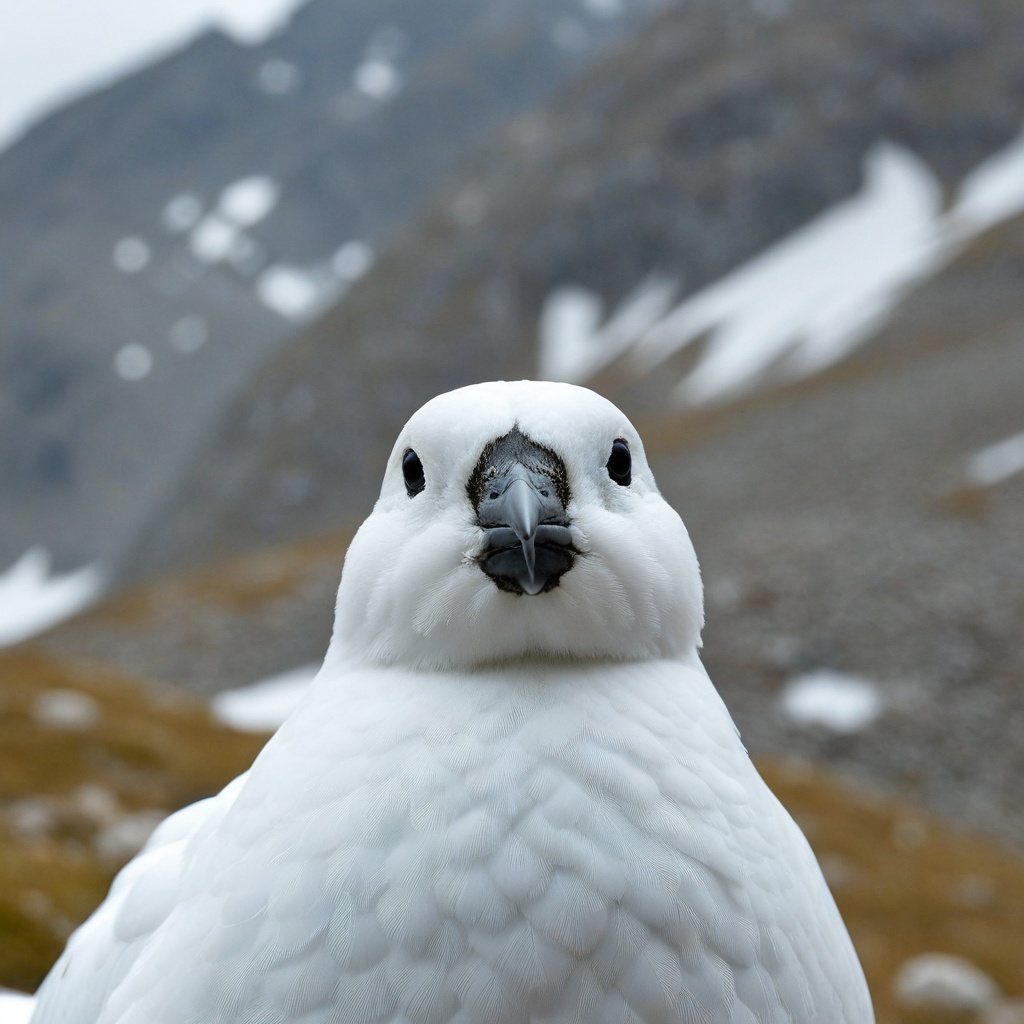} &
    \includegraphics[width=0.145\textwidth]{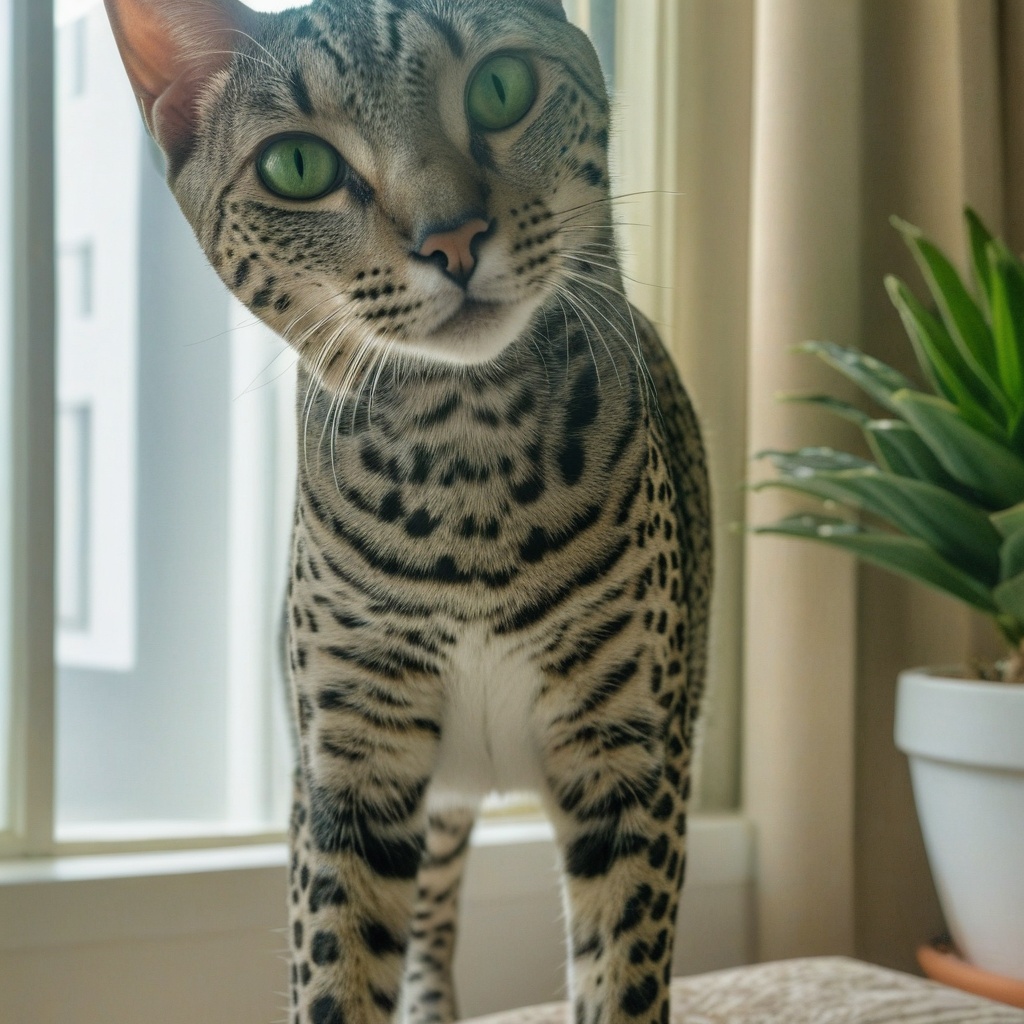} &
    \includegraphics[width=0.145\textwidth]{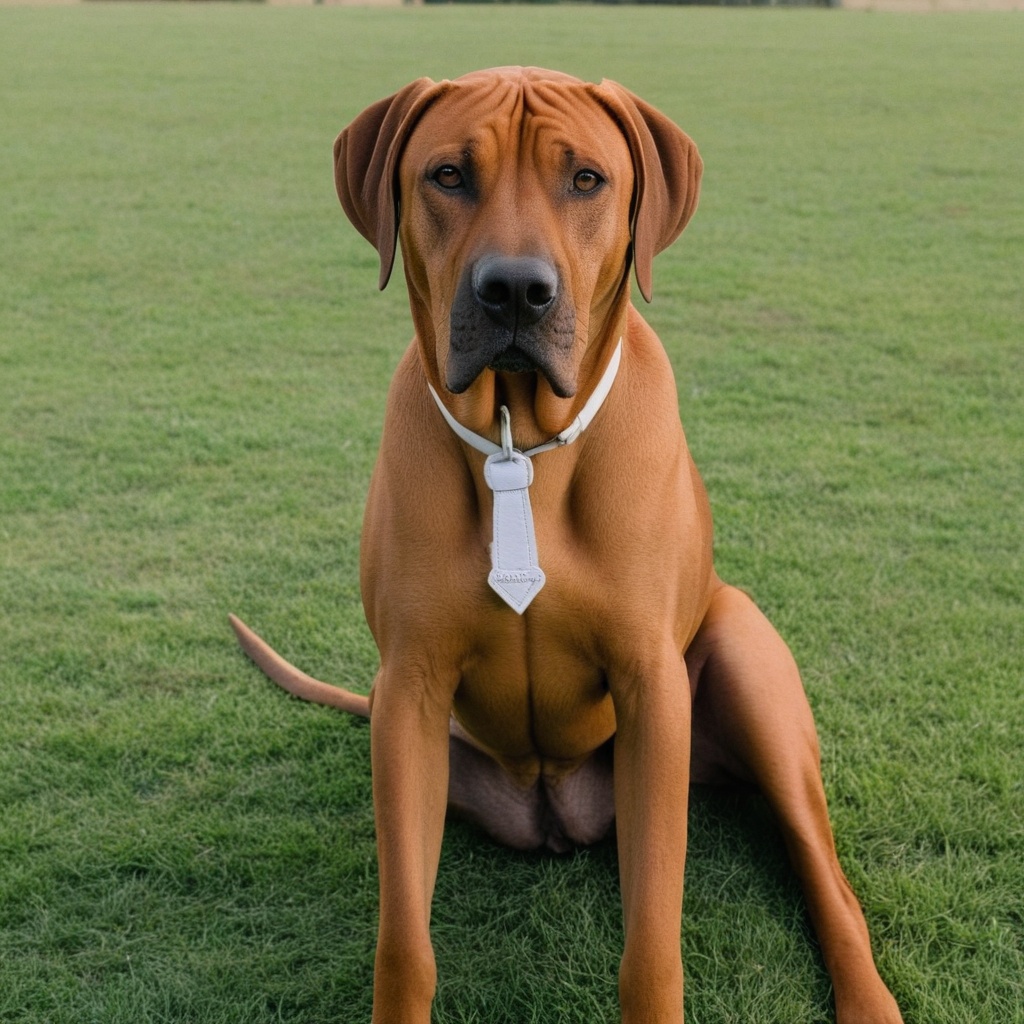} 
    \\
   \rotatebox{90}{\textbf{Front Left}} & 
    \includegraphics[width=0.145\textwidth]{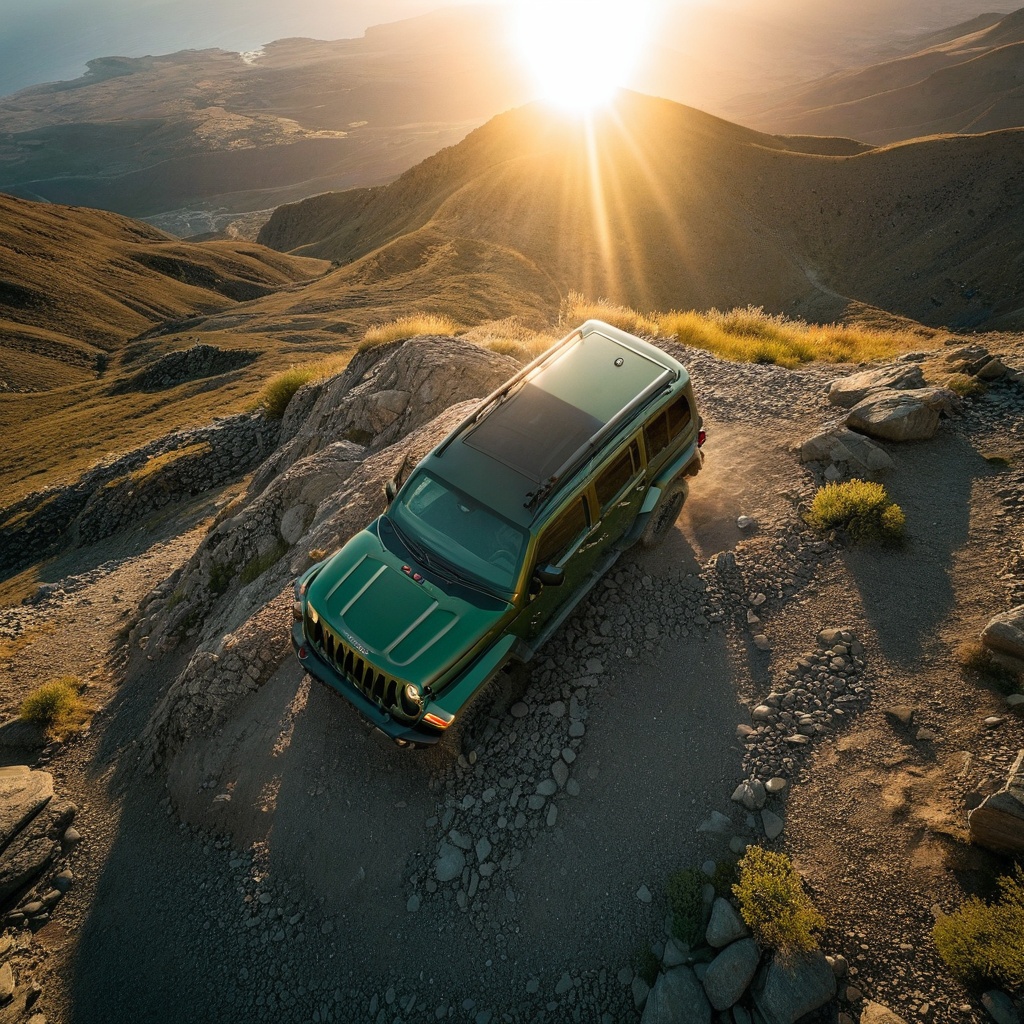} &
   \includegraphics[width=0.145\textwidth]{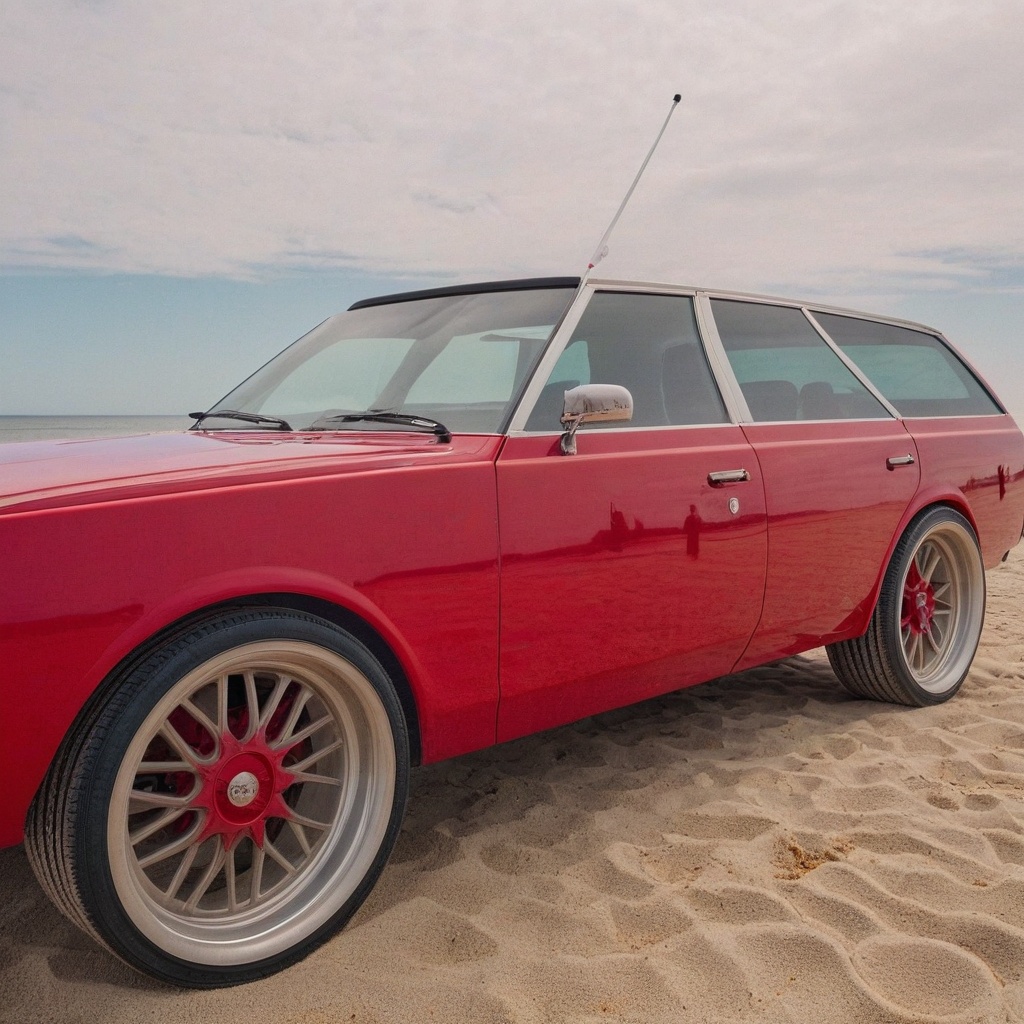} &    
   \includegraphics[width=0.145\textwidth]{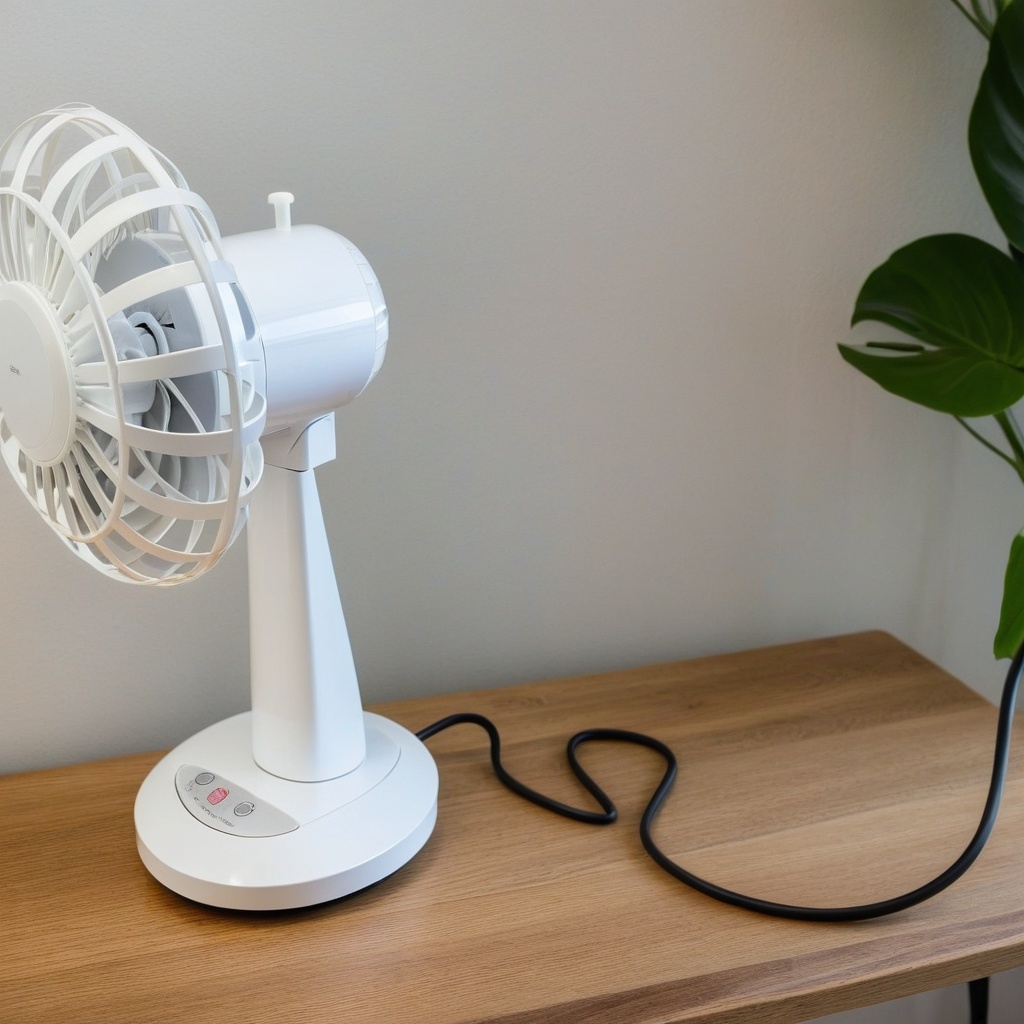} &
    \includegraphics[width=0.145\textwidth]{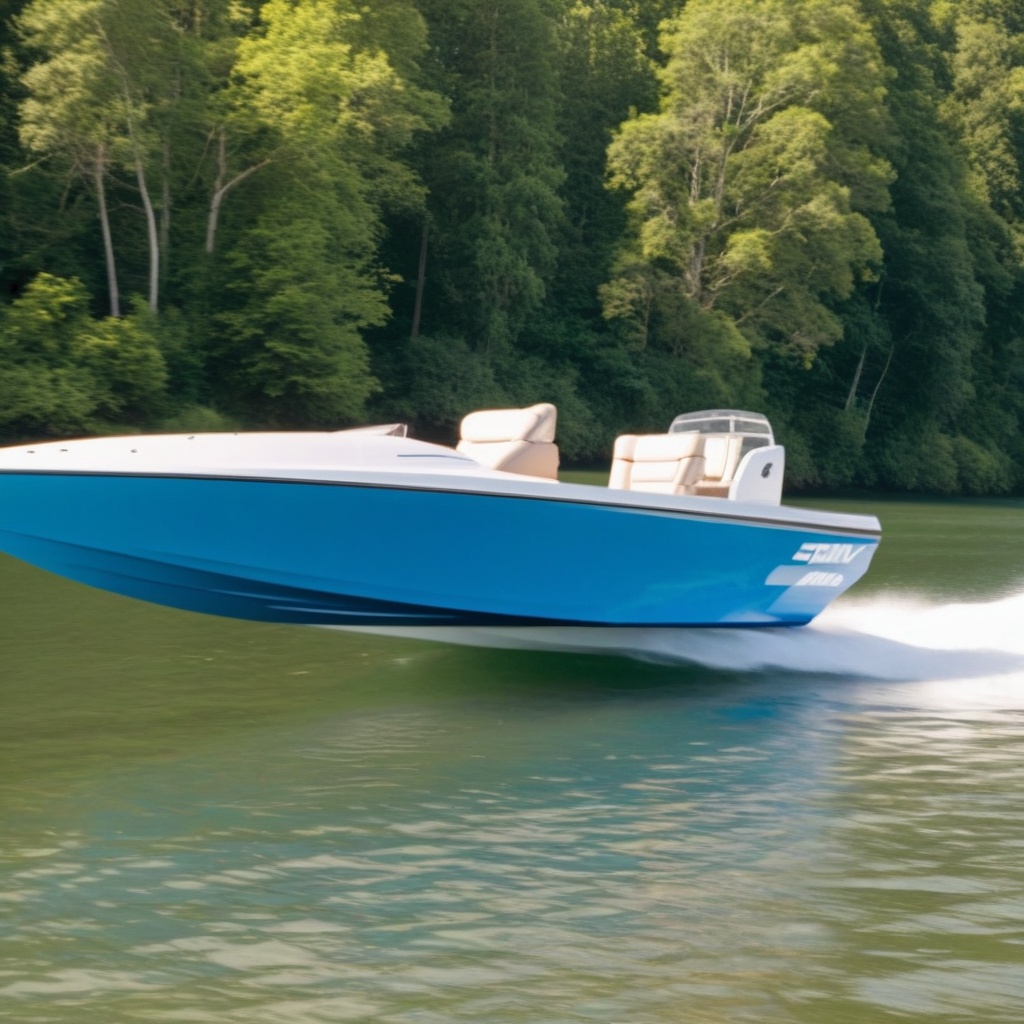} & 
    \includegraphics[width=0.145\textwidth]{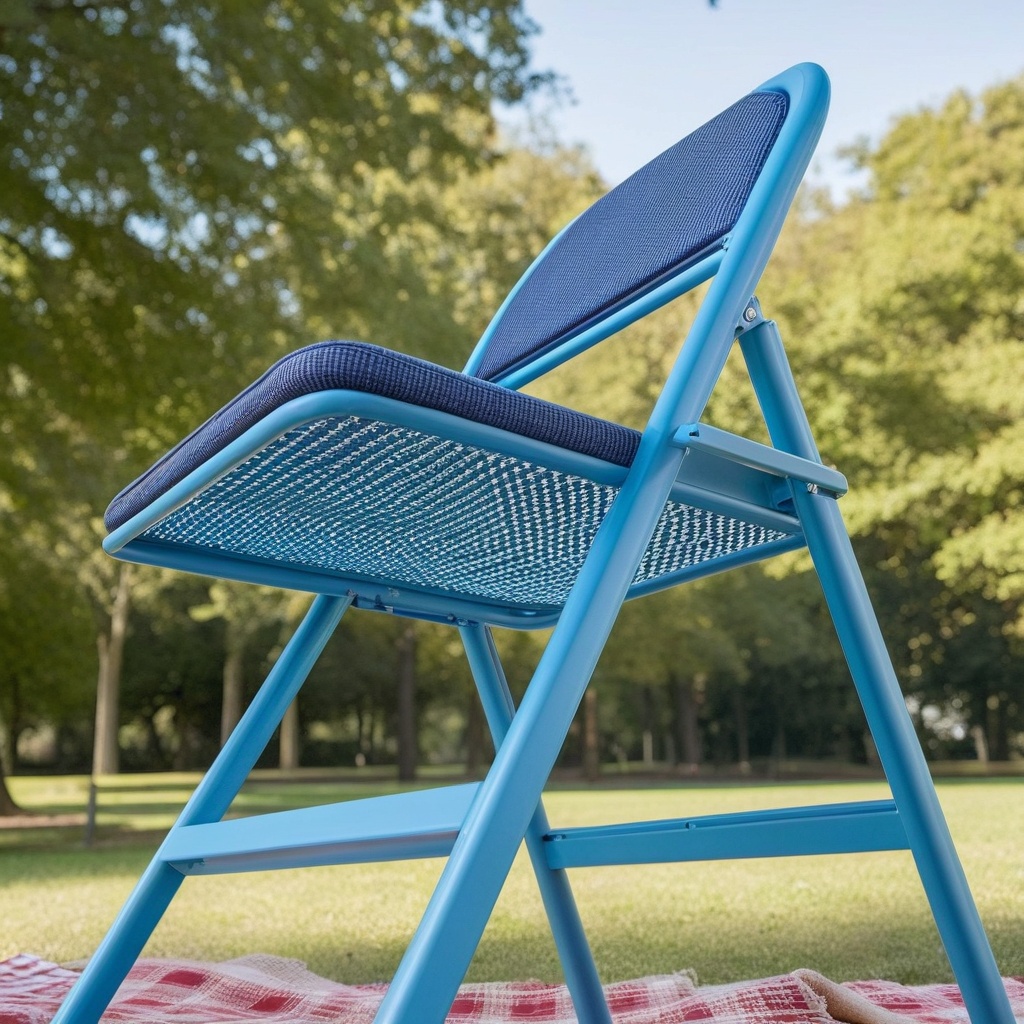} &
    \includegraphics[width=0.145\textwidth]{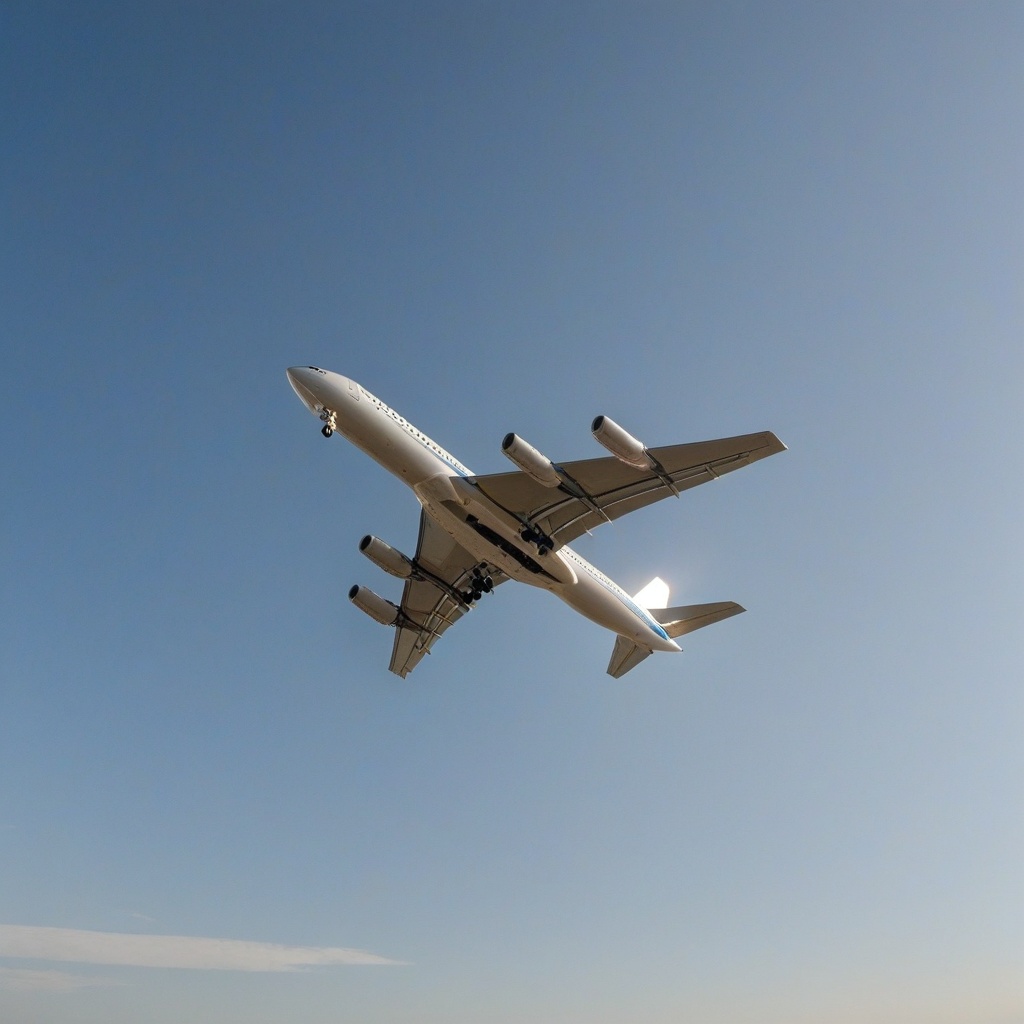} 
    \\
   \rotatebox{90}{\textbf{Left}} & 
    \includegraphics[width=0.145\textwidth]{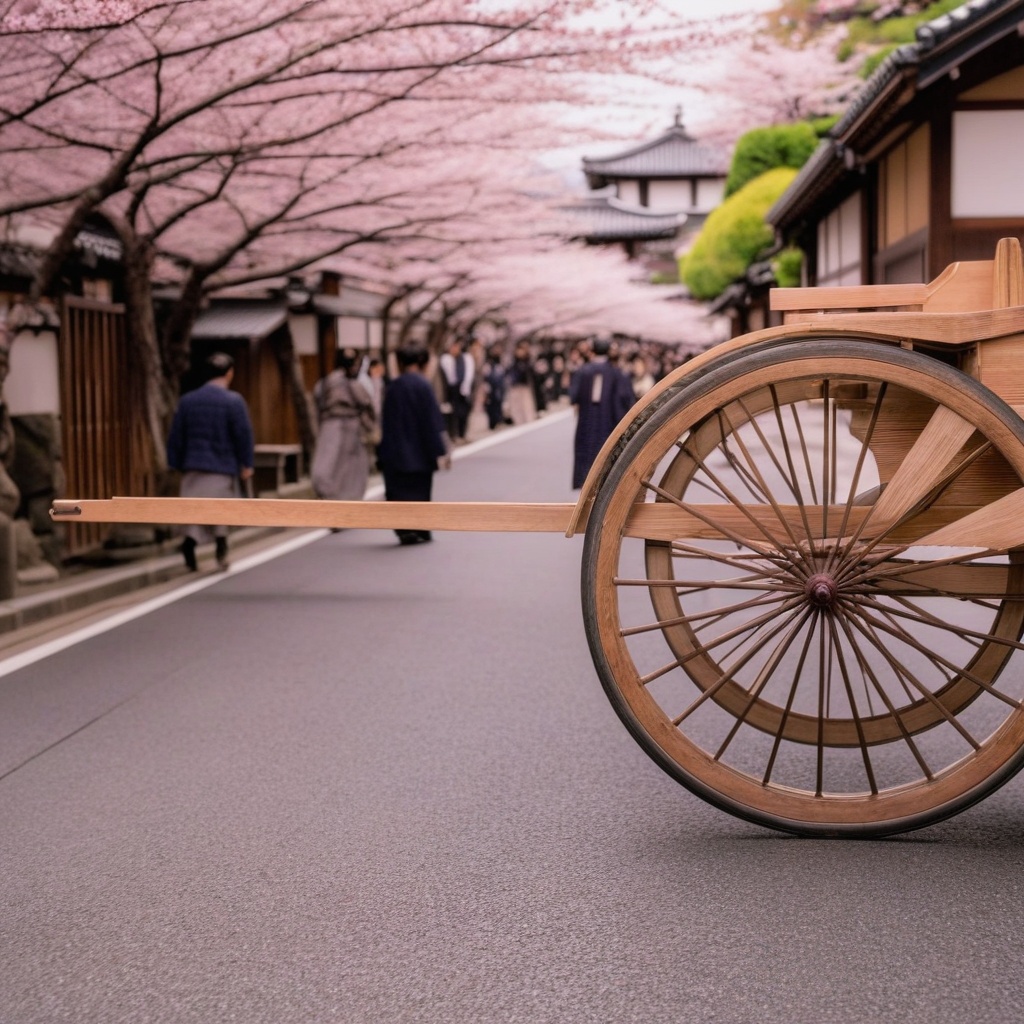} &
   \includegraphics[width=0.145\textwidth]{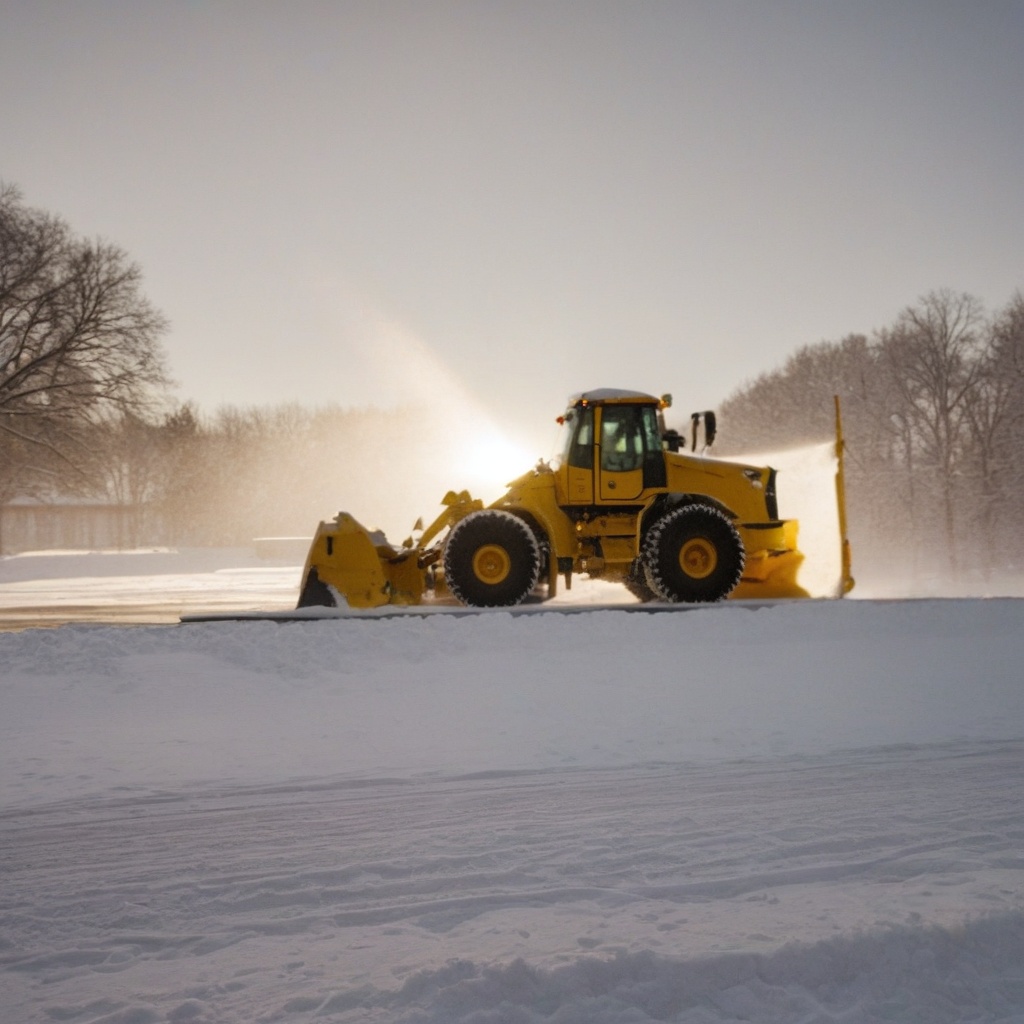} &    
   \includegraphics[width=0.145\textwidth]{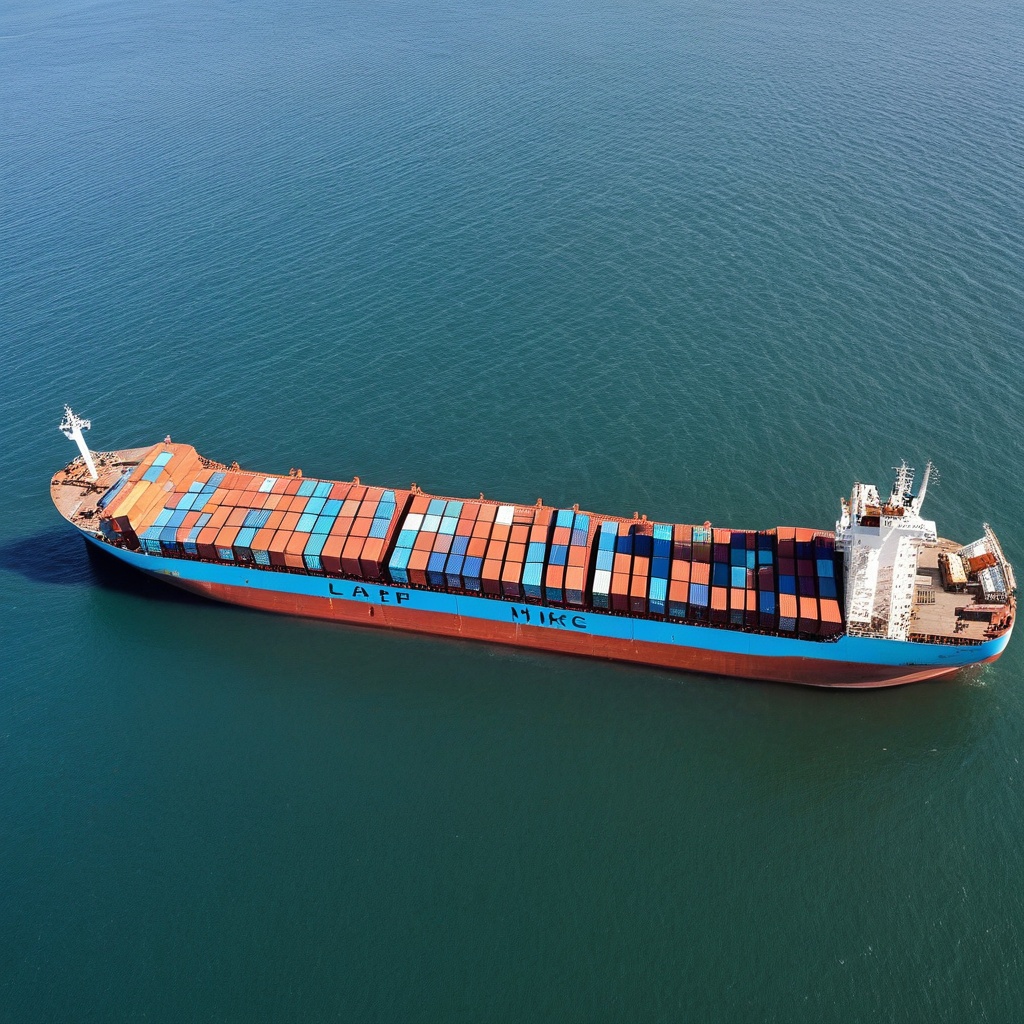} &
    \includegraphics[width=0.145\textwidth]{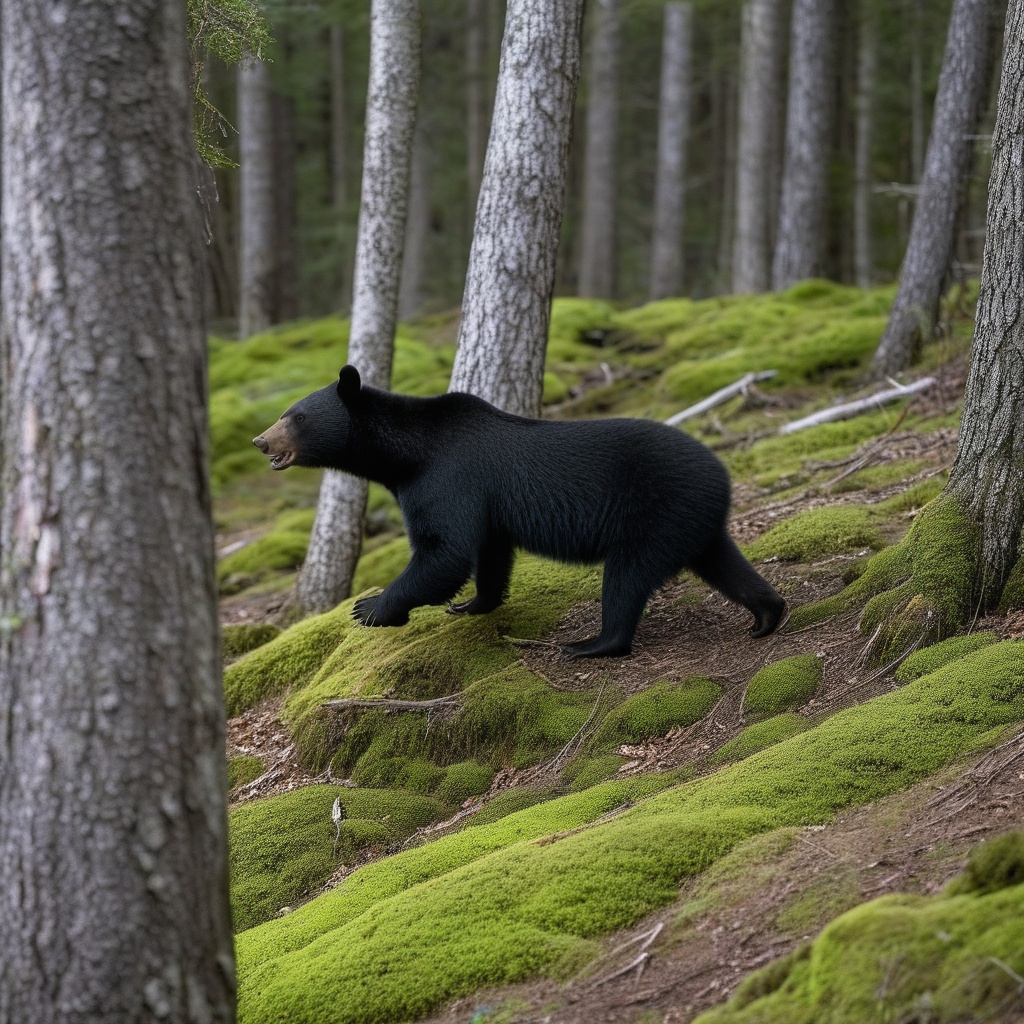} &
    \includegraphics[width=0.145\textwidth]{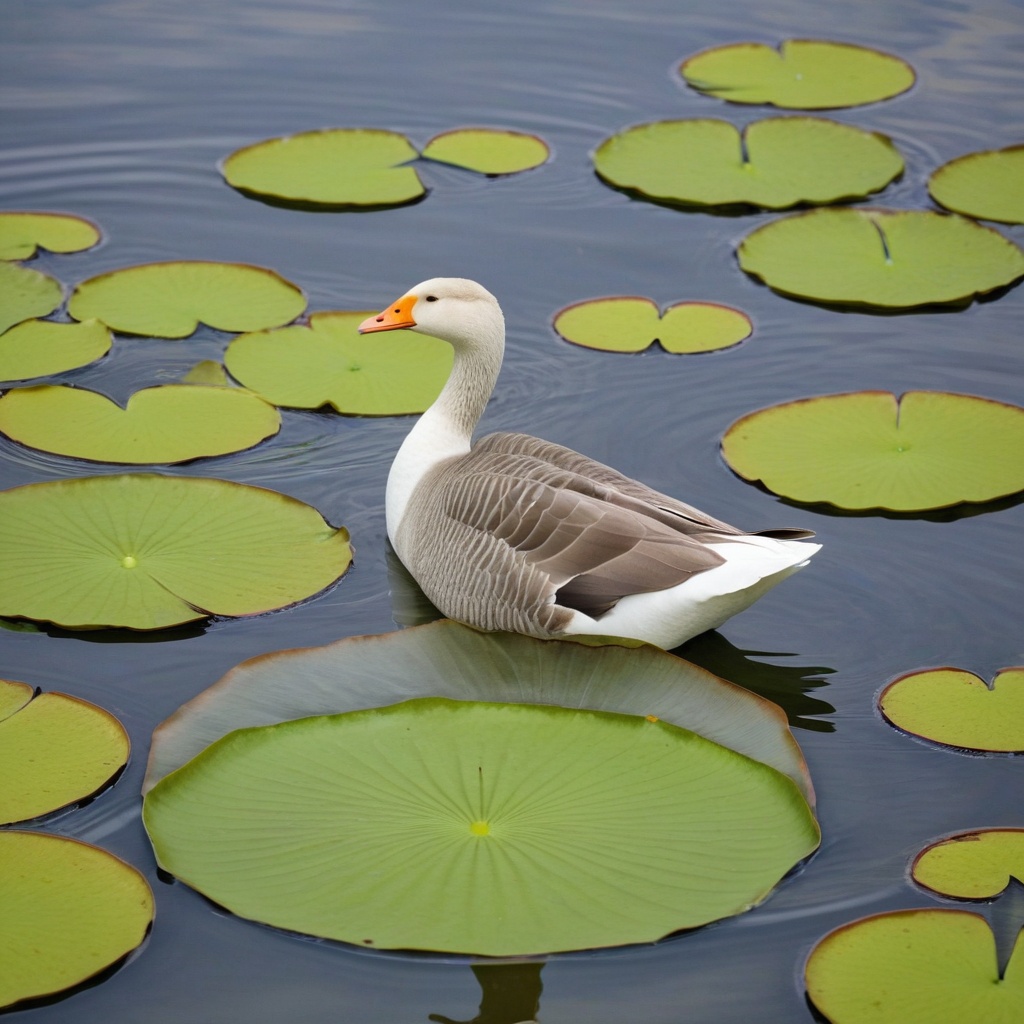} &
    \includegraphics[width=0.145\textwidth]{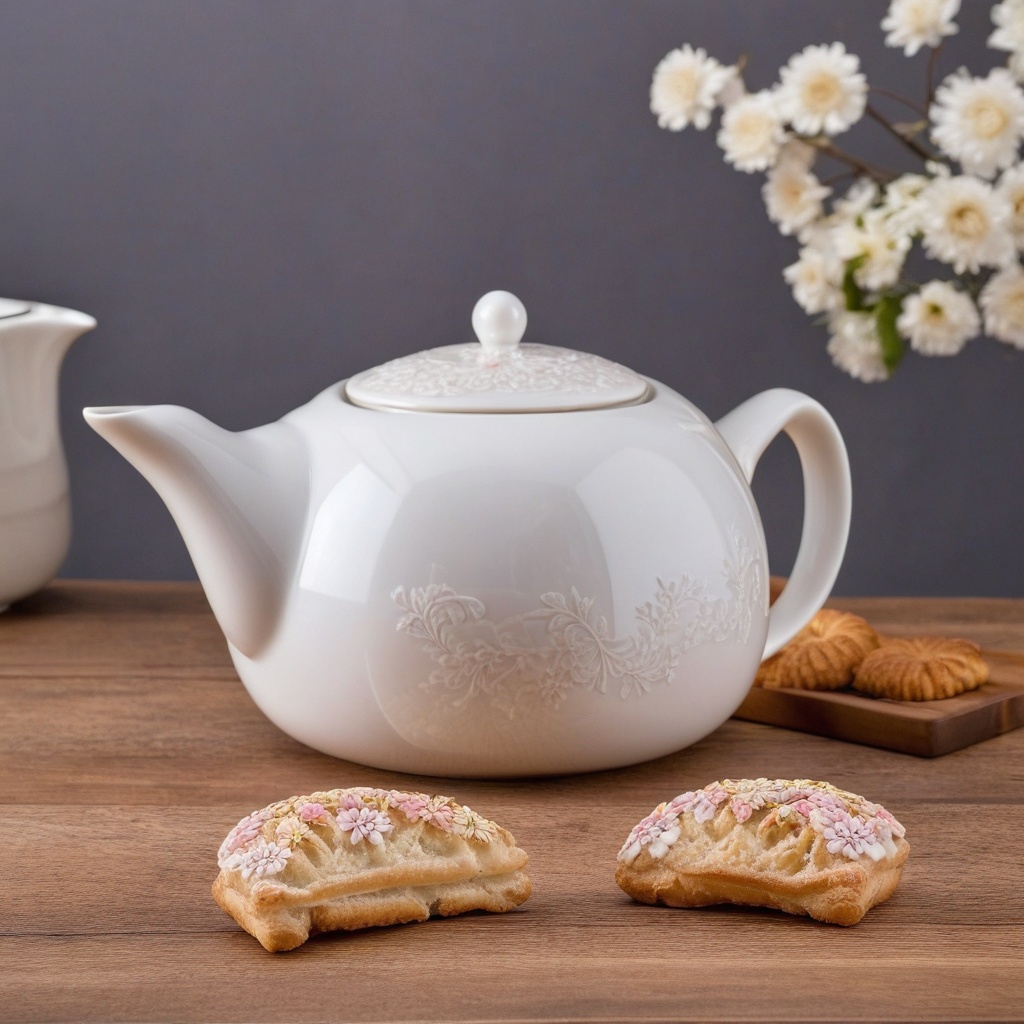} 
    \\
   \rotatebox{90}{\textbf{Back Left}} & 
    \includegraphics[width=0.145\textwidth]{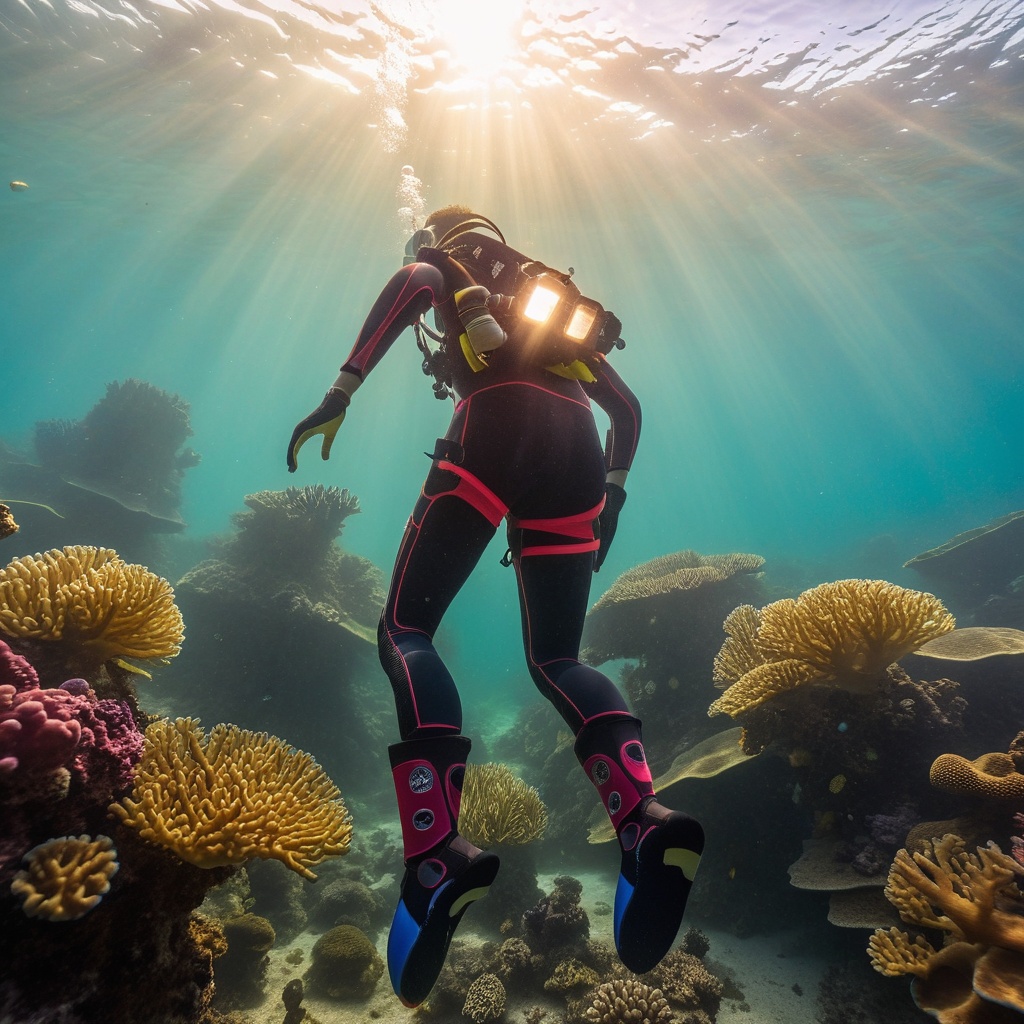} &
   \includegraphics[width=0.145\textwidth]{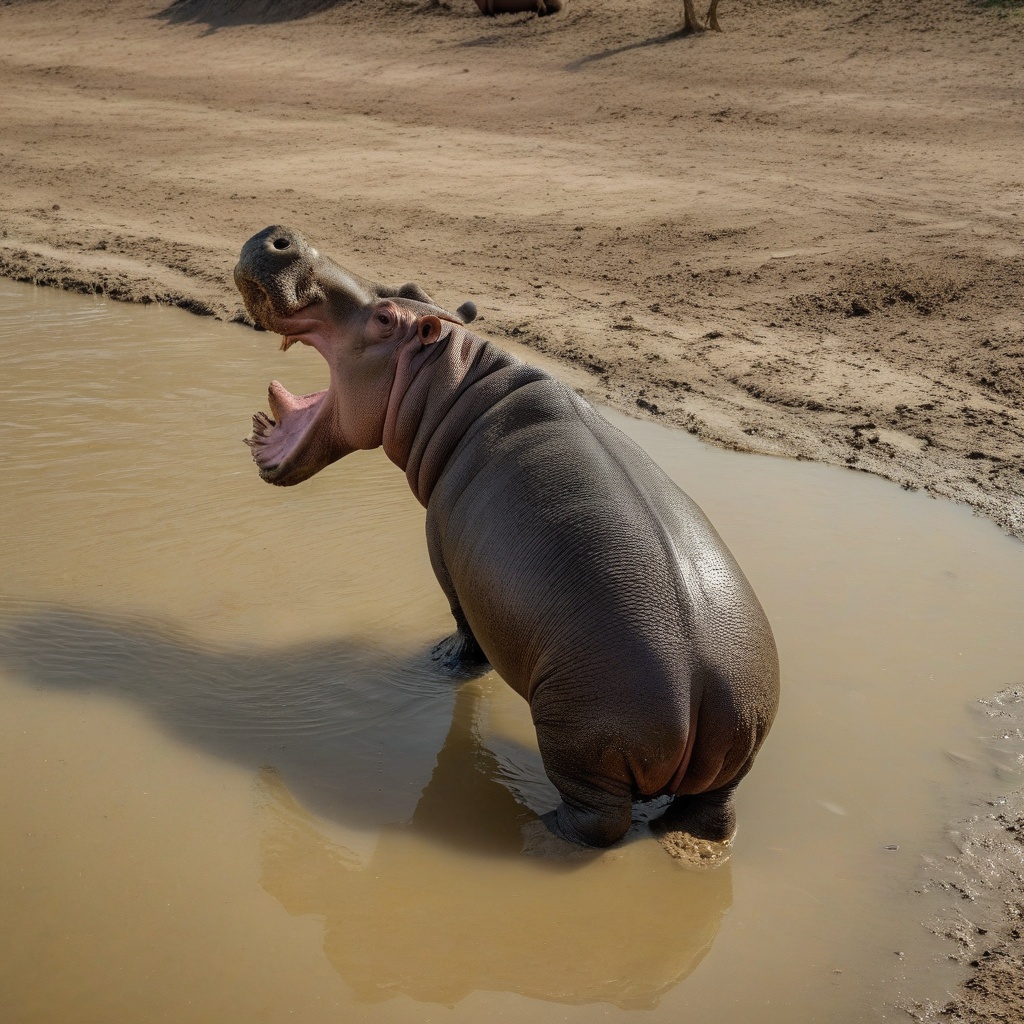} &    
   \includegraphics[width=0.145\textwidth]{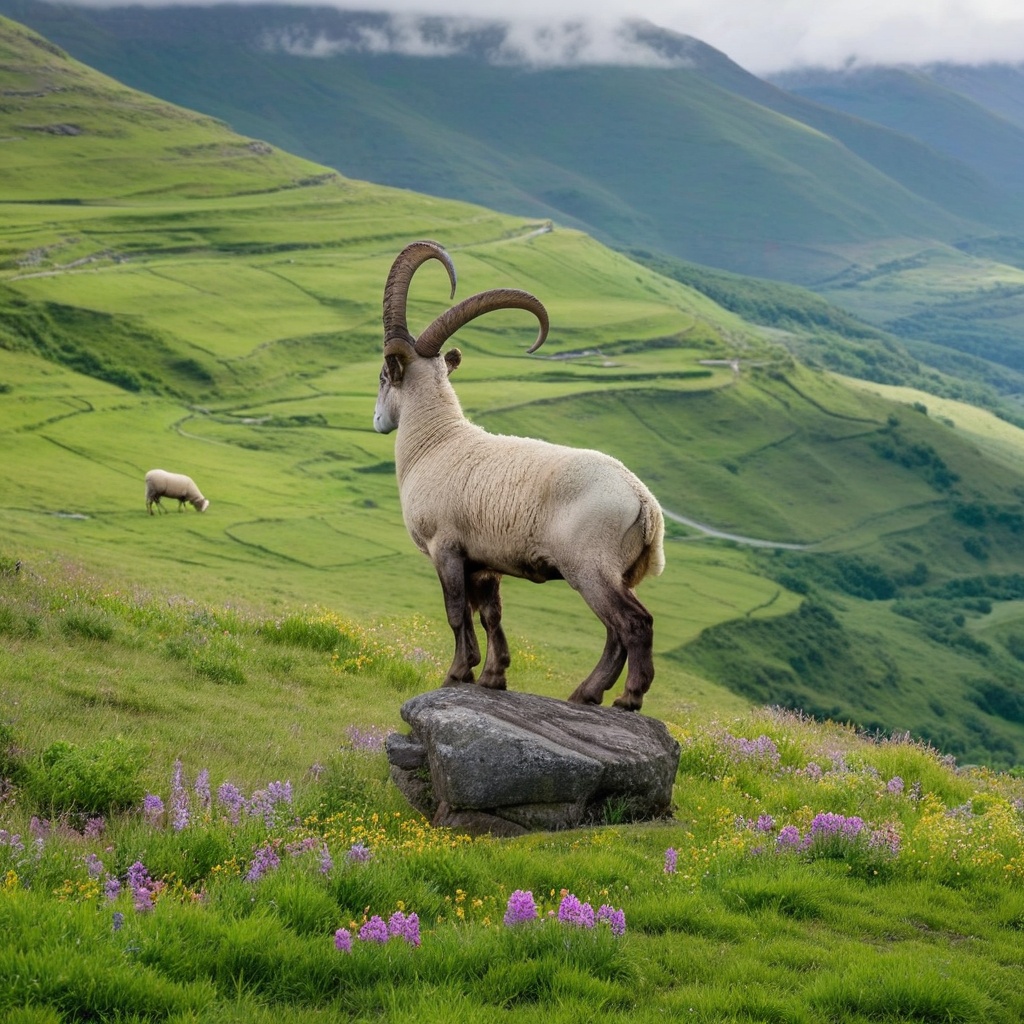} &
    \includegraphics[width=0.145\textwidth]{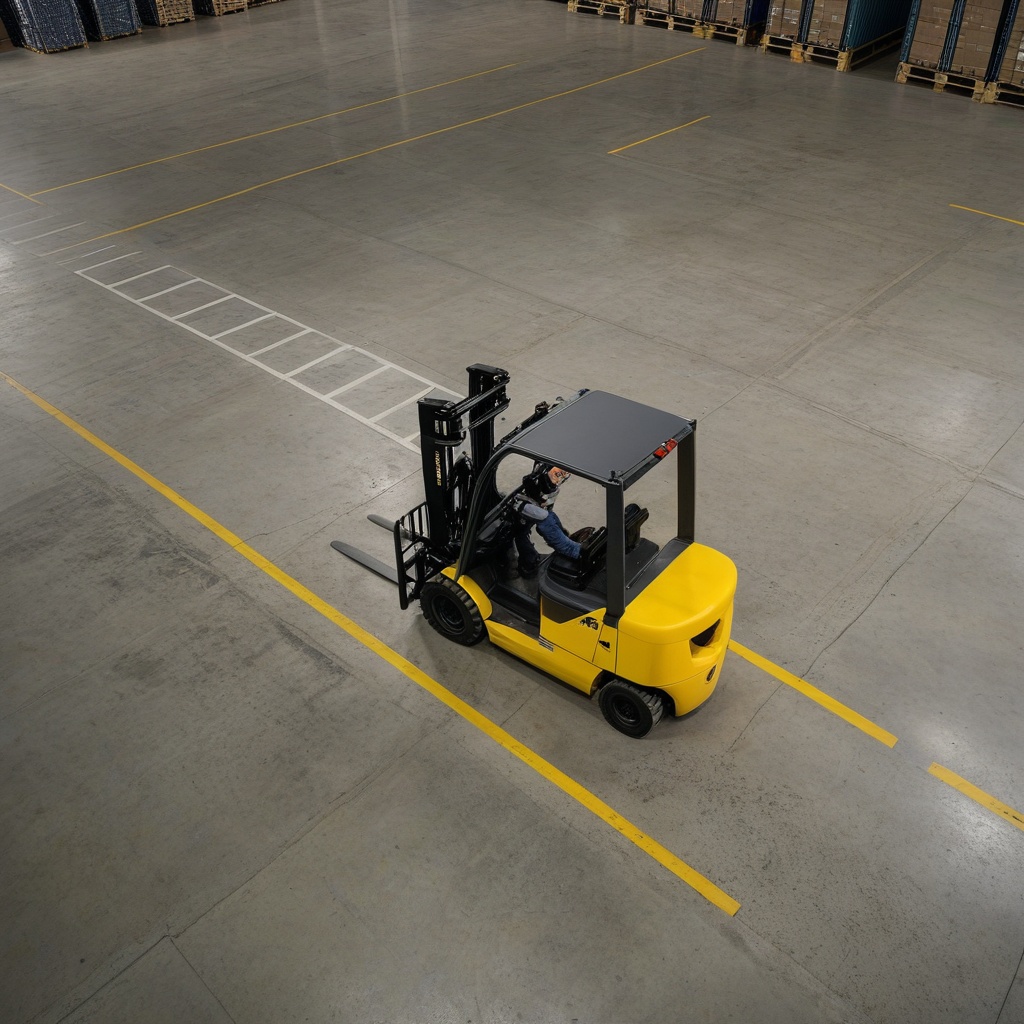} &
    \includegraphics[width=0.145\textwidth]{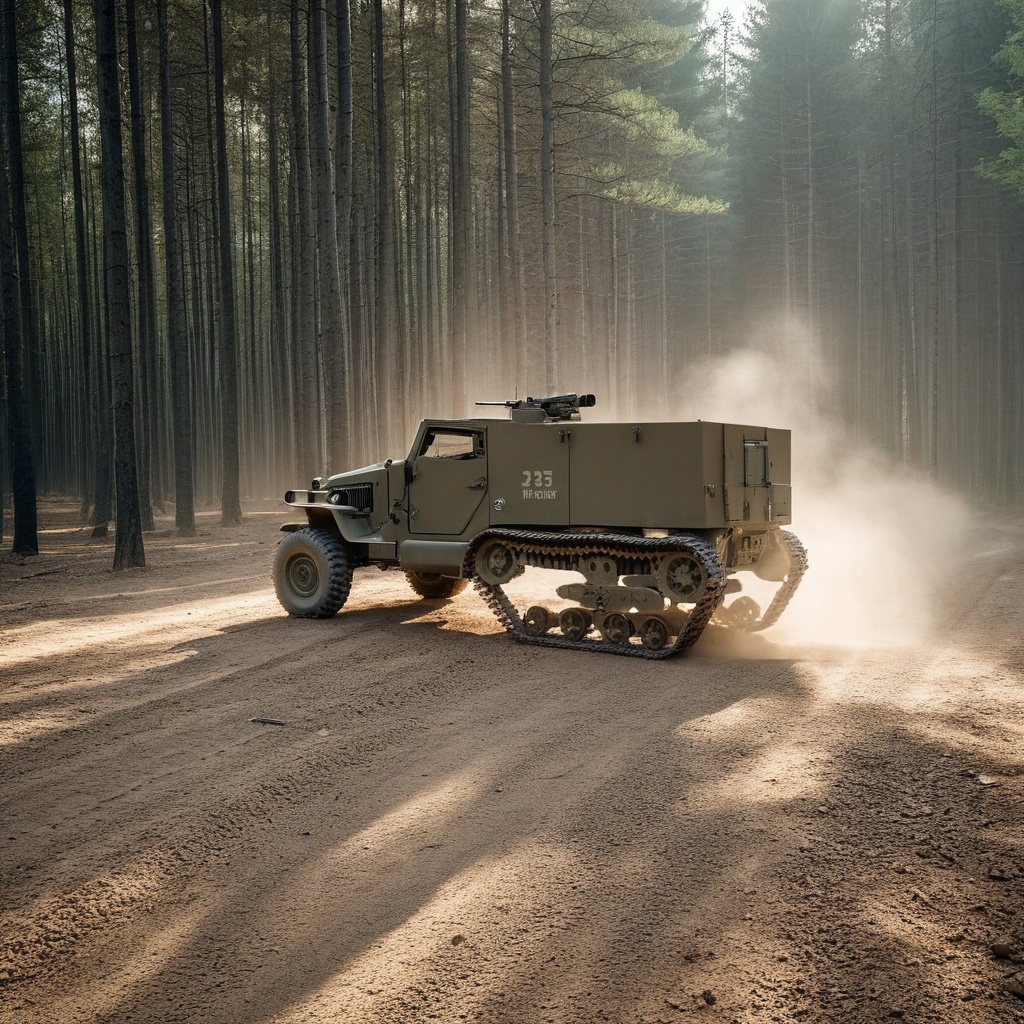} &
    \includegraphics[width=0.145\textwidth]{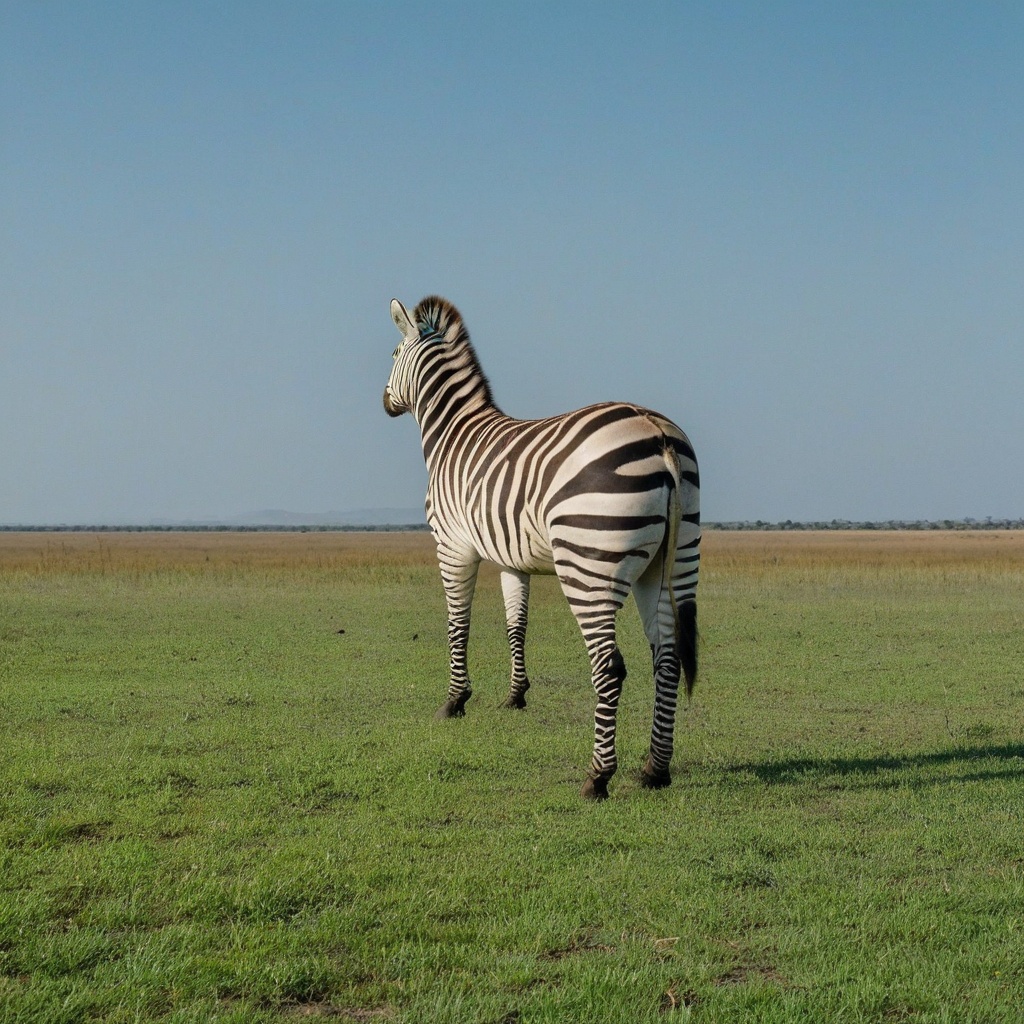} 
    \\
   \rotatebox{90}{\textbf{Back}} & 
    \includegraphics[width=0.145\textwidth]{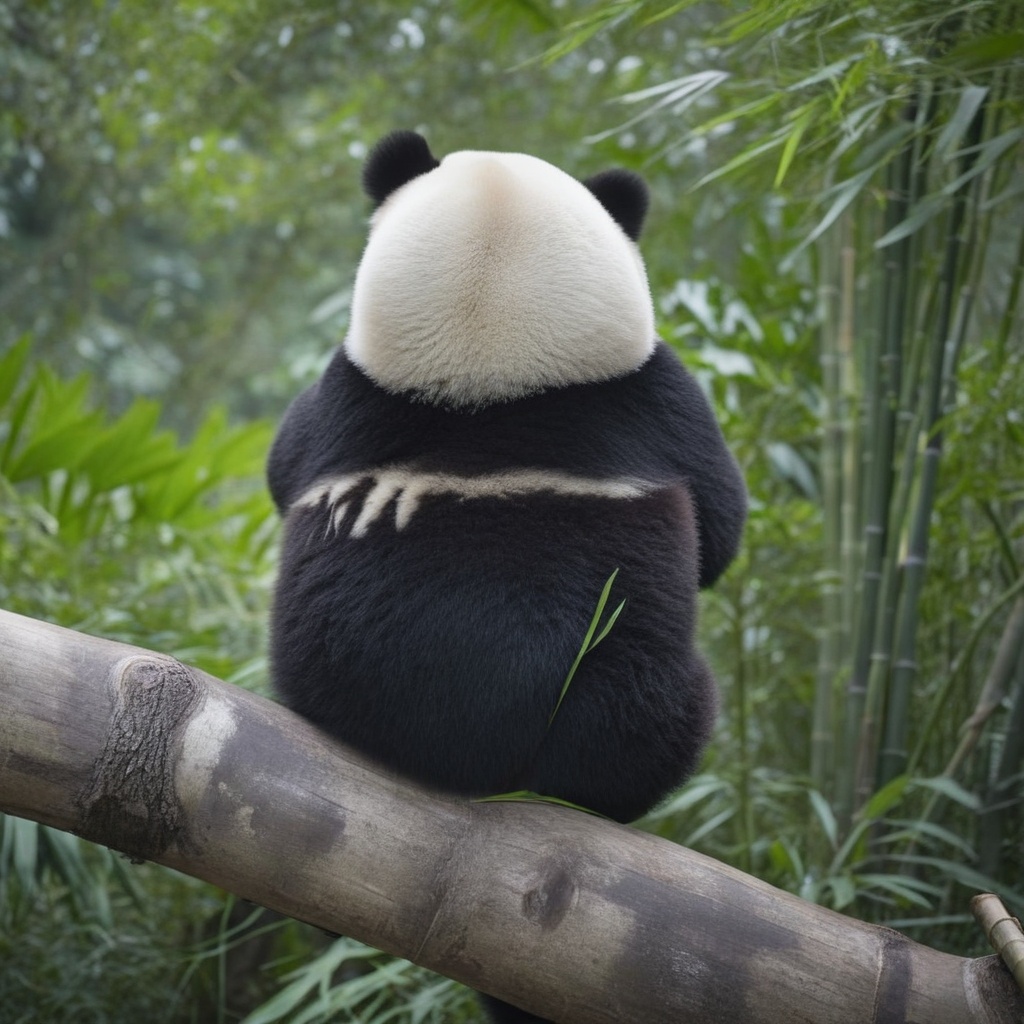} &
   \includegraphics[width=0.145\textwidth]{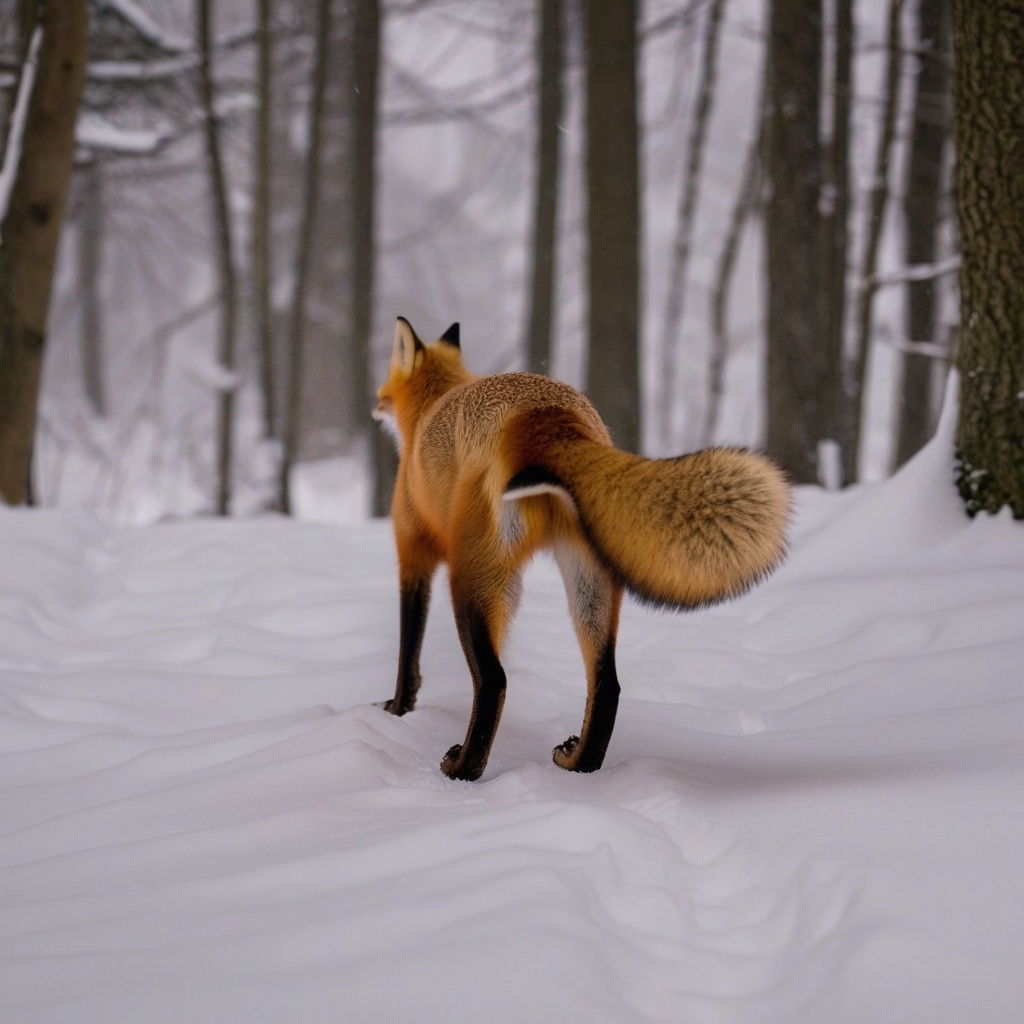} &    
    \includegraphics[width=0.145\textwidth]{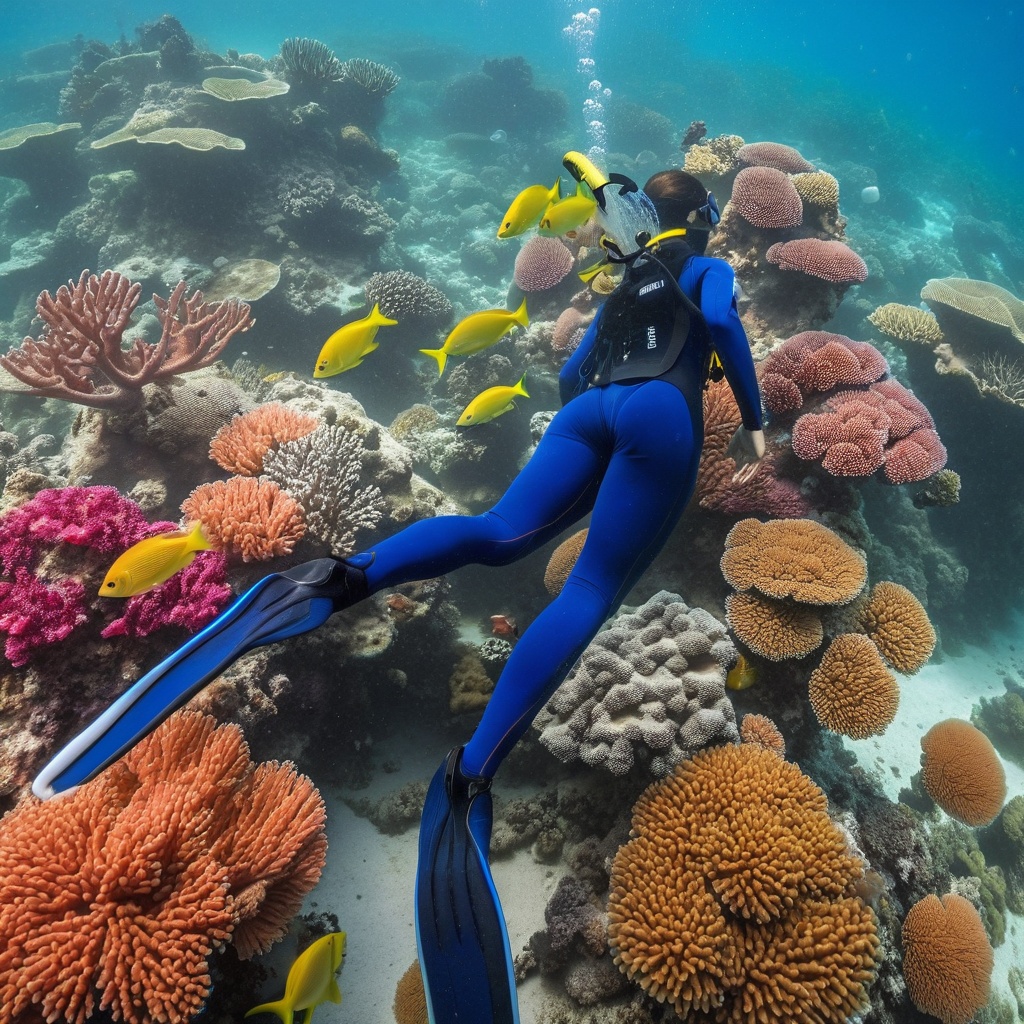} &
    \includegraphics[width=0.145\textwidth]{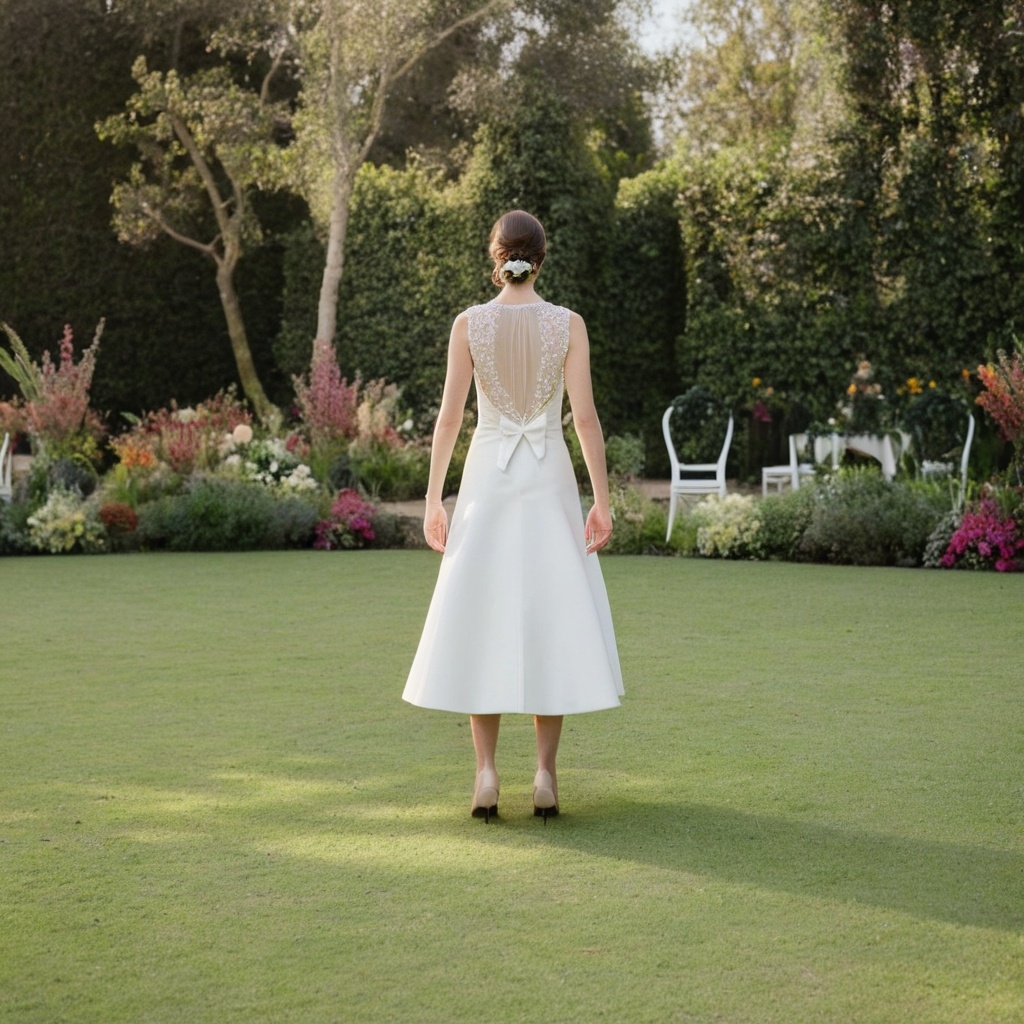} &
    \includegraphics[width=0.145\textwidth]{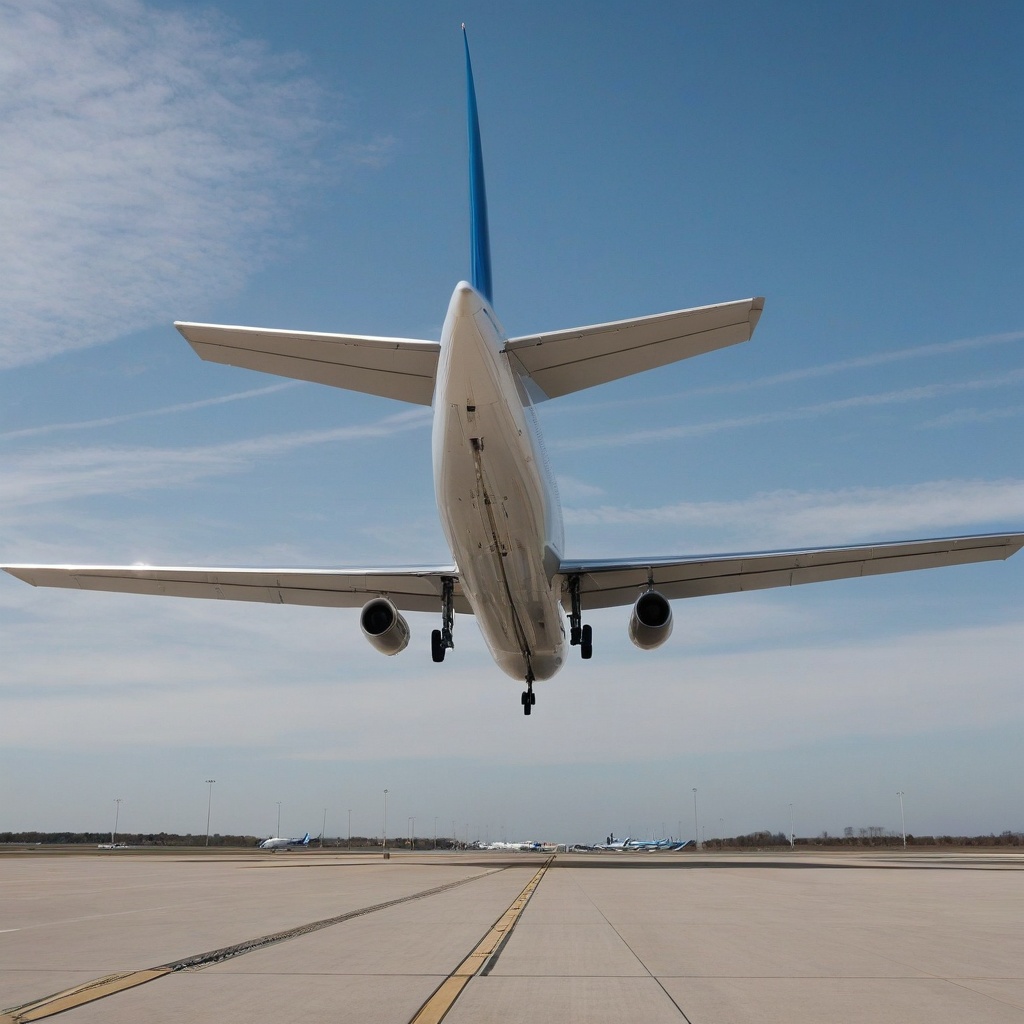} &
   \includegraphics[width=0.145\textwidth]{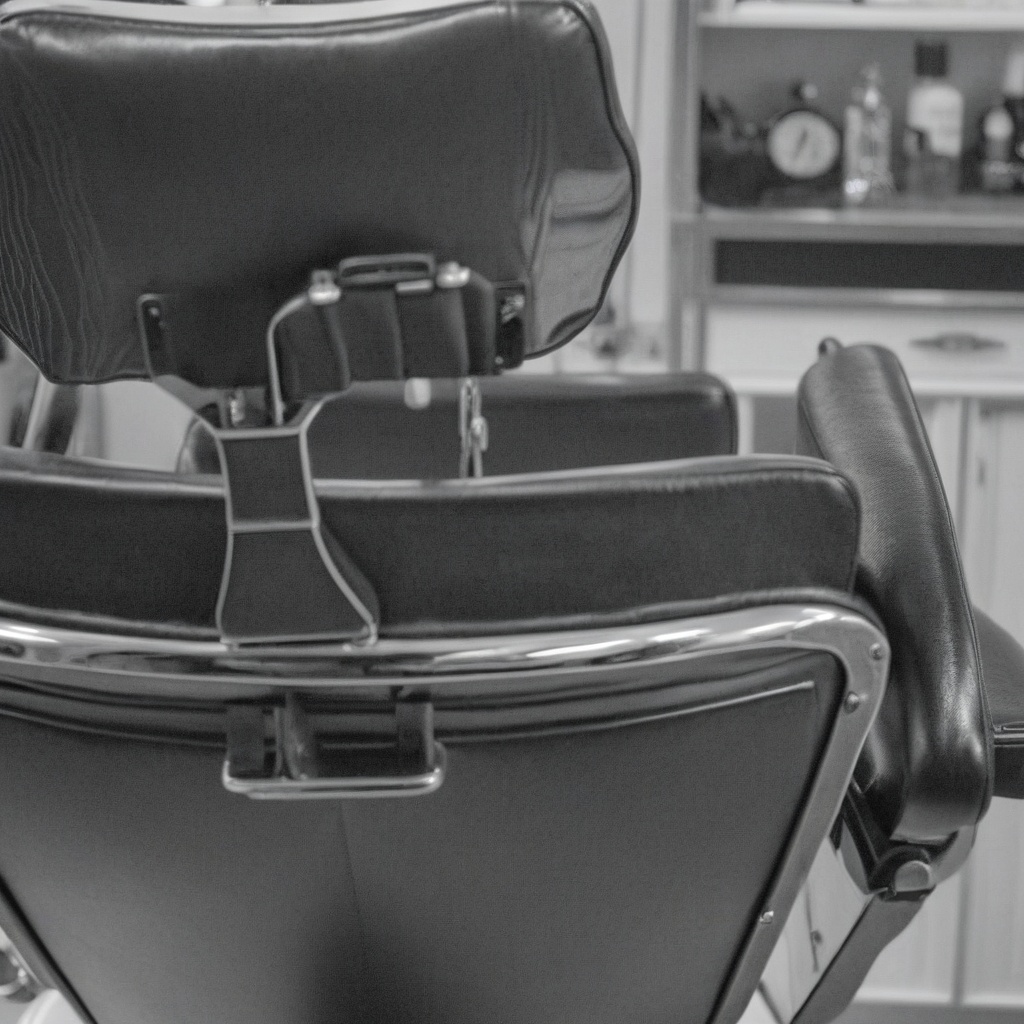} 

    \\
   \rotatebox{90}{\textbf{Back Right}} & 
    \includegraphics[width=0.145\textwidth]{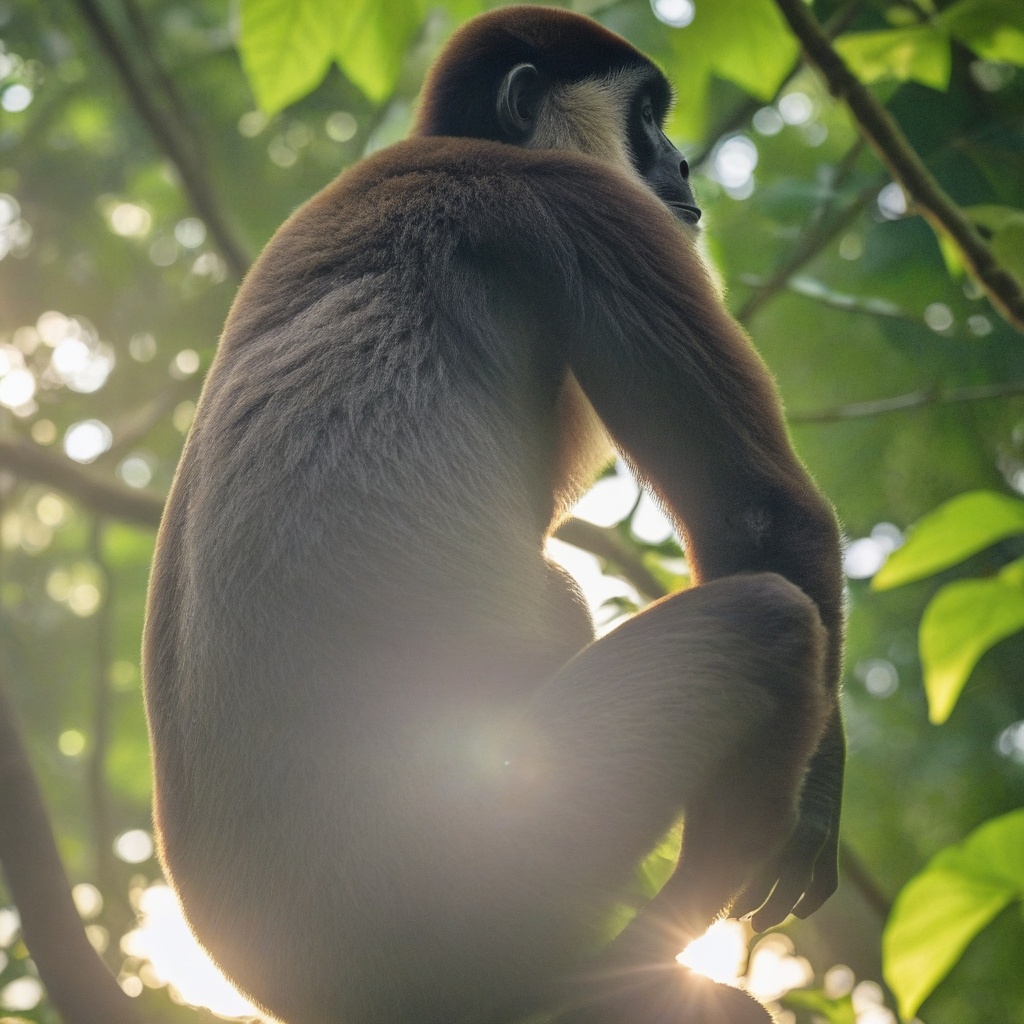} &
   \includegraphics[width=0.145\textwidth]{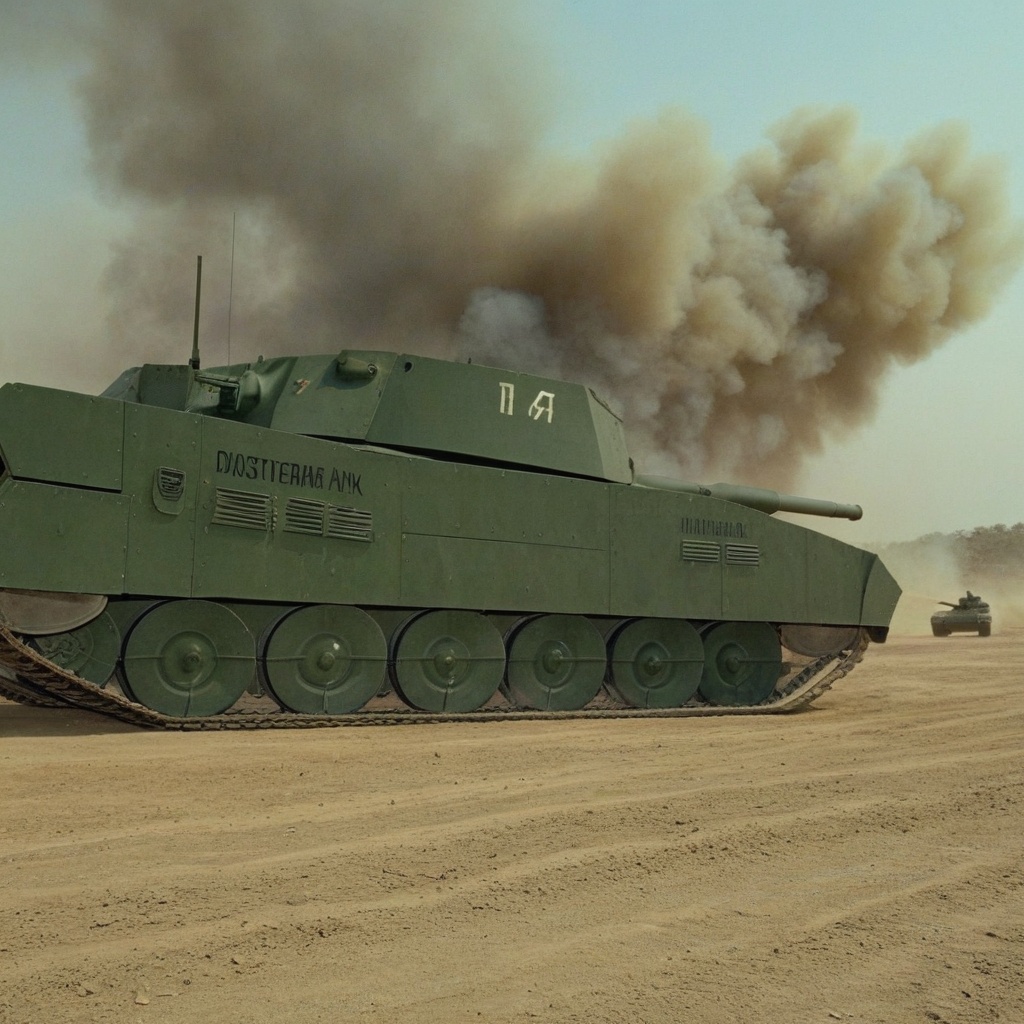} &    
   \includegraphics[width=0.145\textwidth]{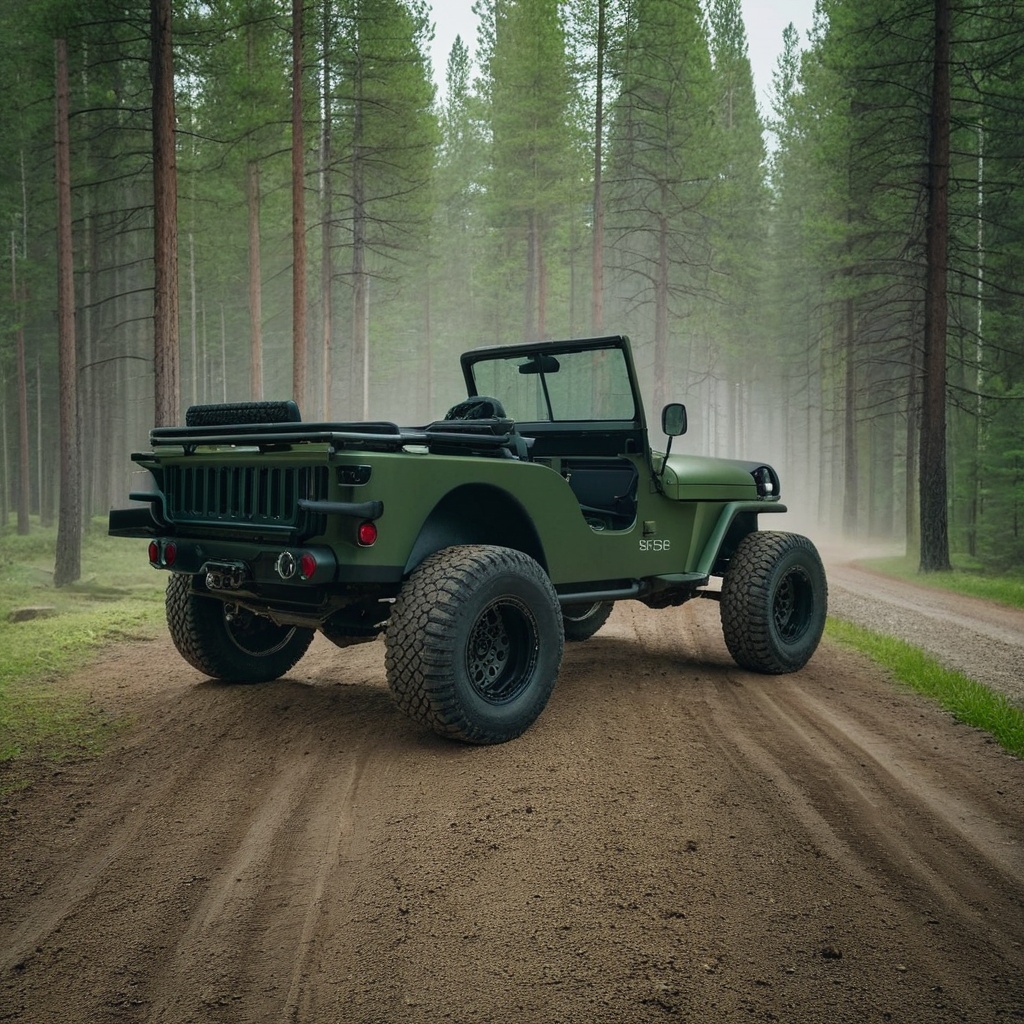} &
    \includegraphics[width=0.145\textwidth]{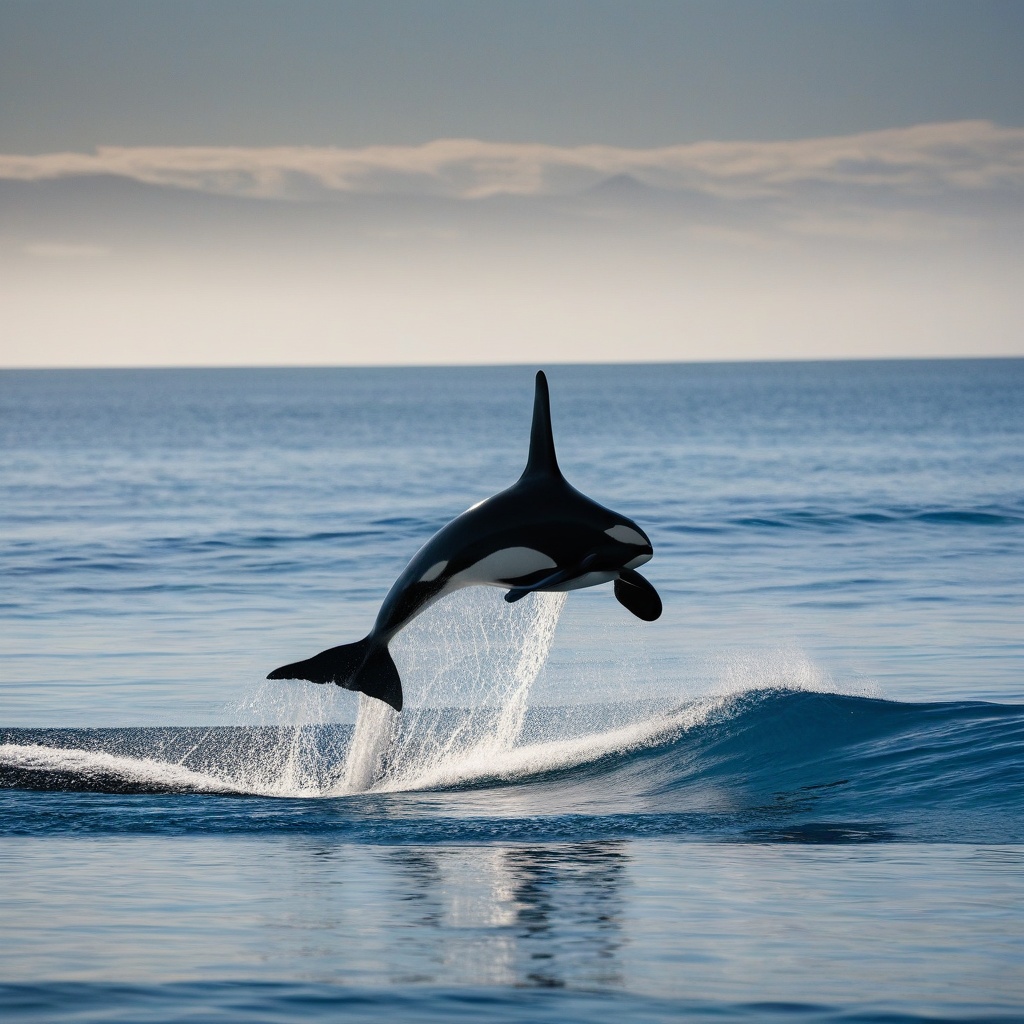} &
    \includegraphics[width=0.145\textwidth]{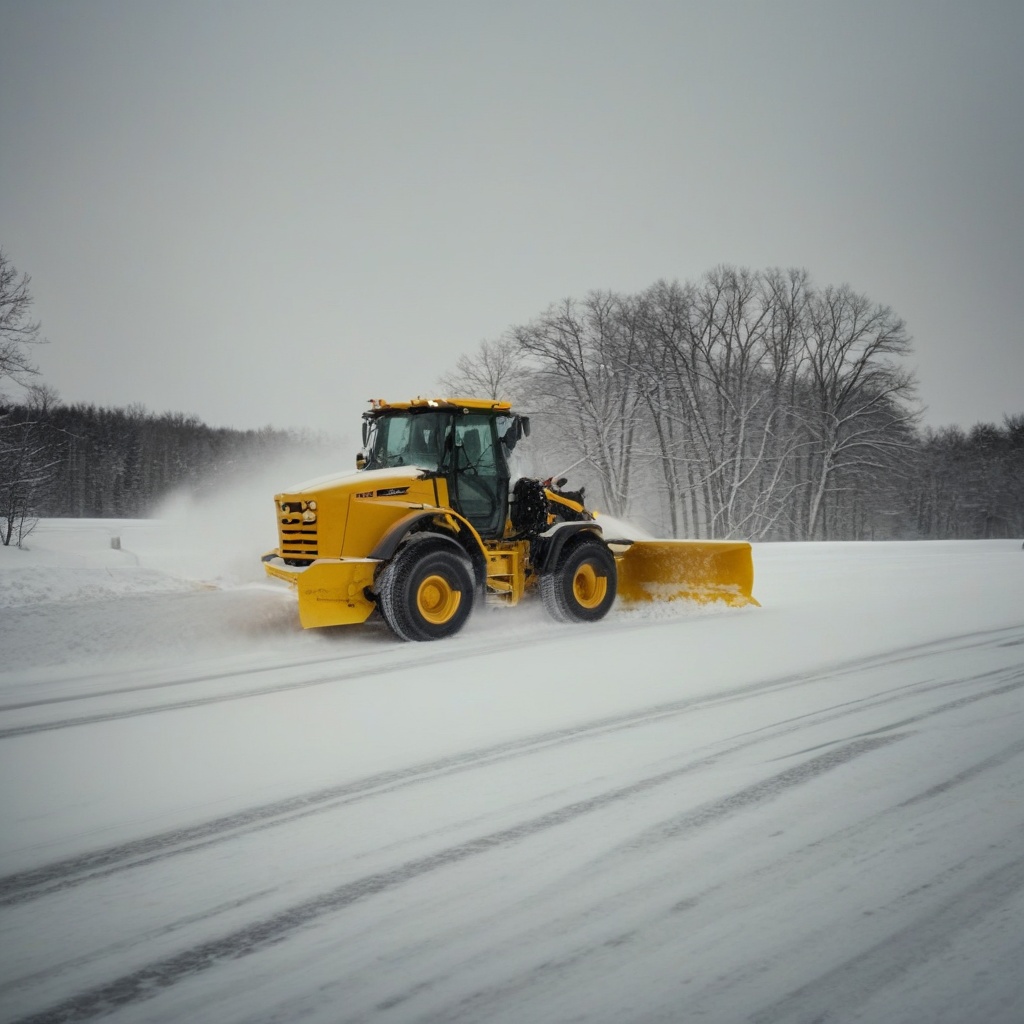} &
    \includegraphics[width=0.145\textwidth]{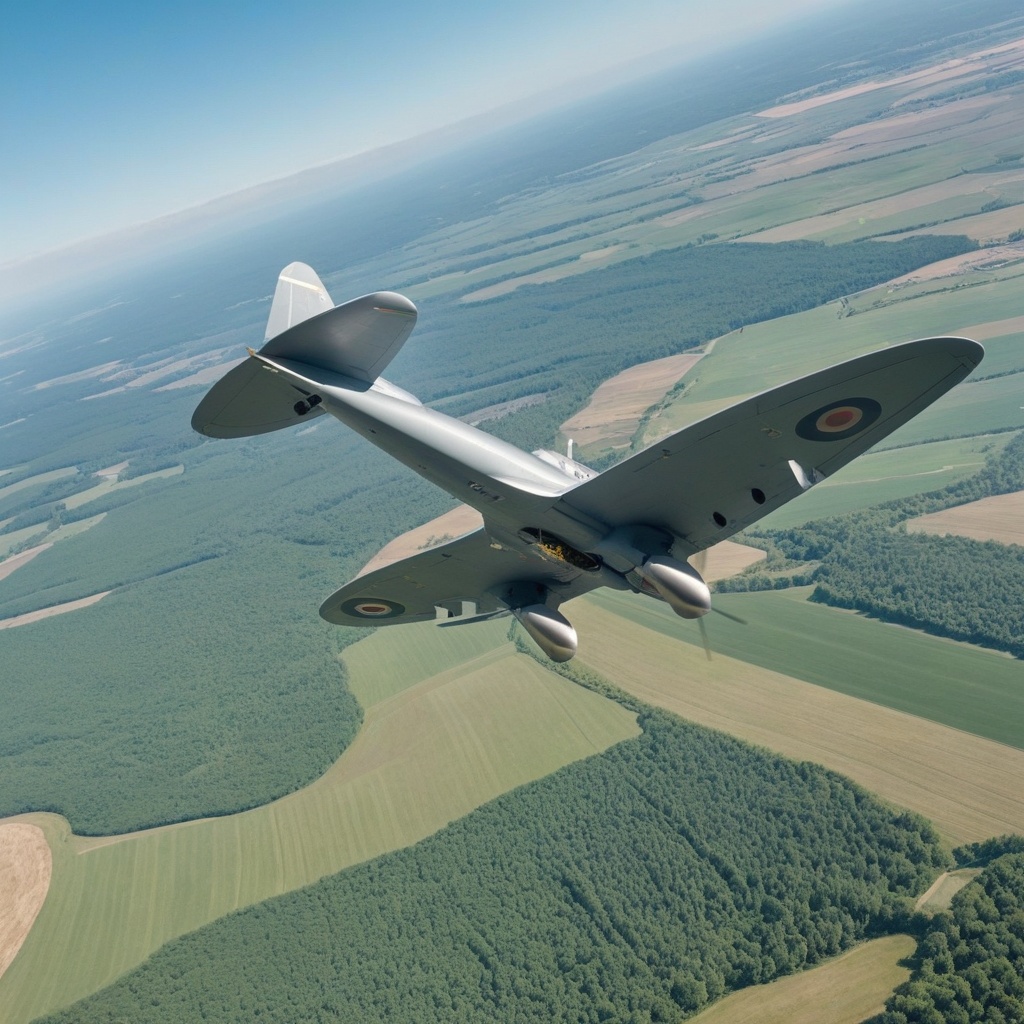} 
    \\
   \rotatebox{90}{\textbf{Right}} & 
    \includegraphics[width=0.145\textwidth]{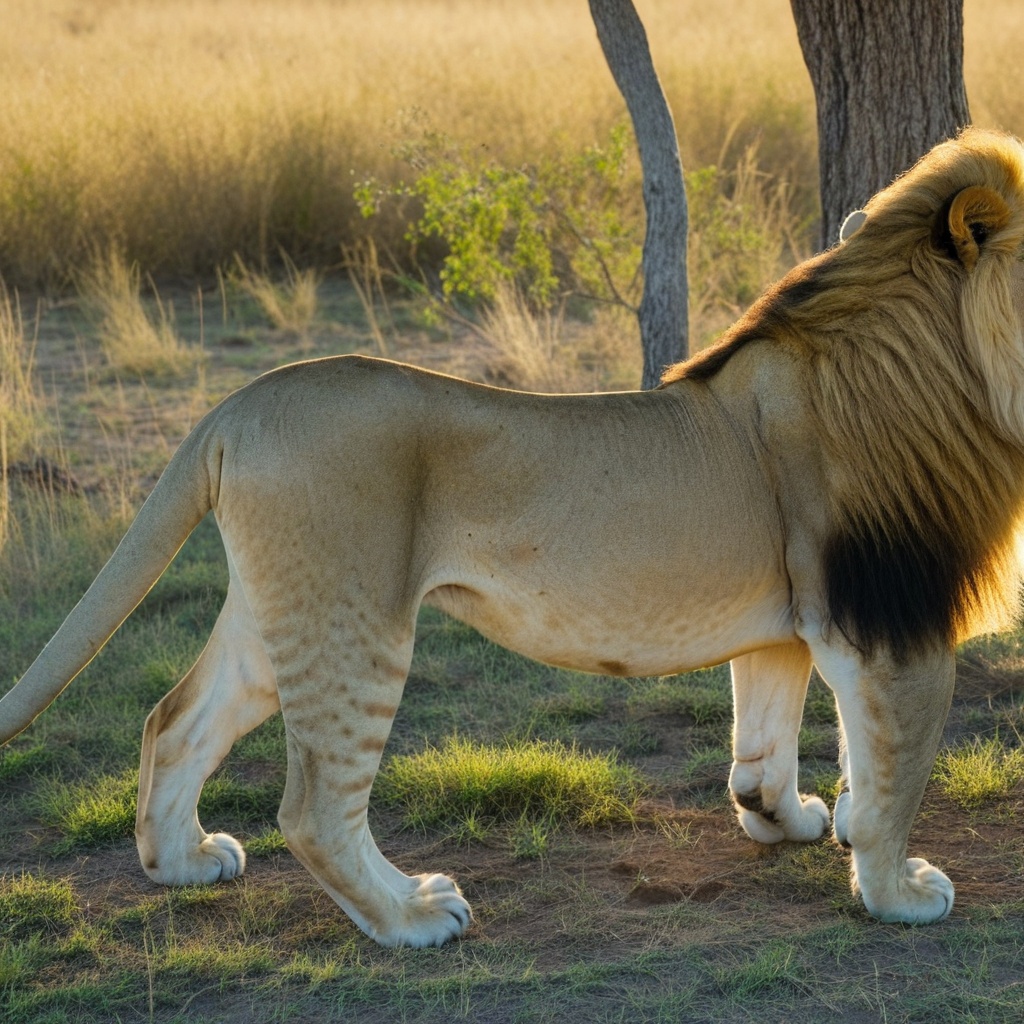} &
    \includegraphics[width=0.145\textwidth]{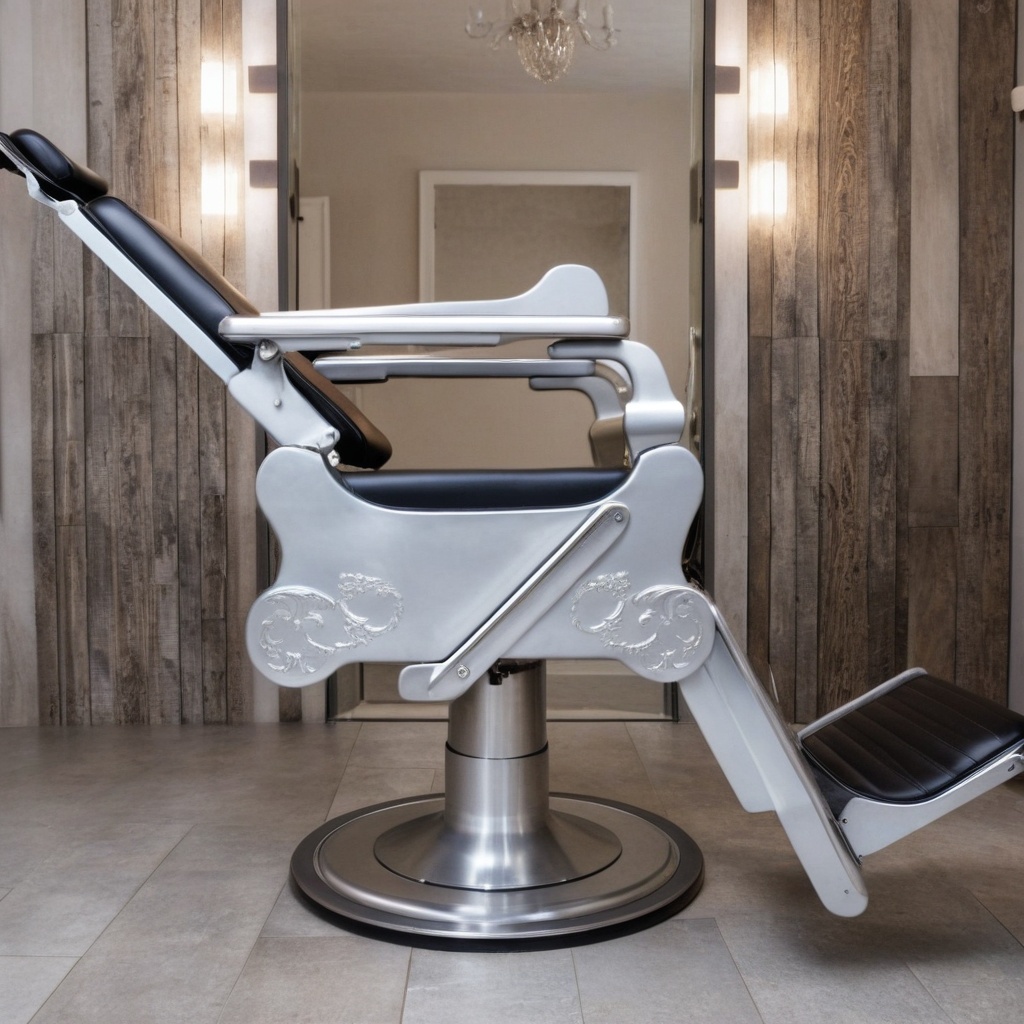} &
   \includegraphics[width=0.145\textwidth]{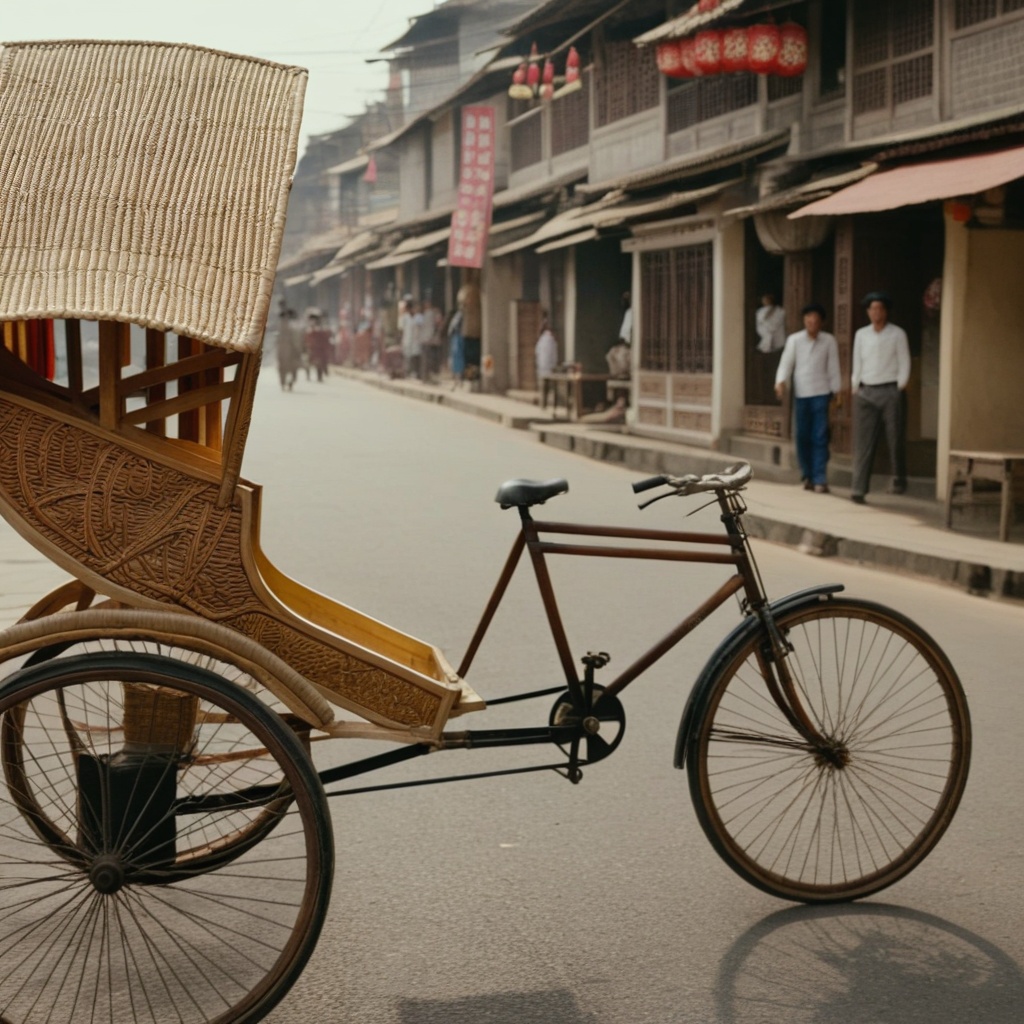} &    
   \includegraphics[width=0.145\textwidth]{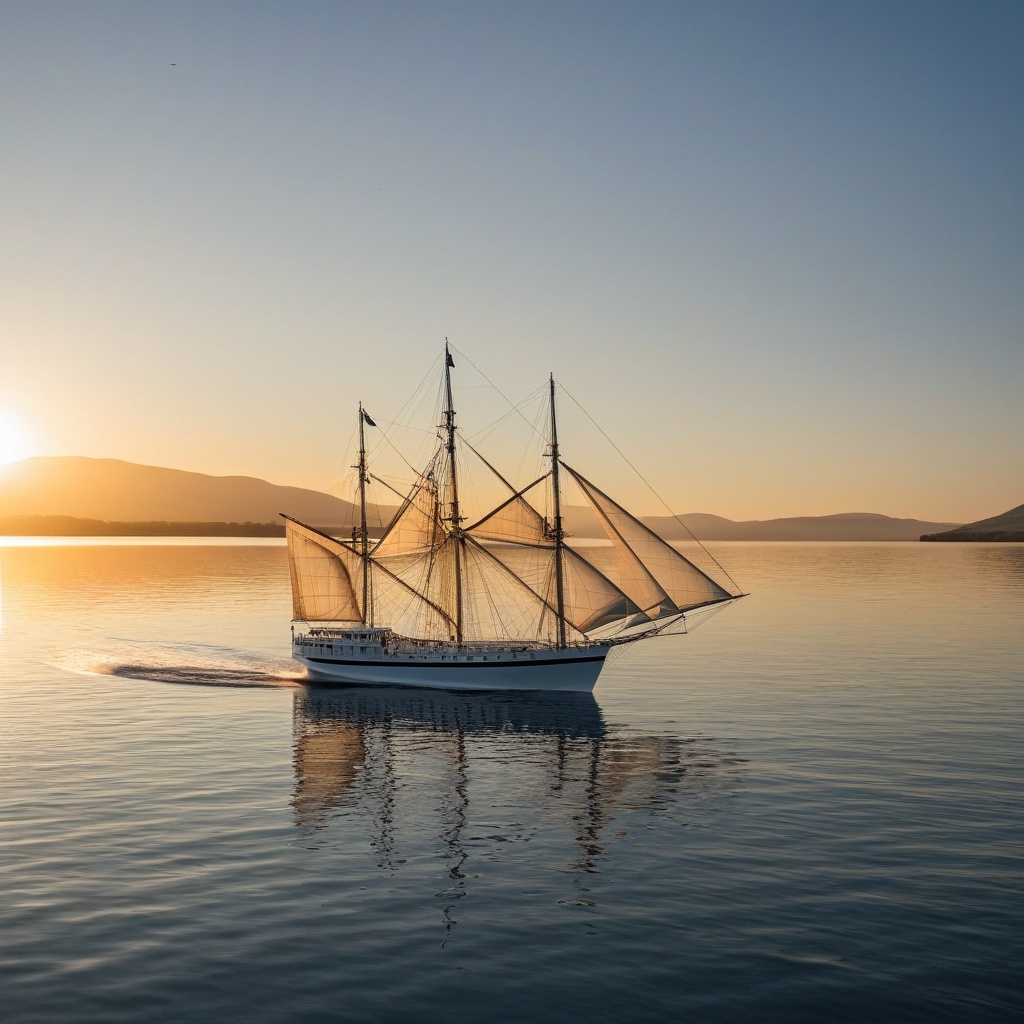} &
    \includegraphics[width=0.145\textwidth]{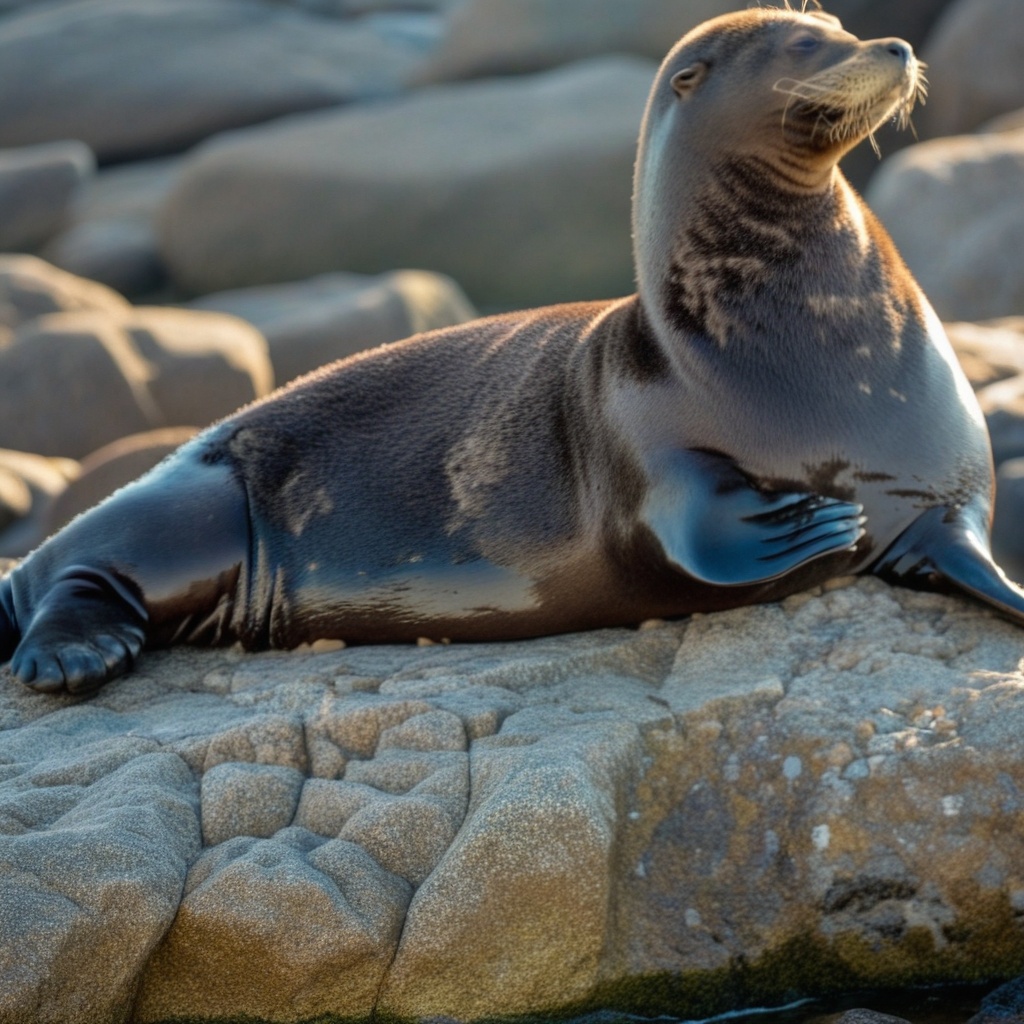} &
    \includegraphics[width=0.145\textwidth]{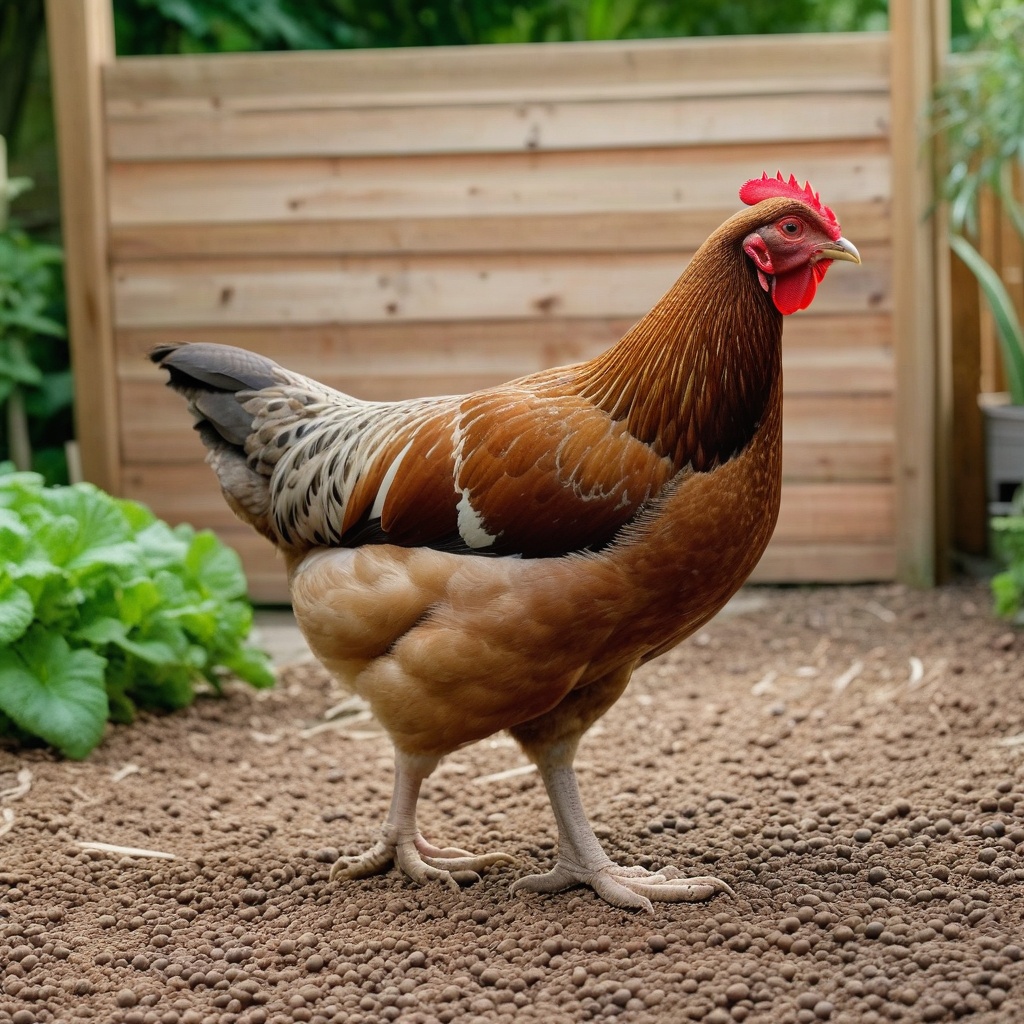} 
    \\
   \rotatebox{90}{\textbf{Front Right}} & 
   \includegraphics[width=0.145\textwidth]{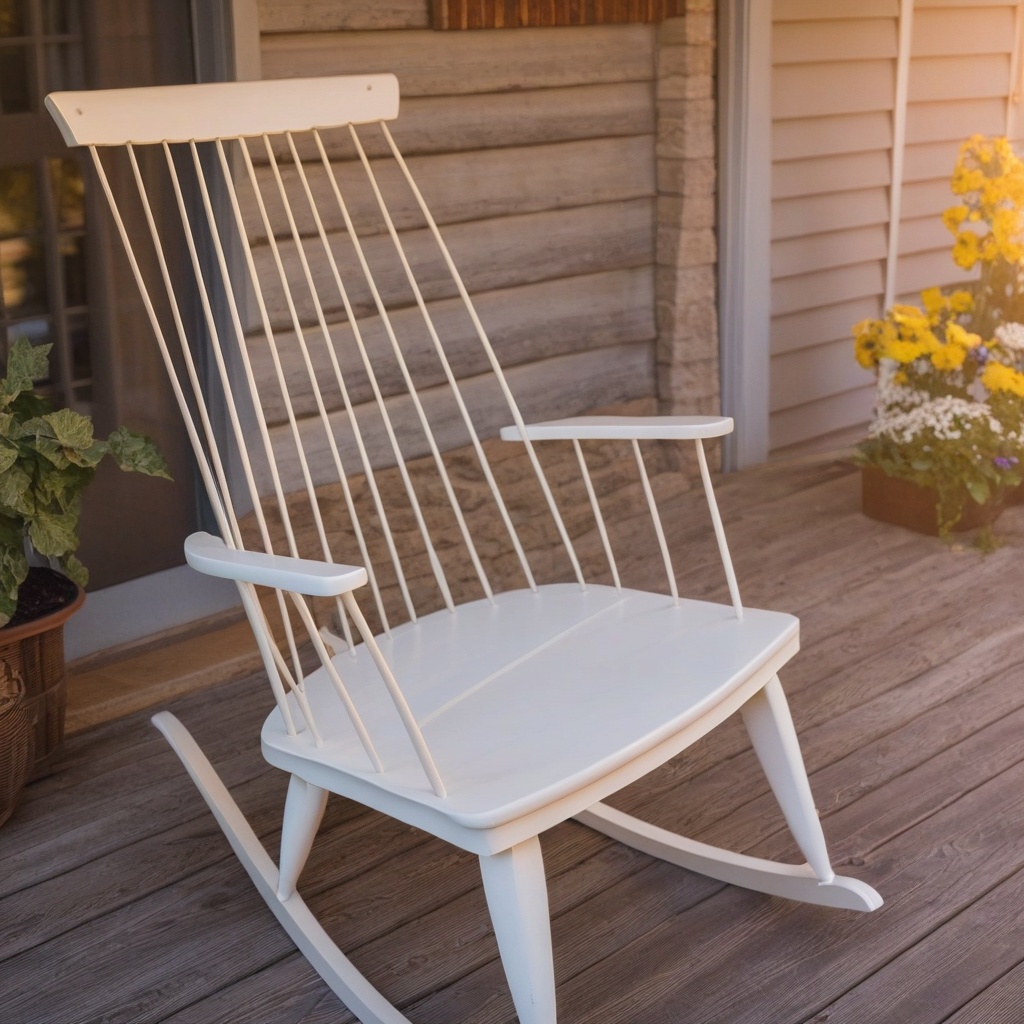} &    
   \includegraphics[width=0.145\textwidth]{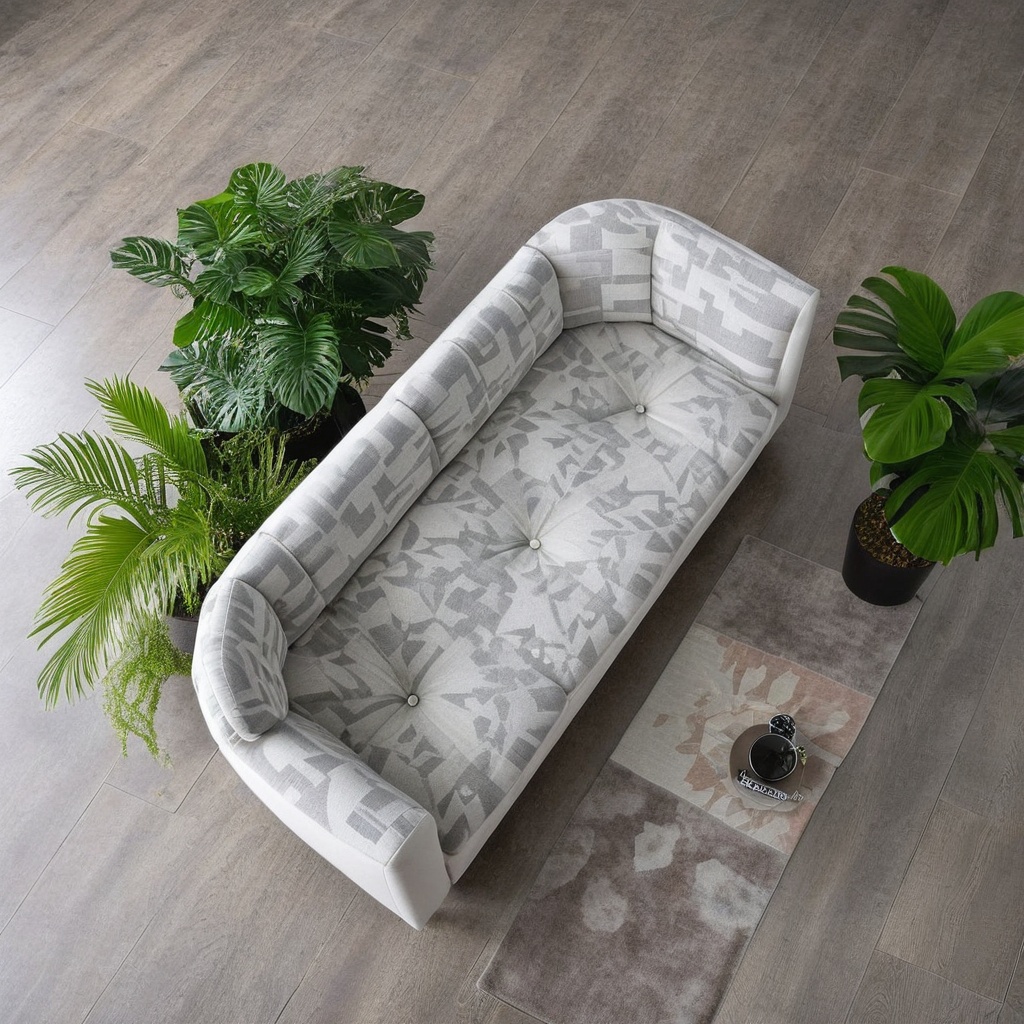} &
    \includegraphics[width=0.145\textwidth]{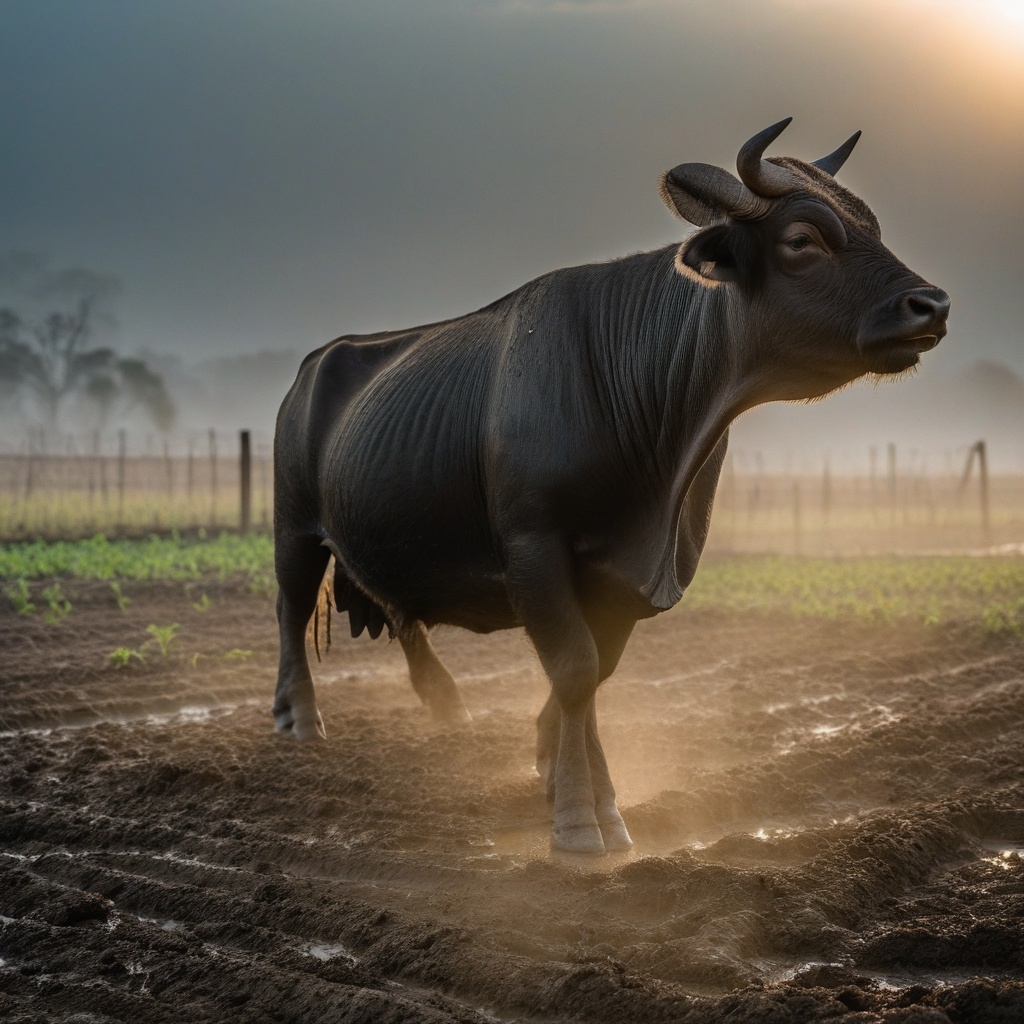} &
    \includegraphics[width=0.145\textwidth]{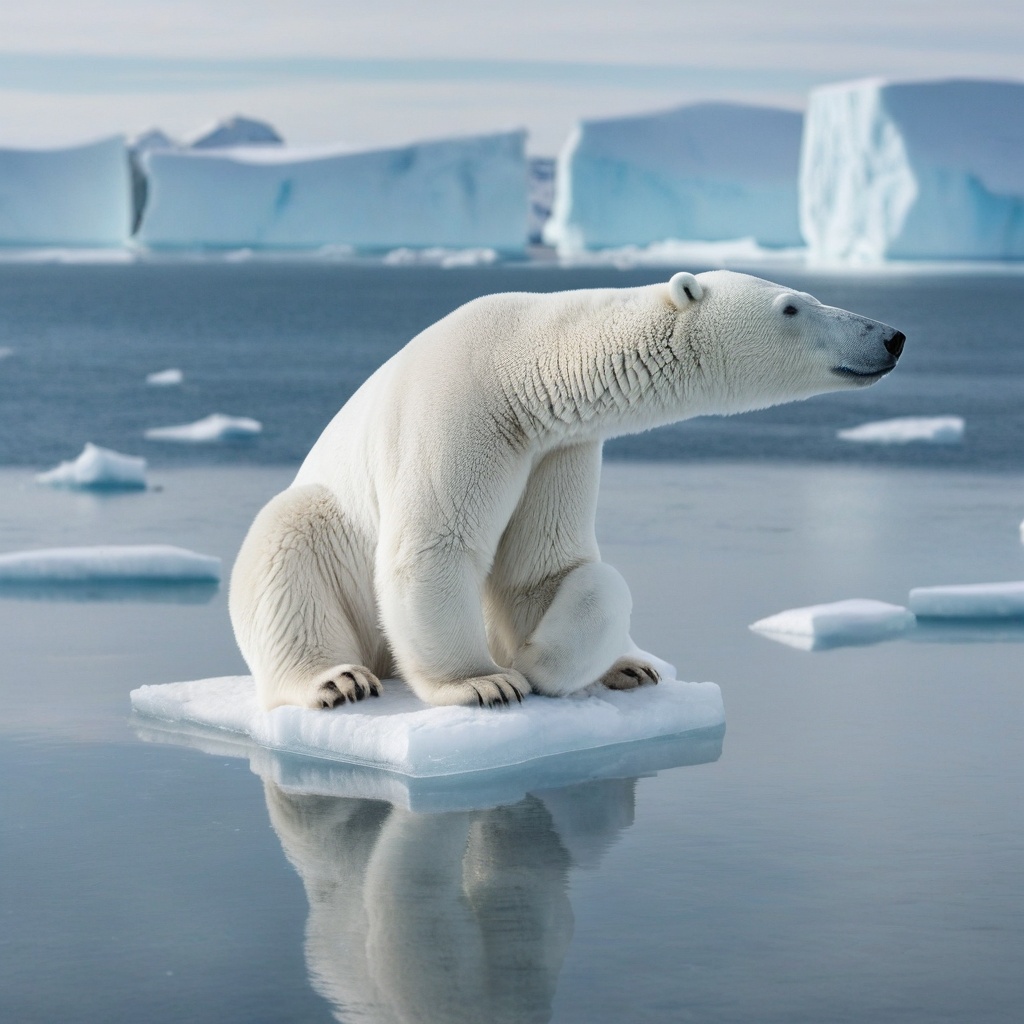} &
    \includegraphics[width=0.145\textwidth]{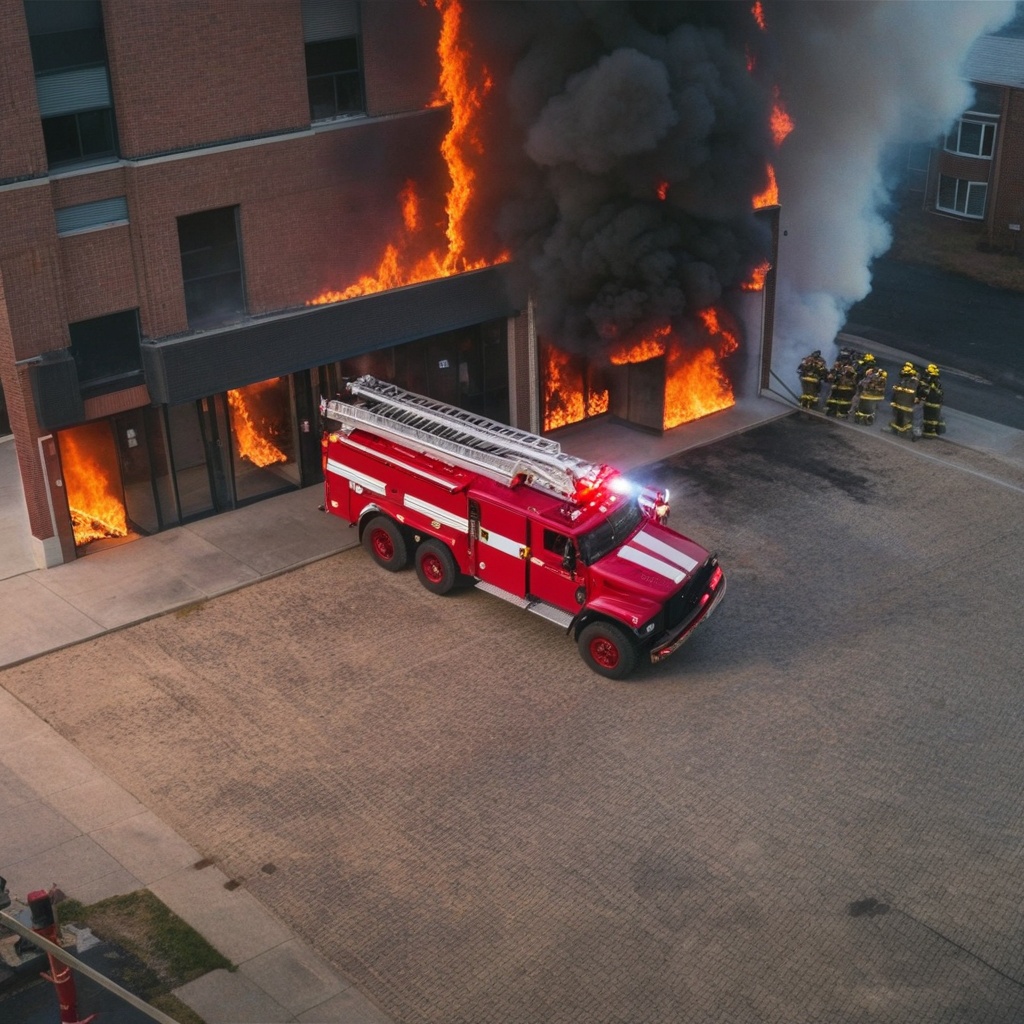} &
    \includegraphics[width=0.145\textwidth]{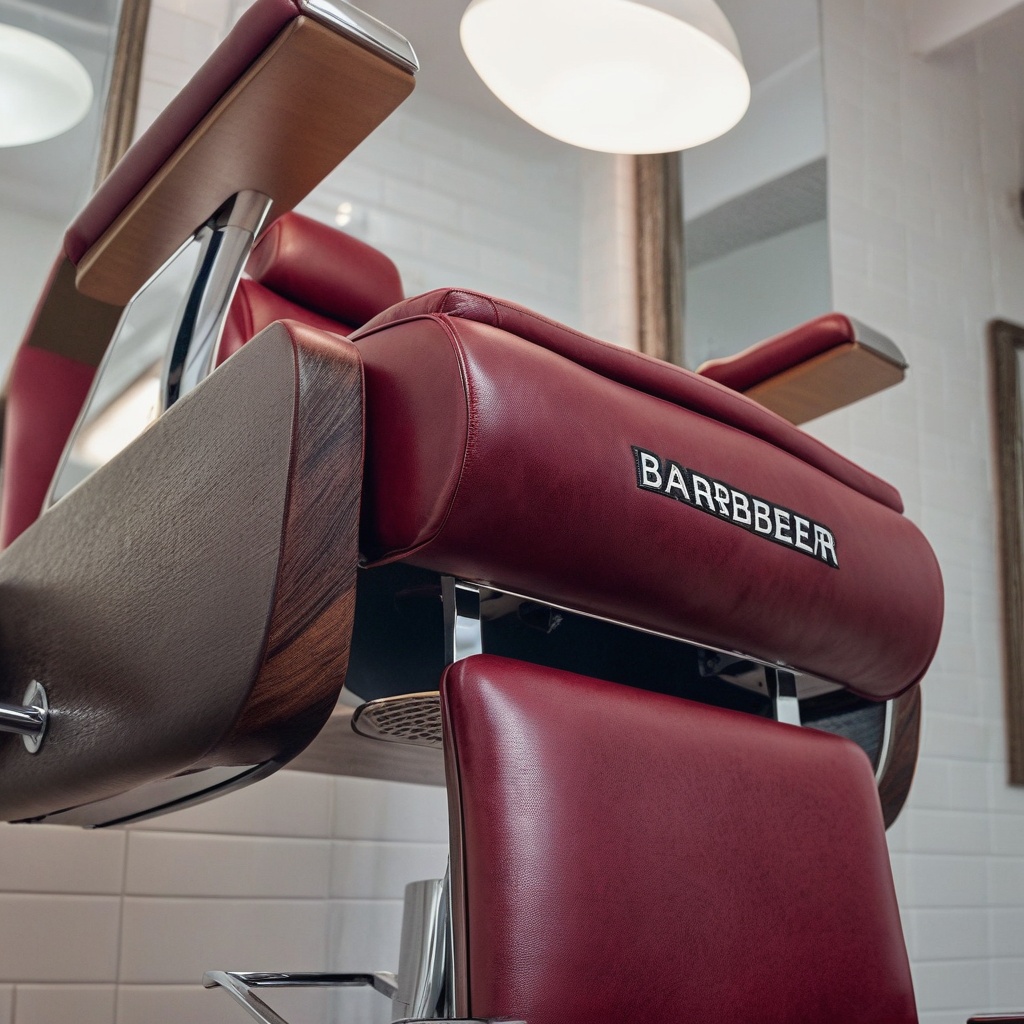}
    \\

\end{tabular}%
     \caption{\textbf{Diversity of \textit{Ultimate3D}}. Our \textit{Ultimate3D} dataset and benchmark cover 100 categories of objects, range diverse camera-object relation settings, and provide plausible image quality. (Each row shows images with the same orientation but in diverse subject and context.)}
      \label{suppfig:ultimate3D}
\end{figure*}

\FloatBarrier
\section{Algorithm of 3D visual instruction dataset generation pipeline}
\label{suppsec:algorithm}

\begin{algorithm}[!ht]
\caption{Synthetic VQA generation}
\label{alg:algorithm}
\textbf{Input}: 3D asset $\mathcal{A}$, asset category $c$, camera-object relation $\beta$.\\
\textbf{Parameter}: Renderer $\mathcal{R}$, DM-based image generator $\mathcal{G}$, image decoder $\mathcal{D}$, LLM text generator $\mathcal{L}$, system prompt given to $\mathcal{L}$ for generating image context description $p_{img}$, and for generating QA pairs $p_{qa}$, DM sampling steps $T$. \\
\textbf{Output}: Generated synthetic image $I_{syn}$, corresponding text QA pairs $\mathcal{T}_{qa}$. 
\begin{algorithmic}[1] 
    \State $I_{\beta} = \mathcal{R}(A, \beta)$
    \dotfill $\#$ Render 3D priors

    \State  $\mathcal{T}_{img} = \mathcal{L} (c, p_{img})$
    \dotfill $\#$ Generate image description

    \State $z_T \sim \mathcal{N}(0, I)$
    \For{$t=T, \cdots, 1$} 
        \State $z_{t-1} = \mathcal{G}(z_{t}, I_{\beta}, \mathcal{T}_{img})$ \dotfill $\#$ DM denoising 
    \EndFor
    \State $I_{syn} = \mathcal{D}(z_0)$
    
    \State $\mathcal{T}_{qa} = \mathcal{L} (c, \beta, p_{qa})$
    \dotfill $\#$ Generate QA pairs 

    \State \textbf{return} $I_{syn}, \mathcal{T}_{qa}$
    
\end{algorithmic}
\end{algorithm}    




\end{appendices}

\clearpage
\clearpage
\bibliography{sn-bibliography}

\end{document}